\documentclass[10pt,twocolumn,letterpaper]{article}

\usepackage[pagenumbers]{cvpr} %

\usepackage{tabularx}
\usepackage{multirow}
\usepackage{colortbl}

\definecolor{cvprblue}{rgb}{0.21,0.49,0.74}
\usepackage[pagebackref,breaklinks,colorlinks,allcolors=cvprblue]{hyperref}

\def\confName{CVPR}
\def\confYear{2027}

\title{RSReasoner: Reformulating and Auditing Geospatial Reasoning Segmentation}

\author{
Xu Zhang, Xinpeng Zhang, Yunrui Zou, Hongliang Chen,
Xiaotian Jiang\\ Yang Zheng, Kaitai Guo, Jimin Liang\thanks{Corresponding author (jimleung@mail.xidian.edu.cn)}\\
[0.4em]
Xidian University
}

\begin{document}
\maketitle

\begin{abstract}

Geospatial reasoning segmentation aims to localize geographic targets from language requiring semantic inference beyond direct visual perception. 
However, existing benchmarks often operationalize reasoning as category omission while overlooking attribute mixing, which neglects explicit-category reasoning, introduces perceptual shortcuts, and fails to precisely evaluate reasoning capability. 
To diagnose and address these limitations, we first reformulate geospatial reasoning segmentation along category expression (explicit vs.\ implicit) and attribute nature (perceptual vs.\ inferential), and audit 90,708 queries across existing benchmarks to reveal that pure explicit-inferential queries account for only 0.61\% whereas 31.98\% mix both attribute types. 
We then develop a three-stage pipeline that harnesses OpenStreetMap attributes while mitigating geometric and semantic noise to construct \textbf{RSReasoner}, a large-scale dataset of 106,610 images with 213,220 reasoning queries, and \textbf{RSReasoner-Bench}, a diagnostic benchmark of 75,400 queries across four canonical types and controlled variants. 
Extensive experiments demonstrate that RSReasoner-Bench precisely evaluates model capability across multiple dimensions while exposing shortcut behaviors, and fine-tuning on RSReasoner boosts both reasoning and cross-type generalization. 
Our findings offer key insights for future geospatial reasoning research and benchmark design.
Datasets and code will be released at \url{https://github.com/Ricardo-XZ/RSReasoner}.

\end{abstract}

\section{Introduction}

\begin{figure}[!htbp]
\centering
\includegraphics[width=\columnwidth]{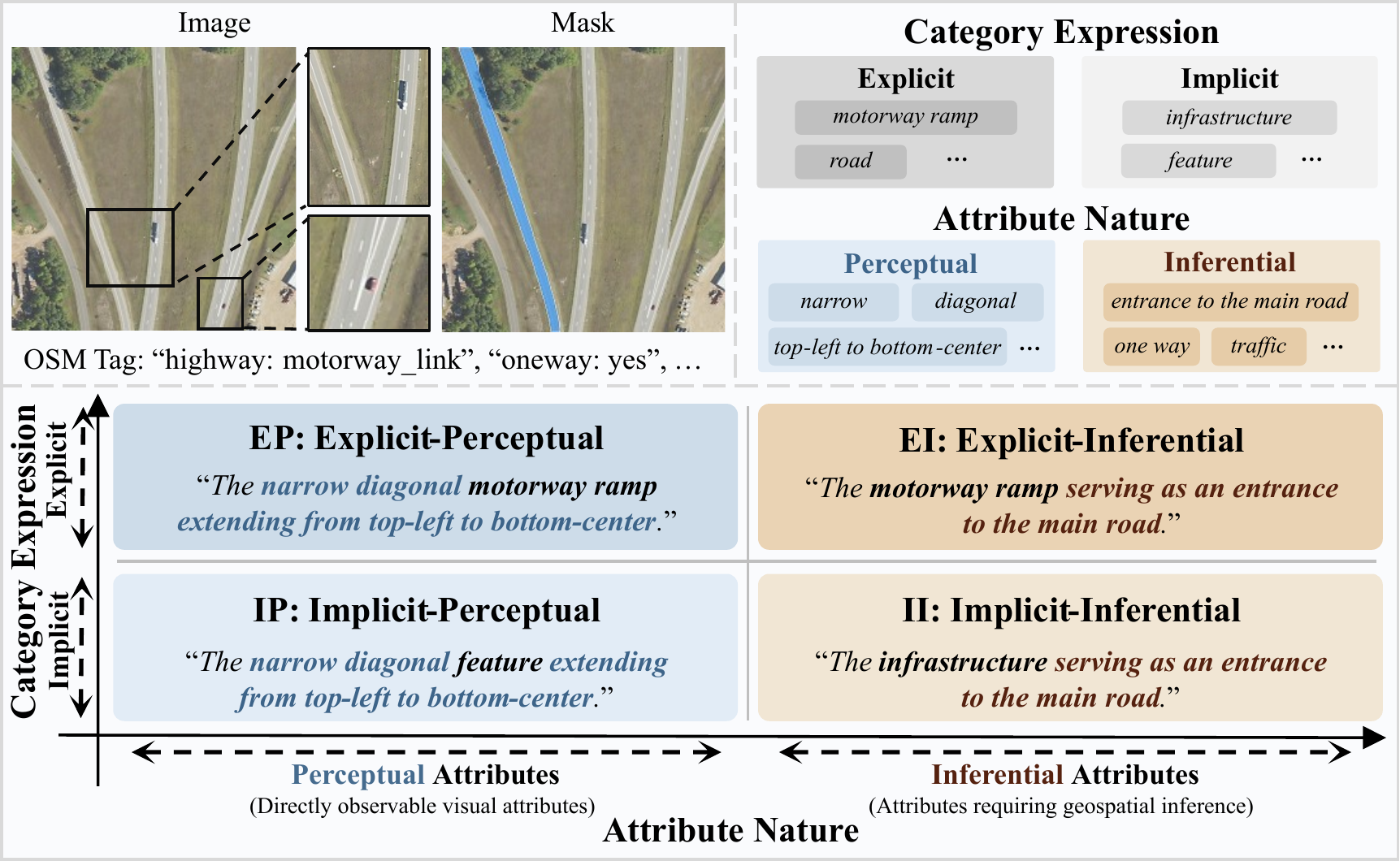}
\caption{Our two-dimensional formulation of geospatial reasoning segmentation along category expression and attribute nature.}
\label{fig:definition}
\end{figure}

Language-guided remote sensing segmentation localizes geographic targets from natural language expressions, with early studies focusing on appearance-based referring expressions~\citep{yuan2024rrsis,liu2024rrsis-d,dong2024cross,yao2025remotesam,ou2025geopix,yang2025large,zhou2024geoground}.
To move beyond explicit categories and visual appearance, \citet{lai2024lisa} formulate reasoning segmentation in natural images as predicting a binary mask from a complex and implicit query requiring reasoning or world knowledge, establishing a paradigm widely adopted by subsequent works~\citep{yang2023lisa++,wang2024llmseg,ren2024pixellm,wang2025segllm,jang2025mmr,liu2025segzero,liu2026visionreasoner,rasheed2024glamm}.
Following this implicit-query formulation, recent works extend reasoning segmentation to geospatial scenes and introduce dedicated benchmarks to ground targets from spatial context, domain knowledge, and user intent~\citep{li2025segearthr1,xin2026segearthr2,ni2026unigeoseg,quenum2026lisat,zhang2026bridging,fan2026georeason,yao2026remotereasoner,wu2026rsreason}.

However, without a clear criterion for implicit queries, existing datasets~\citep{li2025segearthr1,xin2026segearthr2,ni2026unigeoseg,quenum2026lisat} often operationalize reasoning as category omission (verified in Table~\ref{tab:dataset_audit}), replacing a canonical category with a coarser hypernym (e.g., ``motorway ramp'' to ``feature'').
This granularity shift broadens the candidate space~\citep{gualdoni2024objects,graf2016animal} without guaranteeing reasoning (Figure~\ref{fig:definition}, IP): ``the narrow diagonal feature extending from top-left to bottom-center'' omits the category yet is resolvable from shape and location alone. 
Conversely, explicit-category queries that still require reasoning are largely overlooked (Figure~\ref{fig:definition}, EI): ``the motorway ramp serving as an entrance to the main road'' explicitly names the category yet requires reasoning about traffic flow and merging geometry to disambiguate competing roadways.
Moreover, existing benchmarks frequently mix perceptual and inferential attributes in uncontrolled queries, introducing perceptual shortcuts that entangle category recognition, visual perception, and geospatial inference. 
This leaves benchmark evaluation confounded and unauditable, raising a fundamental question: \emph{what do existing benchmarks actually measure, and how can we precisely disentangle and evaluate geospatial reasoning capability?}

To answer this question and make query demands explicitly auditable, we reformulate geospatial reasoning segmentation along two complementary dimensions: category expression and attribute nature (Figure~\ref{fig:definition}). 
Category expression characterizes whether the target category is explicitly named or implicitly generalized. Attribute nature characterizes whether the target attributes are perceptual, directly observable from the image (e.g., color, shape, and position), or inferential, requiring geospatial knowledge or contextual reasoning (e.g., land-use function, traffic semantics, and environmental risk).
This formulation yields four canonical query types: explicit-perceptual (EP), implicit-perceptual (IP), explicit-inferential (EI), and implicit-inferential (II).
Auditing 90,708 queries across existing benchmarks~\citep{li2025segearthr1,quenum2026lisat,xin2026segearthr2,ni2026unigeoseg} under this formulation quantifies the mismatch between nominal task definitions and genuine inferential demands: pure EI queries account for only 0.61\%, while existing splits either skew toward a single regime or mix perceptual and inferential attributes in 31.98\% of queries.

To address these limitations, we construct \textbf{RSReasoner}, a large-scale geospatial reasoning segmentation dataset of 106,610 images and masks with 213,220 queries, and \textbf{RSReasoner-Bench}, a diagnostic benchmark of 5,800 targets and 75,400 queries across four canonical types and controlled variants. Specifically, we ground data construction in OpenStreetMap (OSM)~\citep{haklay2008openstreetmap} from RSTeller~\citep{ge2025rsteller}, whose key--value tags supply rich geospatial attributes beyond category labels, and develop a three-stage pipeline to mitigate semantic~\citep{vargas2020openstreetmap,johnson2022opensentinelmap,biljecki2023quality} and geometric noise~\citep{kaiser2017learning,johnson2022opensentinelmap,vargas2019correcting} in raw OSM. Built upon our formulation, RSReasoner-Bench evaluates model capabilities across broader dimensions than existing benchmarks, precisely disentangling attribute nature, category expression, and same-category competition while exposing shortcut behaviors. Moreover, continual fine-tuning on RSReasoner substantially improves both in-domain reasoning and cross-type generalization. 
These findings offer key insights for future benchmark design and reasoning research.

In summary, our contributions are threefold:
\begin{itemize}
    \item We reformulate geospatial reasoning segmentation along category expression and attribute nature for fine-grained, auditable evaluation, and audit existing benchmarks to expose mismatches between nominal task definitions and genuine reasoning demands.
    \item We develop a three-stage OSM-guided denoising and annotation pipeline to construct RSReasoner for large-scale reasoning supervision and RSReasoner-Bench for controlled multi-dimensional evaluation across canonical types and diagnostic variants.
    \item Extensive experiments disentangle reasoning capability across attribute nature, category expression, and same-category competition, expose shortcut behaviors, and offer insights for reasoning research and benchmark design.

\end{itemize}

\section{Related Work}

\textbf{Reasoning Segmentation in Natural Images.}
Reasoning segmentation extends language-guided segmentation beyond explicit categories and straightforward referring expressions. 
LISA~\citep{lai2024lisa} formulates the task as predicting a binary mask from a complex and implicit query requiring reasoning or world knowledge. 
Subsequent studies expand this paradigm to instance-level and multi-target grounding~\citep{yang2023lisa++,ren2024pixellm,GSVA_Xia_2024_CVPR}, multiple granularities~\citep{jang2025mmr}, and unified task settings~\citep{zhang2024psalm,InstructSeg_Wei_2025_ICCV}, alongside grounded conversation and multi-round interaction~\citep{rasheed2024glamm,wang2025segllm}.
To align language reasoning with pixel-level localization, other methods explore mask-proposal selection~\citep{wang2024llmseg}, structured reasoning chains~\citep{bao2024cores}, token-similarity guidance~\citep{Qian_READ_2025_CVPR}, and reinforcement learning~\citep{liu2025segzero,liu2026visionreasoner,huang2026samr1,you2025seg}.
Meanwhile, prior studies introduce controlled diagnostic benchmarks for compositional visual reasoning~\citep{johnson2017clevr,hudson2019gqa} and referring comprehension~\citep{Chen_2020_CVPR,liu2024finecops}, while exposing shortcut reliance on category priors or superficial cues~\citep{cirik2018visual,RefAdvICLR2026}. 
However, beyond initial analyses of referring attributes~\citep{zhang2026hierarchical} and VQA biases~\citep{chappuis2025evaluating}, controlled diagnostic benchmarks and shortcut analyses for reasoning segmentation in remote sensing remain lacking.

\textbf{Language-Guided Geospatial Segmentation.}
Early language-guided geospatial segmentation focuses on referring expressions with explicit category and appearance descriptions~\citep{yuan2024rrsis,liu2024rrsis-d,dong2024cross,yao2025remotesam,ou2025geopix,yang2025large,zhou2024geoground}. 
Following the implicit-query paradigm in natural images, recent benchmarks expand geospatial reasoning across query descriptions, target multiplicity, and task coverage: \citet{li2025segearthr1} introduce EarthReason for implicit reasoning, LISAt~\citep{quenum2026lisat} curates GRES from detection annotations, \citet{xin2026segearthr2} broaden LaSeRS to multi-target and part-level granularities, and UniGeoSeg~\citep{ni2026unigeoseg} constructs GeoSeg-1M and GeoSeg-Bench with attribute and context reasoning. 
While these datasets have driven various reasoning segmentation models~\citep{zhang2026bridging,jiang2026geoseg,wu2026rsreason,yao2026remotereasoner}, they suffer from skewed query distributions and attribute mixing, failing to precisely train and evaluate reasoning capability. 
Moreover, they are built primarily from category-level labels without rich geospatial attributes to ground diverse, verifiable reasoning queries.

\textbf{OSM-Grounded Geospatial Data Construction.}
Addressing this limitation requires data sources coupling spatial delineations with target-specific geospatial semantics. OpenStreetMap (OSM)~\citep{haklay2008openstreetmap} provides global vector geometries and key--value tags that have supported building, road, and multi-task annotation construction~\citep{kaiser2017learning,bastani2018roadtracer,bastani2023satlaspretrain,hu2024gbss}, land-use and land-cover mapping~\citep{schultz2017open,johnson2022opensentinelmap,schultz2025osmlanduse}, and vision-language supervision~\citep{wang2024skyscript,muhtar2024lhrs,ge2025rsteller}. 
However, raw OSM data may suffer from incomplete coverage, inconsistent semantics~\citep{vargas2020openstreetmap,zhou2022exploring,biljecki2023quality}, and geometric misalignment~\citep{vargas2019correcting,fobi2020learning}. 
Existing pipelines address these issues through semantic filtering and tag disambiguation~\citep{wang2024skyscript,ge2025rsteller,schultz2025osmlanduse}, geometric correction and robust learning~\citep{vargas2019correcting,fobi2020learning,li2025dragosm}, or pseudo-label generation~\citep{henry2024worldwide} with SAM~\citep{kirillov2023segment}, but mainly for categorical mapping, vision-language annotation, or conventional segmentation.
Jointly harnessing OSM geometries and instance-level tags to construct verified masks and queries for reasoning segmentation remains underexplored.

\section{Two-Dimensional Query Formulation}
\label{sec:revisiting}

Given a remote sensing image \(I \in \mathbb{R}^{H \times W \times C}\) and a natural-language query \(q\), language-guided segmentation aims to predict a binary mask
\begin{equation}
    \hat{M} = f_{\theta}(I,q), \qquad
    \hat{M} \in \{0,1\}^{H \times W},
\end{equation}
where \(f_{\theta}\) is the segmentation model and \(\hat{M}\) delineates the geographic region referred to by \(q\). 
The query \(q\) specifies the target through both a categorical reference and a set of target-defining attributes. 
Accordingly, we characterize its reasoning demand along two dimensions: \emph{category expression} and \emph{attribute nature}.

\textbf{Category expression} characterizes whether the target category is explicitly named or implicitly generalized. Relative to the target's canonical category, a query is \emph{explicit} when it directly names that category or uses an unambiguous lexical equivalent, and \emph{implicit} otherwise. Implicit expressions do not eliminate categorical information, but shift the nominal reference to a coarser hypernym. For example, an ``orange'' may be referred to as a ``fruit,'' ``food,'' or ``object,'' while a ``building'' in remote sensing may be described as a ``structure,'' ``feature,'' or ``site.'' Such coarser references broaden the candidate space without necessarily introducing a reasoning requirement.

\textbf{Attribute nature} characterizes whether the target-defining attributes rely on direct visual perception or require inference. Specifically, \emph{perceptual} attributes capture properties directly observable from the image, including color, shape, texture, size, position, and spatial relation. By contrast, \emph{inferential} attributes cannot be established from visual appearance alone and require semantic interpretation of spatial context or geospatial knowledge, such as land-use function, infrastructure role, access or traffic semantics, hydrological function, environmental risk, and contextual role.

\textbf{Query taxonomy.}
Since category expression determines the linguistic granularity of target reference whereas attribute nature determines the evidence required for target selection, combining these two dimensions yields a taxonomy of pure, single-target segmentation queries, as illustrated in Figure~\ref{fig:definition}. For a query \(q\), let \(c(q)\in\{\mathrm{explicit},\mathrm{implicit}\}\) denote whether the target category is directly named, and let \(a(q)\in\{\mathrm{perceptual},\mathrm{inferential}\}\) denote the nature of the target-defining attributes. Its query type is defined as
\begin{equation}
    \tau(q)=\bigl(c(q),a(q)\bigr).
\end{equation}

This formulation yields four canonical query types: explicit-perceptual (EP), implicit-perceptual (IP), explicit-inferential (EI), and implicit-inferential (II). EP and IP queries are defined by perceptual attributes and remain resolvable through direct visual evidence, whereas EI and II queries require geospatial inference and differ only in whether the target category is directly named. 
In unconstrained queries, the two dimensions have different exclusivity properties. For a fixed target, explicit and implicit category expressions are mutually exclusive, whereas perceptual and inferential attributes may co-occur, yielding a compound \emph{mixed} query. A query may also refer to multiple target regions~\citep{liu2023gres}. We therefore use EP, IP, EI, and II as the canonical pure settings for single-target localization, while analyzing mixed queries separately when auditing existing benchmarks and quantifying shortcut effects.

\section{RSReasoner Dataset}
\label{sec:dataset}

\begin{figure*}[!htbp]
    \centering
    \includegraphics[width=\textwidth]{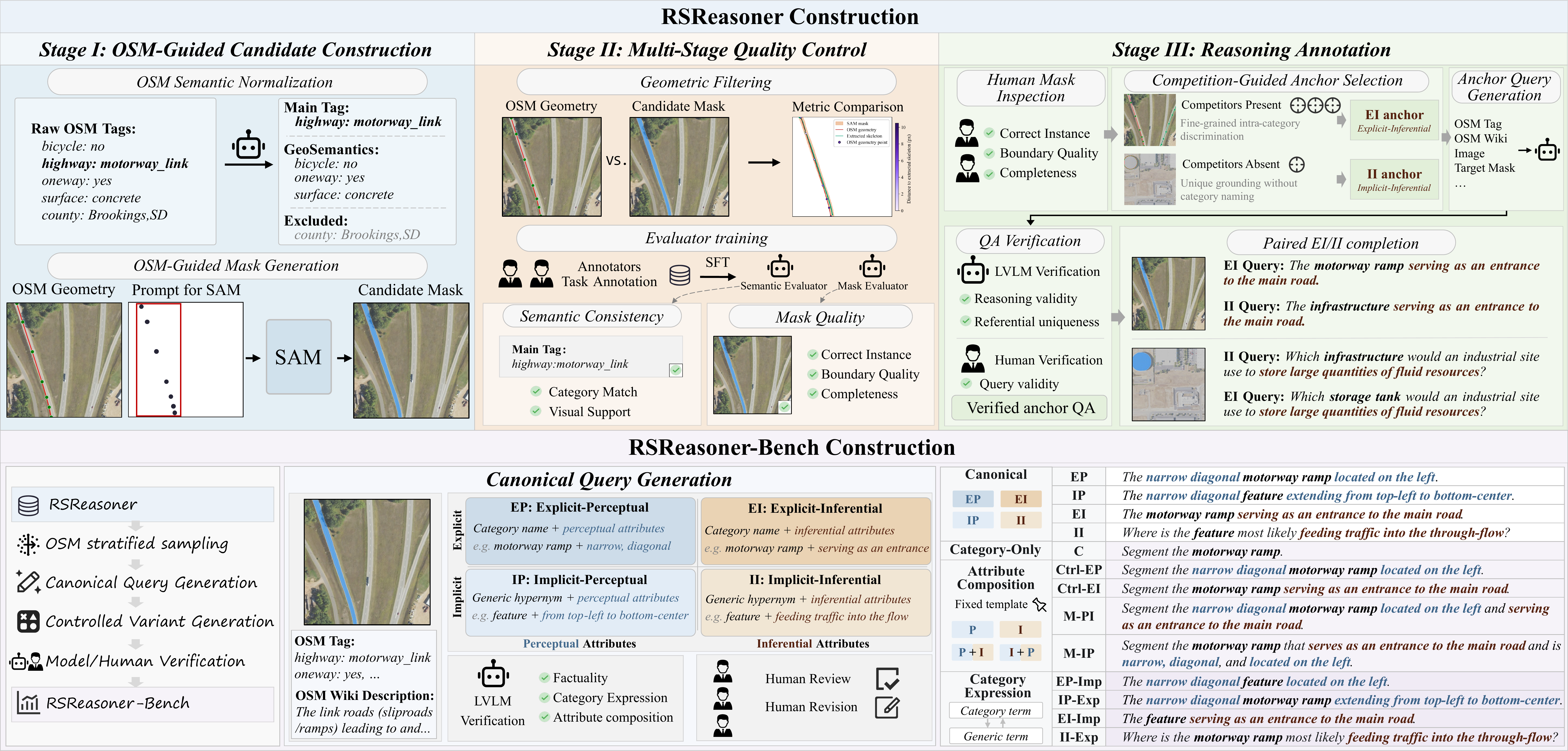}
    \caption{Overview of the construction pipeline for RSReasoner and RSReasoner-Bench, illustrating the three-stage OSM-guided pipeline for RSReasoner (top) and the construction of canonical and controlled diagnostic queries for RSReasoner-Bench (bottom).}
    \label{fig:pipeline}
\end{figure*}

We construct RSReasoner and RSReasoner-Bench from RSTeller~\citep{ge2025rsteller}, which provides 1,309,926 NAIP aerial images (\(448\times448\), \(0.6\,\mathrm{m}\) resolution) retrieved through Google Earth Engine~\citep{gorelick2017google} and their associated OpenStreetMap (OSM)~\citep{haklay2008openstreetmap} vector geometries and key--value tags. While the geometries provide approximate spatial priors and the tags encode target categories and fine-grained geosemantics, raw OSM annotations exhibit geometric misalignment and semantic noise.

\subsection{RSReasoner Dataset Construction}
To mitigate the geometric and semantic noise in raw OSM, we design a three-stage pipeline to construct RSReasoner, as illustrated in the top panel of Figure~\ref{fig:pipeline}.

\textbf{Stage I: OSM-Guided Candidate Construction.}
Because raw OSM elements contain heterogeneous key--value tags mixing categories, functions, and metadata, we first employ Qwen3.5-27B~\citep{qwen3.5} to select a canonical \emph{main tag} from each tag set that best represents the target's geographic category. The main tag serves as a stable anchor for category statistics, semantic verification, and same-category instance detection, while the remaining tags provide fine-grained geospatial attributes for reasoning annotation. 
For mask generation, since raw OSM geometries may be displaced or simplified, we treat them only as spatial priors and prompt SAM2~\citep{ravi2025sam} with geometry-specific point and box prompts for area and non-area objects following Seg-Zero~\citep{liu2025segzero}.

\textbf{Stage II: Multi-Stage Quality Control.}
Since noisy OSM priors do not guarantee accurate instance delineation or semantic consistency, we combine calibrated geometric filtering with decoupled multimodal assessment. To prevent frequent categories from dominating threshold calibration, we manually inspect 5,000 square-root-weighted candidates across 267 main tags, identifying 937 high-quality references. We then calibrate separate quantile thresholds for area and non-area objects across overlap, alignment, orientation, and fragmentation metrics, retaining 303,023 geometrically plausible candidates.

To verify both boundary quality and visual support for the main tag, we then decouple the remaining verification into binary \emph{mask-quality} and \emph{semantic-consistency} assessments. 
Following \citet{ni2026unigeoseg}, both evaluators receive the image with the candidate boundary overlaid, while the semantic evaluator additionally receives the main tag and its OSM Wiki description following RSTeller~\citep{ge2025rsteller}. We manually annotate and split 5,000 geometrically filtered candidates for both tasks to fine-tune dedicated evaluators, selecting Qwen3.6-27B~\citep{qwen3.6-27b} for mask quality and Qwen3.5-27B~\citep{qwen3.5} for semantic consistency. After retraining on all 5,000 annotations, the two filters accept 67.38\% and 70.98\% of the geometrically screened pool, with a joint intersection of 165,023 candidates.

\textbf{Stage III: Taxonomy-Guided Reasoning Annotation.}
In Stage~III, we first identify visible same-category competitors by combining OSM main-tag matching with visual inspection via Qwen3.5-122B-A10B~\citep{qwen3.5}, and condition the initial anchor query type on same-category competition to ensure genuinely discriminative inferential attributes. 
When no same-category competitor is present, naming the canonical category would allow the category alone to specify the target without testing whether the inferential attribute is discriminative~\citep{cirik2018visual}. We therefore construct an implicit-inferential (II) anchor without the canonical category so that the inferential attribute itself must uniquely identify the target. Conversely, since the category alone is insufficient when same-category competitors are present, we construct an explicit-inferential (EI) anchor to force the annotation to capture fine-grained intra-category inferential distinctions rather than generic category-level descriptions.

Before query generation, all 165,023 candidate masks undergo manual inspection for target correctness and boundary quality, retaining 147,472 masks. 
Using the image, full OSM semantics, and OSM Wiki description, we employ Qwen3.7-Plus~\citep{qwen37plus} to generate each anchor query--answer pair with minimal sufficient reasoning cues while excluding redundant perceptual cues to prevent localization shortcuts~\citep{Wu_2026_CVPR,RefAdvICLR2026}. 
Next, Qwen3.7-Plus verifies the reasoning validity and referential uniqueness of the generated anchors to retain 106,610 final targets, where human auditing on 6,523 sampled queries confirms a 95.48\% acceptance rate.
Finally, while keeping the verified inferential clause fixed, we apply controlled lexical transformation with human verification to switch each anchor's nominal reference between explicit and implicit forms, completing paired EI and II queries for every target to decouple category expression from competition conditions.

\subsection{RSReasoner-Bench Construction}

To systematically disentangle the effects of category expression, attribute nature, and same-category competition on the same visual targets, we construct RSReasoner-Bench from RSReasoner for controlled multi-dimensional evaluation, as shown in the bottom panel of Figure~\ref{fig:pipeline}. Specifically, we sample 5,800 targets across OSM categories using category-stratified sampling to mitigate long-tail dominance while maintaining a balanced composition of area and non-area geometries as well as targets with and without same-category competitors. With the image and ground-truth mask fixed, each target is paired with four independently generated queries covering all canonical types, namely EP, IP, EI, and II (23,200 queries in total), which are screened by GPT-6-Astra~\citep{openai2026gpt6} for factuality, category expression, and attribute purity, followed by human review and revision of 2,371 queries. 
Building on these verified benchmark queries, we further construct a category-only reference (C), category-replacement variants (EP-Imp, IP-Exp, EI-Imp, and II-Exp) that switch nominal references between explicit and implicit forms, and template-aligned single-attribute (Ctrl-EP and Ctrl-EI) as well as mixed-attribute (perceptual-first M-PI and inferential-first M-IP) queries via Qwen3.7-Plus for subsequent controlled diagnostic analysis. 
Further pipeline details, benchmark construction rules, quality-control statistics, and prompt templates are provided in the supplementary material.

\subsection{Dataset Statistics}
\label{sec:dataset_statistics}

\begin{figure}[!htbp]
    \centering
    \includegraphics[width=\columnwidth]{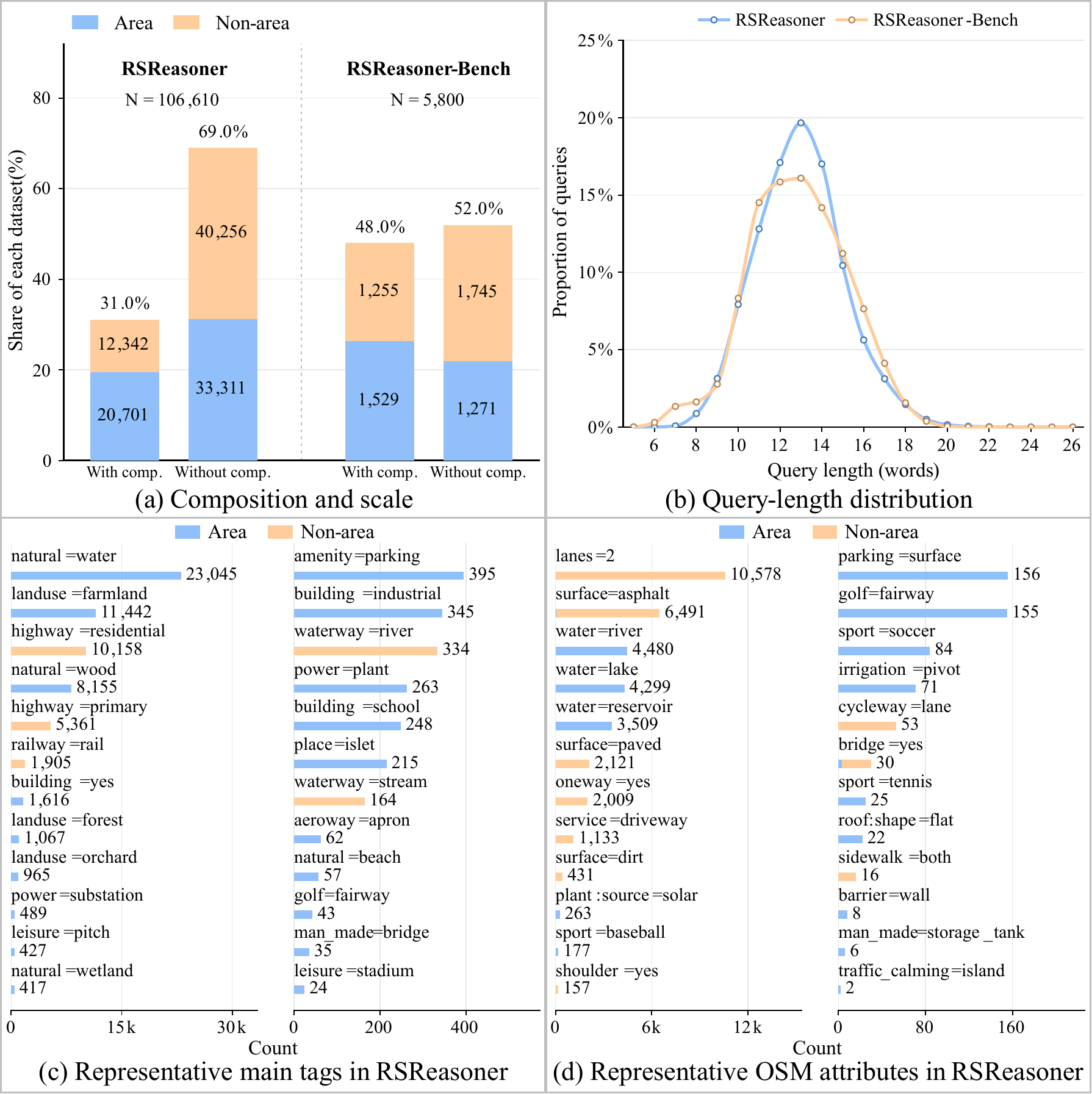}
\caption{Dataset statistics and representative OSM tags.}
    \label{fig:main_semantic_distribution}
\end{figure}

RSReasoner contains 106,610 quality-controlled geospatial reasoning segmentation targets, each comprising a \(448\times448\) image, a binary mask, paired EI and II queries (213,220 queries in total), and associated OSM semantics. As shown in Figure~\ref{fig:main_semantic_distribution}(a,b), the targets originate from 33,043 EI anchors with same-category competitors and 73,567 II anchors without competitors, exhibiting nearly balanced area and non-area geometries alongside a compact query-length distribution. For OSM semantics, the normalized main tag provides the target category anchor while the full tag set retains instance-level attributes, covering 228 main tags from 36 main keys and 44,564 unique key--value pairs across 913 full-tag keys to span diverse geographic categories and fine-grained inferential attributes (Figure~\ref{fig:main_semantic_distribution}(c,d)).

RSReasoner-Bench comprises 5,800 images and masks with balanced geometries and competition conditions, preserving 215 main tags across 35 main keys. It provides 23,200 canonical queries (EP, IP, EI, and II) together with 52,200 controlled diagnostic queries (75,400 queries in total across all 13 conditions), enabling comprehensive evaluation across category expression, attribute nature, competition conditions, and target geometries. 
We use RSReasoner-Bench as the held-out test set and partition the remaining RSReasoner targets into 95,150 training and 5,660 validation targets, assigning samples that share source imagery or OSM element identities to the same split to prevent spatial and instance leakage.

\begin{table}[!htbp]
    \centering
    \footnotesize
    \setlength{\tabcolsep}{1.8pt}
    \renewcommand{\arraystretch}{1.03}
    \caption{Comparison of language-guided geospatial segmentation datasets. Cat.\ and GeoSem.\ denote category count and structured target-specific geographic metadata.}
    \label{tab:dataset_comparison}
    \begin{tabular*}{\columnwidth}{@{\extracolsep{\fill}}lrrrrc@{}}
        \toprule
        Dataset & Images & Masks & Queries & Cat. & GeoSem. \\
        \midrule
        RefSegRS~\citep{yuan2024rrsis} & 285 & 4,420 & 4,420 & 14 & No \\
        RRSIS-D~\citep{liu2024rrsis-d} & 17,402 & 17,402 & 17,402 & 20 & No \\
        RISBench~\citep{dong2024cross} & 29,614 & 52,472 & 52,472 & 26 & No \\
        EarthReason~\citep{li2025segearthr1} & 5,434 & 5,434 & 32,604 & 28 & No \\
        GRES~\citep{quenum2026lisat} & 9,205 & 27,615 & 27,615 & 35 & No \\
        LaSeRS~\citep{xin2026segearthr2} & 10,329 & 40,396 & 30,830 & 122 & No \\
        GeoSeg-1M~\citep{ni2026unigeoseg} & 590,413 & 1,148,504 & 1,148,504 & 117 & No \\
        RSReasoner & 106,610 & 106,610 & 213,220 & 228$^{*}$ & Yes \\
        RSReasoner-Bench & 5,800 & 5,800 & 75,400 & 215$^{*}$ & Yes \\
        \bottomrule
    \end{tabular*}
    \parbox{\columnwidth}{\footnotesize
    \textit{Note:} $^{*}$Normalized OSM main tags rather than a closed set.
    }
\end{table}

\subsection{Comparison with Existing Benchmarks}

Table~\ref{tab:dataset_comparison} compares RSReasoner and RSReasoner-Bench with existing language-guided geospatial segmentation datasets. Most existing datasets repurpose classification, detection, segmentation, or grounding resources, providing supervision primarily through category labels and spatial annotations. In contrast, our dataset grounds each target in OSM geometry and key--value tags, supplying structured geospatial semantics to support both large-scale reasoning supervision in RSReasoner and controlled multi-dimensional diagnostics in RSReasoner-Bench.

To examine what existing benchmarks actually measure, we employ GPT-6-Astra~\citep{openai2026gpt6} with human verification to audit all 90,708 language queries under our definition across EarthReason~\citep{li2025segearthr1}, GRES~\citep{quenum2026lisat}, LaSeRS~\citep{xin2026segearthr2}, and GeoSeg-Bench~\citep{ni2026unigeoseg} (excluding interactive point/box prompts), alongside the 23,200 canonical queries in RSReasoner-Bench. Aside from queries without target attributes, all queries are categorized by category expression (explicit/implicit) and attribute composition (perceptual-only/inferential-only/mixed) into six classes: EP, EI, Explicit-Mixed (EM), IP, II, and Implicit-Mixed (IM).
Detailed audit protocols and verification results are provided in the supplementary material.

\begin{table}[!tbp]
    \centering
    \footnotesize
    \setlength{\tabcolsep}{1.7pt}
    \renewcommand{\arraystretch}{1.00}
    \caption{Query taxonomy distribution (\%) across existing benchmarks and subsets (\(N\): query count; EM/IM: explicit-/implicit-mixed). Suffixes -rea/-ref denote reasoning/referring splits, and -attr/-con denote attribute/context reasoning in GeoSeg-Bench.}
    \label{tab:dataset_audit}
    \begin{tabular*}{\columnwidth}{@{\extracolsep{\fill}}lr|rrr|rrr@{}}
        \toprule
        \multirow{2}{*}{Dataset} & \multirow{2}{*}{\(N\)} & \multicolumn{3}{c|}{Explicit (\%)} & \multicolumn{3}{c}{Implicit (\%)} \\
        \cmidrule{3-8}
        & & EP & EI & EM & IP & II & IM \\
        \midrule
        EarthReason-rea & 32,595 & 0.01 & 1.25 & 0.48 & 0.26 & 84.88 & 13.12 \\
        GRES-rea & 23,623 & 71.57 & 0.07 & 4.59 & 19.38 & 0.03 & 4.11 \\
        \midrule
        LaSeRS & 30,353 & 4.66 & 0.40 & 24.62 & 1.73 & 21.64 & 46.92 \\
        LaSeRS-ref & 8,729 & 16.08 & 0.63 & 82.80 & 0.01 & 0.01 & 0.34 \\
        LaSeRS-rea & 21,624 & 0.05 & 0.31 & 1.13 & 2.43 & 30.37 & 65.72 \\
        \midrule
        GeoSeg-Bench & 4,137 & 38.02 & 0.15 & 1.67 & 15.83 & 26.44 & 17.89 \\
        GeoSeg-Bench-ref & 2,354 & 66.78 & 0.00 & 2.17 & 26.72 & 0.00 & 4.33 \\
        GeoSeg-Bench-attr & 1,150 & 0.00 & 0.52 & 0.00 & 0.00 & 93.74 & 5.74 \\
        GeoSeg-Bench-con & 633 & 0.16 & 0.00 & 2.84 & 4.11 & 2.53 & 90.36 \\

        \midrule
        RSReasoner-Bench & 23,200 & 25.20 & 24.51 & 0.30 & 25.05 & 24.67 & 0.27 \\
        RSReasoner-Bench-EP & 5,800 & 99.84 & 0.00 & 0.14 & 0.02 & 0.00 & 0.00 \\
        RSReasoner-Bench-IP & 5,800 & 0.00 & 0.02 & 0.00 & 99.98 & 0.00 & 0.00 \\
        RSReasoner-Bench-EI & 5,800 & 0.91 & 97.95 & 1.07 & 0.00 & 0.07 & 0.00 \\
        RSReasoner-Bench-II & 5,800 & 0.03 & 0.07 & 0.00 & 0.21 & 98.60 & 1.09 \\

        \bottomrule
    \end{tabular*}
\end{table}

As shown in Table~\ref{tab:dataset_audit}, our audit reveals mismatches between conventional task definitions and genuine inferential demands. The reasoning splits of EarthReason, LaSeRS, and GeoSeg-Bench are 97.00\%--99.48\% category-implicit, confirming that prior benchmarks operationalize reasoning primarily as category omission. 
GRES is predominantly category-explicit (76.23\%), and although framed as reasoning segmentation, it consists largely of perceptual queries, with only 8.80\% containing inferential attributes. 
In LaSeRS, mixed queries dominate both its referring and reasoning splits (71.54\% overall) by heavily stacking perceptual and inferential attributes. 
GeoSeg-Bench generates implicit attribute-reasoning queries for non-competing targets, while assigning competing targets to context reasoning that omits categories yet specifies instances via contextual cues (typically spatial relations), rendering 93.20\% of GeoSeg-Bench-con mixed.
Overall, across all 90,708 audited queries, pure explicit-inferential queries account for only 0.61\% and leave a key regime largely unaddressed, whereas 31.98\% mix perceptual and inferential attributes, introducing localization shortcuts and obscuring which evidence governs target selection. 
In contrast, RSReasoner-Bench achieves balanced coverage across the four canonical types with high purity (0.57\% mixed), supporting controlled variant construction and diagnostic experiments.

\section{Experiments}

We conduct zero-shot and diagnostic evaluations on RSReasoner-Bench, comprising 5,800 test targets partitioned into 2,784 with same-category competitors and 3,016 without.
We evaluate nine zero-shot configurations across seven model families: LISA-7B and LISA-13B~\citep{lai2024lisa}, PixelLM-7B~\citep{ren2024pixellm}, LISAt-7B~\citep{quenum2026lisat}, GeoPixel-7B-RES~\citep{shabbir2025geopixel}, SegEarth-R1~\citep{li2025segearthr1}, UniGeoSeg~\citep{ni2026unigeoseg}, and Think2Seg-RS-3B and Think2Seg-RS-7B~\citep{zhang2026bridging}. In addition, we fine-tune representative models on RSReasoner to assess in-domain reasoning gains and cross-type transfer on RSReasoner-Bench. Following standard protocols~\citep{kazemzadeh2014referitgame,yu2016modeling,liu2023gres,lai2024lisa}, we report generalized IoU (gIoU) and cumulative IoU (cIoU).

\subsection{Zero-Shot Evaluation across Query Types}

\begin{table}[!htbp]
    \centering
    \footnotesize
    \setlength{\tabcolsep}{2pt}
    \renewcommand{\arraystretch}{1.00}
    \caption{Zero-shot segmentation performance (gIoU/cIoU, \%) across the four canonical query types on RSReasoner-Bench.}
    \label{tab:zeroshot_quadrants}
    \begin{tabularx}{\columnwidth}{l*{4}{>{\centering\arraybackslash}X}}
        \toprule
        Method & EP & IP & EI & II \\
        \midrule
        \multicolumn{5}{l}{\textit{Overall (5,800 targets)}} \\
        LISA-7B & 57.21/65.37 & 51.35/57.74 & 49.28/47.13 & 42.99/37.38 \\
        LISA-13B & 61.32/64.01 & 46.20/54.53 & 52.15/50.37 & 45.35/42.53 \\
        PixelLM-7B & 49.90/62.54 & 40.69/52.83 & 26.54/30.55 & 19.59/20.88 \\
        LISAt-7B & 49.42/62.91 & 39.35/50.49 & 45.51/49.80 & 38.97/40.69 \\
        GeoPixel-7B-RES & 58.68/53.34 & 55.81/48.13 & 47.51/37.13 & 39.61/30.76 \\
        SegEarth-R1 & 31.37/20.15 & 33.26/20.80 & 27.38/19.98 & 32.93/20.63 \\
        UniGeoSeg & 60.18/48.97 & 48.65/27.07 & 53.25/45.65 & 42.85/33.30 \\
        Think2Seg-RS-3B & 69.45/68.57 & 67.79/66.85 & 61.13/55.66 & 55.16/46.36 \\
        Think2Seg-RS-7B & \textbf{75.84}/\textbf{75.27} & \textbf{74.05}/\textbf{73.76} & \textbf{66.19}/\textbf{59.44} & \textbf{59.68}/\textbf{51.15} \\
        \rowcolor{gray!15} Avg. & 57.04/57.90 & 50.79/50.24 & 47.66/43.97 & 41.90/35.96 \\
        \midrule
        \multicolumn{5}{l}{\textit{With same-category competitors (2,784 targets)}} \\
        LISA-7B & 48.29/55.95 & 45.62/55.77 & 35.84/35.63 & 30.49/29.08 \\
        LISA-13B & 51.55/53.93 & 43.07/52.41 & 36.90/37.30 & 31.10/32.05 \\
        PixelLM-7B & 43.93/55.70 & 36.65/49.85 & 18.28/22.46 & 12.45/15.39 \\
        LISAt-7B & 43.75/56.69 & 36.05/50.97 & 36.37/41.01 & 28.64/32.12 \\
        GeoPixel-7B-RES & 52.56/50.06 & 50.43/48.54 & 37.33/30.60 & 30.54/25.23 \\
        SegEarth-R1 & 23.71/16.84 & 24.81/17.19 & 21.68/16.55 & 24.30/17.45 \\
        UniGeoSeg & 48.62/39.53 & 36.66/22.05 & 38.52/34.83 & 24.75/18.07 \\
        Think2Seg-RS-3B & 57.34/55.88 & 57.08/57.44 & 45.32/40.63 & 36.74/30.12 \\
        Think2Seg-RS-7B & \textbf{65.72}/\textbf{64.04} & \textbf{64.77}/\textbf{65.76} & \textbf{50.17}/\textbf{44.88} & \textbf{41.73}/\textbf{36.67} \\
        \rowcolor{gray!15} Avg. & 48.39/49.85 & 43.90/46.66 & 35.60/33.77 & 28.97/26.24 \\
        \midrule
        \multicolumn{5}{l}{\textit{Without same-category competitors (3,016 targets)}} \\
        LISA-7B & 65.44/74.80 & 56.64/59.46 & 61.69/60.43 & 54.53/45.52 \\
        LISA-13B & 70.35/74.65 & 49.10/56.39 & 66.23/66.48 & 58.51/54.04 \\
        PixelLM-7B & 55.41/69.09 & 44.42/55.43 & 34.16/38.04 & 26.18/25.97 \\
        LISAt-7B & 54.65/68.48 & 42.39/50.09 & 53.96/58.80 & 48.50/49.40 \\
        GeoPixel-7B-RES & 64.32/56.32 & 60.77/47.76 & 56.91/43.41 & 47.98/35.91 \\
        SegEarth-R1 & 38.44/23.11 & 41.06/23.91 & 32.65/23.17 & 40.89/23.47 \\
        UniGeoSeg & 70.84/58.18 & 59.72/31.48 & 66.85/57.29 & 59.56/48.23 \\
        Think2Seg-RS-3B & 80.62/81.38 & 77.67/75.72 & 75.73/72.55 & 72.17/63.64 \\
        Think2Seg-RS-7B & \textbf{85.17}/\textbf{87.06} & \textbf{82.62}/\textbf{81.47} & \textbf{80.97}/\textbf{77.49} & \textbf{76.25}/\textbf{67.66} \\
        \rowcolor{gray!15} Avg. & 65.03/65.90 & 57.15/53.52 & 58.79/55.30 & 53.84/45.98 \\
        \bottomrule
    \end{tabularx}
\end{table}

To investigate how category expression, attribute nature, and same-category competition govern localization difficulty, we evaluate all nine models across the four canonical query types. Table~\ref{tab:zeroshot_quadrants} reveals three consistent patterns across the two query dimensions. 
First, with category expression fixed, inferential queries are substantially harder than perceptual ones across nearly all models. Second, with attribute nature fixed, implicit queries are harder than explicit ones as replacing canonical categories with coarser hypernyms broadens the candidate space. Nearly all models follow this trend, while SegEarth-R1 favors implicit queries likely due to the implicit bias learned from its EarthReason training set. Third, the performance drop across the attribute nature dimension exceeds that across the category expression dimension. On the overall benchmark, switching from perceptual to inferential queries reduces average gIoU by 9.38\% from EP to EI and 8.89\% from IP to II, whereas switching from explicit to implicit expressions reduces average gIoU by 6.25\% from EP to IP and 5.76\% from EI to II, with an even wider gap under same-category competition. 
This confirms that attribute nature has a greater impact on localization difficulty than category expression.

Comparing the two scene subsets further shows that targets with same-category competitors are consistently harder than those without across all query types, with average gIoU dropping by 13.25\%--24.87\%. This gap indicates that same-category competitors are inherently harder to distinguish than heterogeneous objects. Without same-category competitors, a target can often be separated from other scene elements through broad functional or simple positional cues. In contrast, when same-category instances coexist, localization requires identifying target-specific perceptual or inferential attributes to resolve intra-category ambiguity, making explicit-inferential on competing targets even harder than implicit-inferential on non-competing targets. 
Together, these findings show that category expression, attribute nature, and same-category competition govern performance through distinct mechanisms, validating our formulation over conventional referring and reasoning definitions and enabling RSReasoner-Bench to precisely evaluate multi-dimensional model capabilities.

\begin{figure}[!htbp]
    \centering
    \includegraphics[width=\columnwidth]{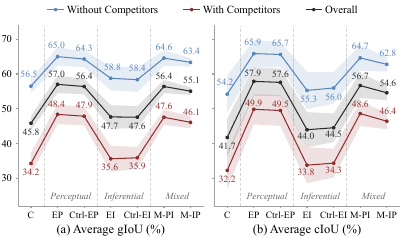}
    \caption{Nine-model average zero-shot performance in (a) gIoU (\%) and (b) cIoU (\%) across canonical explicit queries (EP and EI) and controlled attribute compositions (C, Ctrl-EP, Ctrl-EI, M-PI, and M-IP) on RSReasoner-Bench. Shaded bands denote the standard deviation across mean-centered model profiles.}
    \label{fig:attribute_composition}
\end{figure}

\subsection{Effect of Attribute Composition}
\label{sec:attribute_composition}

To disentangle the effect of attribute composition from syntactic variation, we construct controlled query variants sharing a unified sentence backbone. Specifically, we format the canonical category name into a concise template without attributes to form a category-only reference (C), rewrite the verified EP and EI queries under the same template to form single-attribute queries (Ctrl-EP and Ctrl-EI), and merge both attribute descriptions in either order to form perceptual-first (M-PI) and inferential-first (M-IP) mixed queries. 
As shown in Figure~\ref{fig:attribute_composition}, across all subsets in both gIoU and cIoU, EP and Ctrl-EP, as well as EI and Ctrl-EI, exhibit negligible differences while Ctrl-EP consistently outperforms Ctrl-EI, with small mean-centered standard deviations confirming consistent relative trends across all nine models (tighter in gIoU than in area-sensitive cIoU). This indicates that sentence structure has minimal impact on localization performance, and that the observed performance gaps genuinely arise from our two defined dimensions rather than surface phrasing. Per-model results are provided in the supplementary material.

With syntax fixed, comparing C against Ctrl-EP and Ctrl-EI disentangles the contributions of category recognition, perceptual grounding, and geospatial inference. On the overall benchmark, adding perceptual attributes in Ctrl-EP brings a substantial gain of 10.6\% gIoU and 15.9\% cIoU over C, whereas adding inferential attributes in Ctrl-EI yields only modest gains of 1.8\% and 2.8\%, showing that perceptual cues provide direct localization signals while inferential cues demand more complex semantic analysis. 
Moreover, attributes have a much stronger impact when same-category competitors coexist. While C achieves 56.5\% gIoU on non-competing targets where category recognition alone largely isolates the target, it drops sharply to 34.2\% on competing targets because the category name cannot distinguish multiple instances of the same class. Under such competition, Ctrl-EP improves gIoU by 13.7\% over C as perceptual cues effectively resolve intra-category ambiguity to reduce localization difficulty, whereas Ctrl-EI gains only 1.7\% as disambiguating competing instances through inferential attributes requires more demanding geospatial reasoning.

Comparing single-attribute queries (Ctrl-EP and Ctrl-EI) with mixed-attribute queries (M-PI and M-IP) further reveals a pronounced asymmetry that exposes perceptual shortcuts. Relative to purely inferential Ctrl-EI, incorporating perceptual cues in M-PI and M-IP raises overall gIoU from 47.6\% to 56.4\% and 55.1\%, restoring accuracy to the level of perceptual-only Ctrl-EP (56.4\%), whereas adding inferential descriptions to Ctrl-EP brings no gain. 
Moreover, perceptual-first M-PI yields a small but consistent gain of 1.2\%--1.5\% over inferential-first M-IP across subsets, suggesting that presenting perceptual cues first facilitates direct visual grounding.
These results indicate that including an inferential attribute in a query does not guarantee that target localization depends on reasoning. When perceptual and inferential cues coexist and both suffice to specify the target, models can exploit perceptual shortcuts to complete localization and bypass more demanding geospatial reasoning, empirically validating our principle of excluding redundant perceptual attributes during dataset construction.

\subsection{Effect of Category Expression}

\begin{table}[!t]
    \centering
    \footnotesize
    \setlength{\tabcolsep}{1.5pt}
    \renewcommand{\arraystretch}{1.00}
    \definecolor{gaincolor}{RGB}{33,102,172}  %
    \definecolor{dropcolor}{RGB}{178,24,43}   %
    \newcommand{\up}[1]{\textcolor{gaincolor}{#1}}
    \newcommand{\dn}[1]{\textcolor{dropcolor}{#1}}
    \caption{Performance changes (\(\Delta\text{gIoU}\,/\,\Delta\text{cIoU}\), \%; \textcolor{gaincolor}{blue}: gain, \textcolor{dropcolor}{red}: drop) under category-expression replacement on RSReasoner-Bench relative to Table~\ref{tab:zeroshot_quadrants}.}

    \label{tab:category_expression}
    \begin{tabularx}{\columnwidth}{l*{4}{>{\centering\arraybackslash}X}}
        \toprule
        Method & \(\Delta\)EP-Imp & \(\Delta\)IP-Exp & \(\Delta\)EI-Imp & \(\Delta\)II-Exp \\
        \midrule
        LISA-7B & \dn{0.51}\,/\,\dn{7.42} & \up{5.06}\,/\,\up{6.72} & \dn{3.04}\,/\,\dn{6.63} & \up{2.15}\,/\,\up{3.07} \\
        LISA-13B & \dn{4.65}\,/\,\dn{6.85} & \up{13.45}\,/\,\up{9.69} & \dn{5.26}\,/\,\dn{5.86} & \up{2.76}\,/\,\up{3.29} \\
        PixelLM-7B & \dn{8.65}\,/\,\dn{12.12} & \up{8.44}\,/\,\up{8.15} & \dn{4.18}\,/\,\dn{4.81} & \up{2.06}\,/\,\up{0.75} \\
        LISAt-7B & \dn{6.61}\,/\,\dn{7.00} & \up{7.06}\,/\,\up{7.15} & \dn{5.56}\,/\,\dn{9.86} & \up{3.12}\,/\,\up{3.30} \\
        GeoPixel-7B-RES & \dn{4.33}\,/\,\dn{4.91} & \up{3.40}\,/\,\up{5.64} & \dn{4.83}\,/\,\dn{4.03} & \up{4.69}\,/\,\up{1.51} \\
        SegEarth-R1 & \up{1.70}\,/\,\up{0.89} & \dn{0.85}\,/\,\dn{0.47} & \up{5.00}\,/\,\up{0.49} & \dn{5.22}\,/\,\dn{0.45} \\
        UniGeoSeg & \dn{11.43}\,/\,\dn{20.43} & \up{10.69}\,/\,\up{21.25} & \dn{9.75}\,/\,\dn{9.90} & \up{5.63}\,/\,\up{8.33} \\
        Think2Seg-RS-3B & \dn{4.28}\,/\,\dn{5.51} & \up{3.08}\,/\,\up{3.83} & \dn{4.85}\,/\,\dn{7.12} & \up{4.73}\,/\,\up{6.51} \\
        Think2Seg-RS-7B & \dn{4.09}\,/\,\dn{5.42} & \up{2.48}\,/\,\up{2.89} & \dn{6.36}\,/\,\dn{7.41} & \up{5.88}\,/\,\up{6.78} \\
        \rowcolor{gray!15} Avg. & \dn{4.76}\,/\,\dn{7.64} & \up{5.87}\,/\,\up{7.21} & \dn{4.31}\,/\,\dn{6.13} & \up{2.87}\,/\,\up{3.68} \\
        \bottomrule
    \end{tabularx}
\end{table}

To isolate the impact of category expression while keeping attribute evidence fixed, we evaluate the category-replacement variants (EP-Imp, IP-Exp, EI-Imp, and II-Exp) that only switch the target nominal reference between its canonical category and a coarser hypernym while preserving the remaining query text. As shown in Table~\ref{tab:category_expression}, nearly all models exhibit a symmetric pattern: replacing explicit categories with hypernyms (EP-Imp and EI-Imp) consistently degrades performance, whereas restoring explicit categories (IP-Exp and II-Exp) consistently improves performance. 
This demonstrates that category expression independently affects localization difficulty by constraining or broadening the candidate space.
Moreover, for inferential queries, the drop from implicitizing EI exceeds the gain from explicitizing II, especially on non-competing targets, as II already specifies targets without categories whereas implicitizing EI removes the category anchor and expands ambiguity (subset breakdowns in supplementary material).

\subsection{Fine-Tuning and Cross-Type Transfer}

\begin{table}[!t]
    \centering
    \footnotesize
    \setlength{\tabcolsep}{1.5pt}
    \renewcommand{\arraystretch}{1.00}
    \caption{Performance gains (\(\Delta\mathrm{gIoU}/\Delta\mathrm{cIoU}\), \%) on RSReasoner-Bench after fine-tuning on RSReasoner over Table~\ref{tab:zeroshot_quadrants}.}
    \label{tab:finetuning}
    \begin{tabularx}{\columnwidth}{l*{4}{>{\centering\arraybackslash}X}}
        \toprule
        Method & \(\Delta\)EP & \(\Delta\)IP & \(\Delta\)EI & \(\Delta\)II \\
        \midrule
        LISA-7B
        & 25.99\,/\,17.14 & 17.04\,/\,8.57
        & 33.18\,/\,32.20 & 36.95\,/\,38.56 \\

        PixelLM-7B
        & 32.37\,/\,24.29 & 36.37\,/\,24.50
        & 48.82\,/\,43.10 & 54.10\,/\,49.61 \\

        LISAt-7B
        & 34.43\,/\,20.29 & 31.23\,/\,10.30
        & 36.49\,/\,29.91 & 39.92\,/\,32.99 \\

        GeoPixel-7B-RES
        & 32.07\,/\,38.99 & 32.02\,/\,39.54
        & 37.84\,/\,46.65 & 42.56\,/\,46.81 \\

        SegEarth-R1
        & 45.99\,/\,53.44 & 15.47\,/\,23.26
        & 52.98\,/\,57.77 & 43.71\,/\,50.42 \\

        \rowcolor{gray!15} Avg. & 34.17\,/\,30.83 & 26.43\,/\,21.23 & 41.86\,/\,41.93 & 43.45\,/\,43.68 \\
        \bottomrule
    \end{tabularx}
\end{table}

To evaluate the training value of RSReasoner, we continually fine-tune representative models from their official checkpoints on RSReasoner using paired EI and II supervision. As shown in Table~\ref{tab:finetuning}, fine-tuning yields substantial in-domain gains across all models, boosting average gIoU/cIoU by 41.86\%/41.93\% on EI and 43.45\%/43.68\% on II while mitigating the implicit bias of SegEarth-R1. 
Moreover, although the training data contains only inferential queries (EI and II) without perceptual queries (EP and IP), fine-tuning brings pronounced cross-type gains of 34.17\%/30.83\% on EP and 26.43\%/21.23\% on IP, with larger gains on EP as explicit category anchors learned from EI transfer to constrain the candidate space. 
These results demonstrate that our instance-level OSM reasoning supervision not only enables models to leverage inferential attributes, but also enhances their generalization to perceptual grounding.
Training details are provided in the supplementary material.

\section{Discussion and Conclusion}

\textbf{Discussion for Future Research.}
Our study highlights several key insights for advancing geospatial reasoning segmentation: (1) reasoning queries should be characterized jointly by attribute nature and category expression, (2) the underexplored explicit-inferential regime warrants further study, (3) benchmark design should account for same-category competition, and (4) reasoning query construction should avoid attribute stacking to prevent perceptual shortcuts.
Finally, while our formulation establishes a clean diagnostic foundation, geospatial reasoning is inherently multifaceted. 
Future work can further subdivide inferential attributes into finer-grained tiers, such as functional roles, contextual relations, risk prediction, and multi-temporal or scenario planning.

\textbf{Conclusion.} We reformulate geospatial reasoning segmentation along category expression and attribute nature, audit 90,708 queries across existing benchmarks to expose scarce explicit-inferential cases and widespread attribute mixing, and develop a three-stage OSM-guided pipeline to construct RSReasoner and RSReasoner-Bench. Experiments show that RSReasoner-Bench precisely disentangles and evaluates model capabilities across attribute nature, category expression, and same-category competition while exposing perceptual shortcuts, and fine-tuning on RSReasoner substantially boosts both in-domain reasoning and cross-type perceptual generalization.

\section*{LLM Usage Statement}

In this work, ChatGPT and Codex were used to assist with manuscript language polishing, table and figure design, literature search, and project code development. As part of the research methodology and experiments, Qwen3.5-27B, Qwen3.6-27B, Qwen3.5-122B-A10B, and Qwen3.7-Plus were used for dataset construction and quality control, while GPT-6-Astra supported benchmark curation, verification, and query-taxonomy auditing. The authors reviewed and verified all AI-assisted text, tables, figures, and code, applied the reported quality-control and human-verification procedures to the generated annotations, and take full responsibility for all content and claims in this paper.

{
    \small
    \bibliographystyle{ieeenat_fullname}
    \bibliography{main}
}

\clearpage

\maketitlesupplementary

\section*{Overview}

This supplementary material provides additional technical details, dataset analyses, and complete experimental results supporting RSReasoner and RSReasoner-Bench, organized as follows:

\begin{itemize}
    \item Section~\ref{app:rsreasoner_construction}. RSReasoner Dataset Construction Details.
    \item Section~\ref{app:bench_verification}. RSReasoner-Bench Construction Details.
    \item Section~\ref{app:dataset_statistics}. Additional Dataset Statistics.
    \item Section~\ref{app:benchmark_audit}. Taxonomy Audit of Existing Benchmarks.
    \item Section~\ref{app:exp_results}. Additional Experimental Results.
    \item Section~\ref{app:qualitative_results}. Qualitative Examples.
    \item Section~\ref{app:prompt_templates}. Prompt Templates.
\end{itemize}

\section{RSReasoner Dataset Construction Details}
\label{app:rsreasoner_construction}

\subsection{OSM-Guided Candidate Construction}
\subsubsection{Main Tag Selection}

Main tag selection maps each heterogeneous OSM tag set to a stable category anchor for subsequent statistics, sampling, semantic verification, and same-category instance detection, while retaining the remaining tags as fine-grained metadata for reasoning annotation. 
For each tag set \(\mathcal{T}_i=\{(k_{ij},v_{ij})\}_{j=1}^{n_i}\), we employ Qwen3.5-27B~\citep{qwen3.5} to select a \emph{main tag} \(t_i^{\mathrm{main}}\in\mathcal{T}_i\), defined as the original key--value pair that best represents each element's canonical geographic category. The model prioritizes concrete, visually groundable category tags, excludes auxiliary attributes and metadata, and must return an exact pair from \(\mathcal{T}_i\).

The selection follows five principles. First, macro-category keys describing the underlying geographic entity, such as \texttt{landuse}, \texttt{natural}, \texttt{highway}, \texttt{waterway}, and \texttt{building}, are prioritized over auxiliary attributes. Second, physically grounded categories are preferred over more abstract functional descriptions when both characterize the same element. Third, the selected tag should denote a concrete geographic object or region that can plausibly correspond to an entity in overhead imagery. Fourth, modifiers and metadata, including addresses, identifiers, administrative fields, source information, and reference codes, are excluded. Finally, the model is required to return an exact key--value pair from the original tag set rather than generating a new label. The prompt template is shown in Figure~\ref{fig:prompt_main_tag_selection}.

Selecting a canonical main tag provides a consistent semantic representation without discarding the finer-grained information encoded in the full OSM tag set, which is retained as structured metadata. 
For example, an OSM element assigned \texttt{natural=water} as its main tag may also contain the key \texttt{water}, whose values specify more detailed water body types or properties. The values observed in RSReasoner include \texttt{basin}, \texttt{bay}, \texttt{borrow pit}, \texttt{canal}, \texttt{cooling\_channel}, \texttt{ditch}, \texttt{estuary}, \texttt{fishpond}, \texttt{intermittent}, \texttt{lagoon}, \texttt{lake}, \texttt{lake;pond}, \texttt{lock}, \texttt{no}, \texttt{oxbow}, \texttt{pond}, \texttt{pond/puddle}, \texttt{reservoir}, \texttt{river}, \texttt{river:canal}, \texttt{stream}, \texttt{stream:river}, and \texttt{wastewater}.

Elements containing a single tag are assigned that tag directly. For elements containing multiple tags, identical tag configurations are deduplicated before inference because they constitute the same semantic selection task. 
Among the 1,309,926 OSM elements, 300,712 contain a single tag, while 344,393 unique tag configurations require model-based selection after deduplication. The semantic normalization process consolidates the original tag space into 1,349 main tags across 120 main keys, which are subsequently reduced through the quality-control pipeline.

\subsubsection{OSM-Guided Mask Generation}

OSM vector geometries provide approximate target locations and extents but cannot serve directly as pixel-level supervision because they may be displaced, simplified, incomplete, or inconsistent with the aerial image. We therefore treat them only as spatial priors and use SAM2~\citep{ravi2025sam} to generate candidate masks.

Following the OSM geometry conventions in RSTeller~\citep{ge2025rsteller}, we categorize targets into \emph{area} objects (closed polygonal regions such as buildings, farmland, water body, and sports fields) and \emph{non-area} objects (elongated linear structures represented by polylines, such as roads, railways, and waterways), and construct type-specific prompts for SAM2.
For area objects, following Seg-Zero~\citep{liu2025segzero}, we rasterize the projected polygon and derive an expanded bounding box with two interior positive points, allowing SAM2 to refine the target boundary from the coarse polygon prior. For non-area objects, whose line geometries typically represent approximate centerlines rather than full target widths, we sample positive points along the projected line and use an expanded bounding box to guide segmentation of the corresponding elongated region. 

\subsection{Multi-Stage Quality Control}

\subsubsection{Distribution-Aware Sampling and Geometric Calibration}
\label{app:geometric_calibration}

Simple random sampling from the full candidate pool would overrepresent frequent main tags and provide insufficient coverage of rare geographic categories. We therefore exclude predefined invalid categories and main tags with fewer than ten instances, and assign each remaining category \(c\) the sampling probability
\begin{equation}
    p_c =
    \frac{\sqrt{n_c}}
    {\sum_{j\in\mathcal{C}}\sqrt{n_j}},
\end{equation}
where \(n_c\) is the number of candidates in category \(c\), and \(\mathcal{C}\) is the set of valid categories. The square-root transformation reduces category imbalance while preserving the relative frequency structure of the original distribution. Each category contributes between one and 200 samples. This procedure yields 5,000 candidates spanning 267 main tags, comprising 3,009 area objects and 1,991 non-area objects.

Human annotators inspect the sampled candidates and identify 937 high-quality reference masks, including 474 area objects and 463 non-area objects. These reference samples characterize the geometric relationship between acceptable SAM2 masks and their OSM priors. Because polygonal and linear objects differ substantially in geometric representation, we calibrate separate metric sets for the two types.

For area objects, geometric consistency is measured by the IoU between the rasterized OSM geometry and the SAM2 mask, the mean nearest distance from sampled OSM-boundary points to the mask boundary, the Euclidean distance between their centroids, and fragmentation, measured as the number of connected components. 

\begin{figure}[!t]
\centering
\includegraphics[width=\columnwidth]{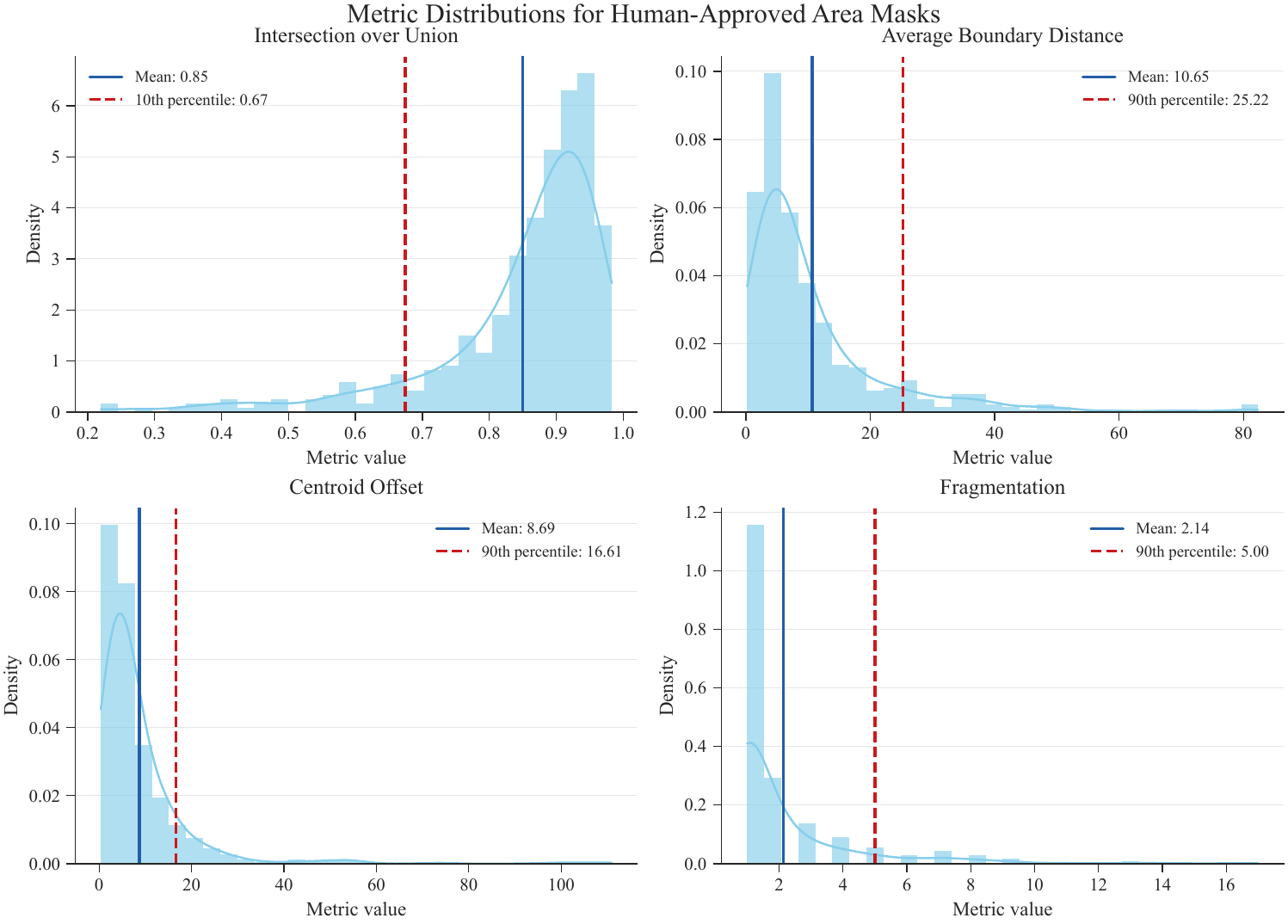}
\caption{Empirical distributions of the four quantile-calibrated geometric metrics over the 474 human-verified area masks.}
\label{fig:area_metric_distributions}
\end{figure}

\begin{figure}[!t]
\centering
\includegraphics[width=\columnwidth]{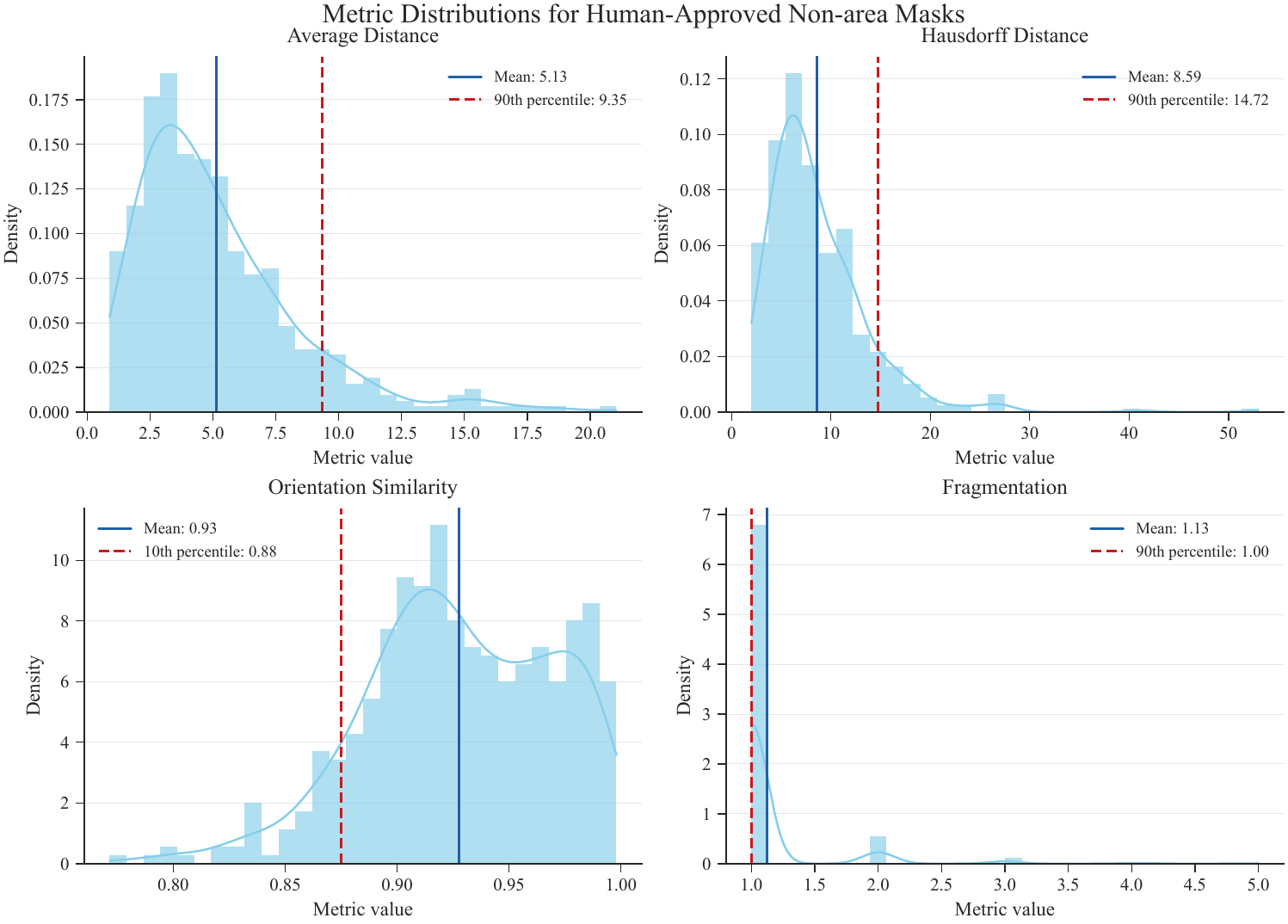}
\caption{Empirical distributions of the four quantile-calibrated geometric metrics over the 463 human-verified non-area masks.}
\label{fig:nonarea_metric_distributions}
\end{figure}

For non-area objects, region overlap is less informative because an OSM line generally represents the approximate centerline of an elongated target rather than its full spatial extent. We therefore skeletonize the SAM2 mask and compute the mean and maximum nearest-point distances from the OSM line to the extracted skeleton, with the latter corresponding to the one-way Hausdorff distance; orientation consistency measures their length-weighted local directional agreement, while fragmentation counts the connected skeleton components.

Figures~\ref{fig:area_metric_distributions} and~\ref{fig:nonarea_metric_distributions} show the empirical metric distributions of the human-verified reference masks. For metrics where higher values indicate stronger agreement, we use the 10th percentile as the lower acceptance bound. For metrics where lower values are preferable, we use the 90th percentile as the upper bound. A candidate is retained only if it satisfies all criteria for its geometric type. Table~\ref{tab:geometric_thresholds} reports the resulting thresholds.

Applying these criteria to the full pool retains 303,023 candidates, corresponding to \(23.13\%\) of the initial candidates. Table~\ref{tab:geometric_filtering_results} reports the results by geometric type.

\begin{table}[!htbp]
    \centering
    \small
    \setlength{\tabcolsep}{3.5pt}
    \caption{Quantile-calibrated geometric acceptance criteria for area and non-area targets. Distance and offset thresholds are reported in pixels on \(448\times448\) images.}
    \begin{tabular}{llc}
        \toprule
        Target type & Metric & Acceptance criterion \\
        \midrule
        Area
        & Intersection over union    & \(\geq 0.6744\) \\
        & Mean boundary distance     & \(\leq 25.22\) \\
        & Centroid offset            & \(\leq 16.61\) \\
        & Fragmentation              & \(\leq 5\) \\
        \midrule
        Non-area
        & Mean line-to-mask distance & \(\leq 9.35\) \\
        & One-way Hausdorff distance & \(\leq 14.72\) \\
        & Orientation consistency    & \(\geq 0.88\) \\
        & Fragmentation              & \(\leq 1\) \\
        \bottomrule
    \end{tabular}
    \label{tab:geometric_thresholds}
\end{table}

\begin{table}[!htbp]
    \centering
    \small
    \setlength{\tabcolsep}{5pt}
    \caption{Retention statistics after geometric coarse filtering.}
    \begin{tabular}{lrrr}
        \toprule
        Target type & Initial & Retained & Retention (\%) \\
        \midrule
        Area     & 593,099   & 125,631 & 21.18 \\
        Non-area & 716,827   & 177,392 & 24.75 \\
        \midrule
        Overall  & 1,309,926 & 303,023 & 23.13 \\
        \bottomrule
    \end{tabular}
    \label{tab:geometric_filtering_results}
\end{table}

\subsubsection{Decoupled Quality Assessment}

Geometric consistency alone does not fully determine annotation quality: a candidate mask may remain aligned with its OSM prior while delineating the wrong instance or exhibiting incomplete or contaminated coverage, whereas a visually accurate mask may still be associated with an unsupported main tag. We therefore decouple the remaining quality assessment into two complementary binary tasks: mask-quality assessment and semantic-consistency assessment.

\textbf{Mask-quality assessment.}
This task determines whether the candidate mask accurately and completely delineates the intended visible target, considering instance correctness, boundary accuracy, completeness, and contamination. Following \citet{ni2026unigeoseg}, the model receives two aerial images: one overlaid with the candidate mask boundary and the other with the projected OSM geometry boundary, together with the main tag as a semantic anchor, preserving the target appearance and spatial context while identifying the region under evaluation.

\textbf{Semantic-consistency assessment.}
This task determines whether the extracted main tag correctly describes the visible geographic entity. The model receives the image overlaid with the projected OSM geometry along with the normalized main tag and its OSM Wiki description, which is included following RSTeller~\citep{ge2025rsteller} to clarify specialized terminology.

\begin{figure}[!htbp]
    \centering
    \includegraphics[width=\columnwidth]{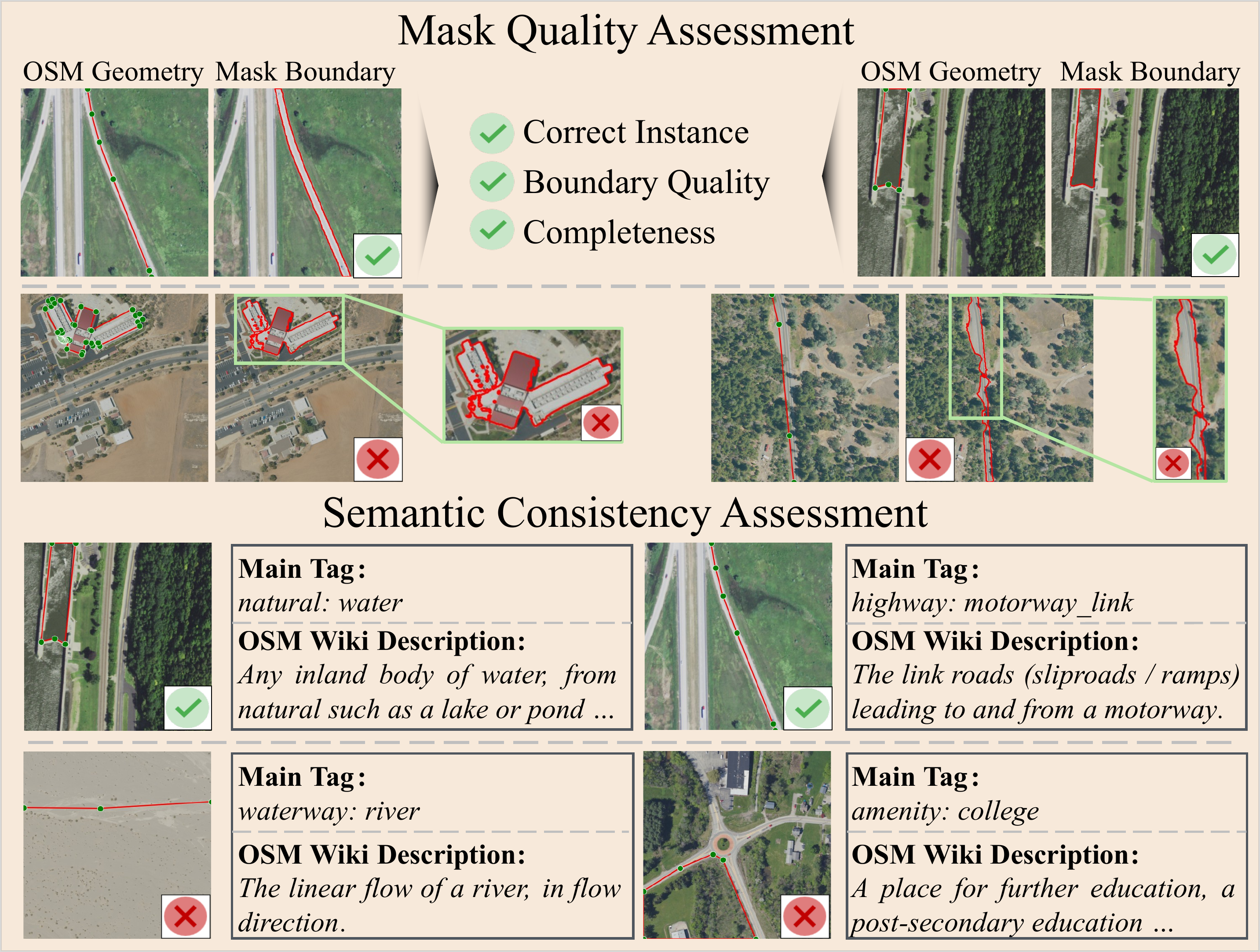}
    \caption{Representative accepted and rejected examples for decoupled quality assessment in Stage~II: mask quality (top) and semantic consistency (bottom).}    \label{fig:quality_assessment_examples}
\end{figure}

The two tasks require complementary capabilities: mask-quality assessment emphasizes fine-grained spatial alignment and boundary perception, whereas semantic-consistency assessment relies on geographic-category understanding and contextual interpretation. Separate prompt templates are used for area and non-area targets in both tasks (see Figures~\ref{fig:prompt_mask_quality_area} and~\ref{fig:prompt_semantic_consistency_area} for the prompt template). 
Figure~\ref{fig:quality_assessment_examples} illustrates accepted and rejected candidates for both assessment tasks. Only candidates accepted by both tasks are retained for Stage~III.

\begin{table}[!htbp]
\centering
\small
\setlength{\tabcolsep}{5pt}
\renewcommand{\arraystretch}{1.0}
\caption{Hyperparameters for task-specific quality-assessment fine-tuning.}
\begin{tabular*}{0.95\columnwidth}{@{\extracolsep{\fill}}ll@{}}
\toprule
Hyperparameter & Setting \\
\midrule
LoRA rank / alpha & 64 / 128 \\
Training epochs & 4 \\
Initial learning rate & \(1\times10^{-4}\) \\
Optimizer & AdamW \\
LR schedule & Cosine decay \\
Warmup ratio & 0.05 \\
Weight decay & 0 \\
Effective batch size & 16 \\
Maximum sequence length & 4,096 \\
Maximum image pixels & 262,144 \\
Numerical precision & BF16 \\
\bottomrule
\end{tabular*}
\label{tab:hyperparameters}
\end{table}

\begin{table*}[!htbp]
    \centering
    \footnotesize
    \setlength{\tabcolsep}{1.2pt}
    \renewcommand{\arraystretch}{0.95}
    \caption{MLLM performance (\%) on mask-quality and semantic-consistency assessment. The first two groups report zero-shot results without task-specific fine-tuning; the third reports Qwen models after task-specific supervised fine-tuning (SFT). Models selected for mask-quality and semantic-consistency filtering are marked with \(\dagger_{\mathrm{M}}\) and \(\dagger_{\mathrm{S}}\), respectively. The best result in each column is shown in bold.}
    \begin{tabular*}{\textwidth}{@{\extracolsep{\fill}}l*{8}{c}@{\hspace{3pt}}*{8}{c}@{}}
        \toprule
        \multirow{3}{*}{Model}
        & \multicolumn{8}{c}{\textit{Mask-quality assessment}}
        & \multicolumn{8}{c}{\textit{Semantic-consistency assessment}} \\
        \cmidrule(lr){2-9}\cmidrule(lr){10-17}
        & \multicolumn{4}{c}{Validation} & \multicolumn{4}{c}{Test}
        & \multicolumn{4}{c}{Validation} & \multicolumn{4}{c}{Test} \\
        \cmidrule(lr){2-5}\cmidrule(lr){6-9}\cmidrule(lr){10-13}\cmidrule(lr){14-17}
        & Acc. & Prec. & Rec. & F1 & Acc. & Prec. & Rec. & F1
        & Acc. & Prec. & Rec. & F1 & Acc. & Prec. & Rec. & F1 \\
        \midrule
        Qwen3.7-Plus
        & 65.40 & 64.03 & 88.69 & 74.37
        & 64.87 & 62.93 & 91.49 & 74.57
        & 75.00 & 74.92 & 83.21 & 78.85
        & 73.65 & 72.90 & 83.87 & 78.00 \\
        GPT-5.5
        & 65.60 & 63.06 & 94.70 & 75.71
        & 65.87 & 63.00 & 95.39 & 75.88
        & 71.00 & 68.91 & 87.86 & 77.24
        & 67.86 & 67.25 & 82.44 & 74.07 \\
        GPT-6.1-Sol
        & 62.60 & 60.57 & 97.17 & 74.63
        & 63.27 & 60.99 & 96.45 & 74.73
        & 69.00 & 65.82 & \textbf{92.86} & 77.04
        & 66.67 & 64.00 & 91.76 & 75.41 \\
        Gemini-3.1-Pro
        & 67.60 & 70.79 & 72.79 & 71.78
        & 67.47 & 69.38 & 75.53 & 72.33
        & 72.20 & 69.53 & 89.64 & 78.32
        & 71.06 & 69.48 & 85.66 & 76.73 \\
        Kimi-K2.7-Code
        & 63.20 & 61.76 & 91.87 & 73.86
        & 63.07 & 61.52 & 91.84 & 73.68
        & 71.60 & 70.78 & 83.93 & 76.80
        & 69.66 & 68.30 & 84.95 & 75.72 \\
        \midrule
        Qwen3.5-9B (Zero-shot)
        & 62.80 & 61.10 & 94.35 & 74.17
        & 57.09 & 57.67 & 89.36 & 70.10
        & 70.80 & 69.14 & 86.43 & 76.83
        & 72.46 & 69.42 & 90.32 & 78.50 \\
        Qwen3.5-27B (Zero-shot)
        & 63.20 & 60.93 & \textbf{97.53} & 75.00
        & 60.88 & 59.31 & \textbf{97.16} & 73.66
        & 67.60 & 64.82 & 92.14 & 76.11
        & 68.86 & 64.68 & \textbf{97.13} & 77.65 \\
        Qwen3.6-27B (Zero-shot)
        & 68.60 & 82.47 & 56.54 & 67.09
        & 70.86 & \textbf{84.00} & 59.57 & 69.71
        & 72.40 & 76.69 & 72.86 & 74.73
        & 71.46 & 75.00 & 73.12 & 74.05 \\
        \midrule
        Qwen3.5-9B (SFT)
        & 80.00 & 78.33 & 89.40 & 83.50
        & 77.05 & 74.78 & 89.36 & 81.42
        & 81.80 & 85.39 & 81.43 & 83.36
        & 85.43 & \textbf{85.03} & 89.61 & 87.26 \\
        Qwen3.5-27B (SFT)\(\dagger_{\mathrm{S}}\)
        & \textbf{83.00} & \textbf{83.45} & 87.28 & \textbf{85.32}
        & 81.04 & 80.06 & 88.30 & 83.98
        & 84.20 & \textbf{86.28} & 85.36 & 85.82
        & \textbf{85.83} & 84.44 & 91.40 & \textbf{87.78} \\
        Qwen3.6-27B (SFT)\(\dagger_{\mathrm{M}}\)
        & 82.40 & 83.05 & 86.57 & 84.78
        & \textbf{82.44} & 81.49 & 89.01 & \textbf{85.08}
        & \textbf{85.00} & 85.71 & 87.86 & \textbf{86.77}
        & 84.63 & 84.35 & 88.89 & 86.56 \\
        \bottomrule
    \end{tabular*}
    \label{tab:mask_understanding_horizontal}
\end{table*}

\paragraph{Training Details.}

Recent studies have employed vision-language models for candidate-mask filtering and segmentation quality assessment~\citep{ni2026unigeoseg,shi2025remote}, yet general-purpose multimodal models remain limited in fine-grained pixel-level understanding~\citep{zhu2025segagent}. Since both components of our decoupled quality control require precise interpretation of masked regions, we evaluate multiple models on mask-quality and semantic-consistency tasks and investigate whether task-specific post-training improves their reliability.
Specifically, we independently annotate 5,000 geometrically filtered candidates for both binary tasks and split the annotations for each task into 3,999 training, 500 validation, and 501 test samples. Under this unified protocol, we benchmark five general-purpose MLLMs in the zero-shot setting and evaluate Qwen3.5-9B, Qwen3.5-27B, and Qwen3.6-27B both before and after task-specific supervised fine-tuning (SFT). These paired comparisons assess the effect of task-specific adaptation within each model. We report accuracy, precision, recall, and F1 for both assessment tasks.

We fine-tune the task-specific quality assessors through LoRA-based supervised fine-tuning using LLaMA-Factory. During model selection, each Qwen-series model is trained on the 3,999 training samples for 4 epochs (1,000 optimizer steps). After selecting Qwen3.6-27B for mask quality and Qwen3.5-27B for semantic consistency, each chosen model is retrained on all 5,000 annotations for 4 epochs (1,252 optimizer steps) before being applied to the full candidate pool. Experiments are conducted on Ubuntu 20.04 using two NVIDIA A800 80GB GPUs. The software environment comprised Python 3.12.13, PyTorch 2.7.1 with CUDA 11.8, and Transformers 5.2.0. Both tasks share the hyperparameters in Table~\ref{tab:hyperparameters}, with LoRA adapters applied to all eligible linear modules in the vision encoder, multimodal projector, and language model.

\paragraph{Evaluation Results.}
Table~\ref{tab:mask_understanding_horizontal} provides the complete validation and test results, including accuracy, precision, recall, and F1 for both tasks. 
The general-purpose models generally favor recall over precision, indicating a tendency to accept unreliable candidates. Task-specific fine-tuning consistently improves performance on both tasks, demonstrating the benefit of adapting models to fine-grained mask interpretation. 
We select Qwen3.6-27B~\citep{qwen3.6-27b} for mask-quality assessment and Qwen3.5-27B~\citep{qwen3.5} for semantic-consistency assessment. 
On the test set, the selected models achieve the highest F1 scores of 85.08\% and 87.78\% for mask-quality and semantic-consistency, outperforming the strongest zero-shot closed-source results by 9.20 and 9.78 percentage points, respectively.

\subsection{Taxonomy-Guided Reasoning Annotation}

\begin{figure}[!htbp]
    \centering
    \includegraphics[width=\columnwidth]{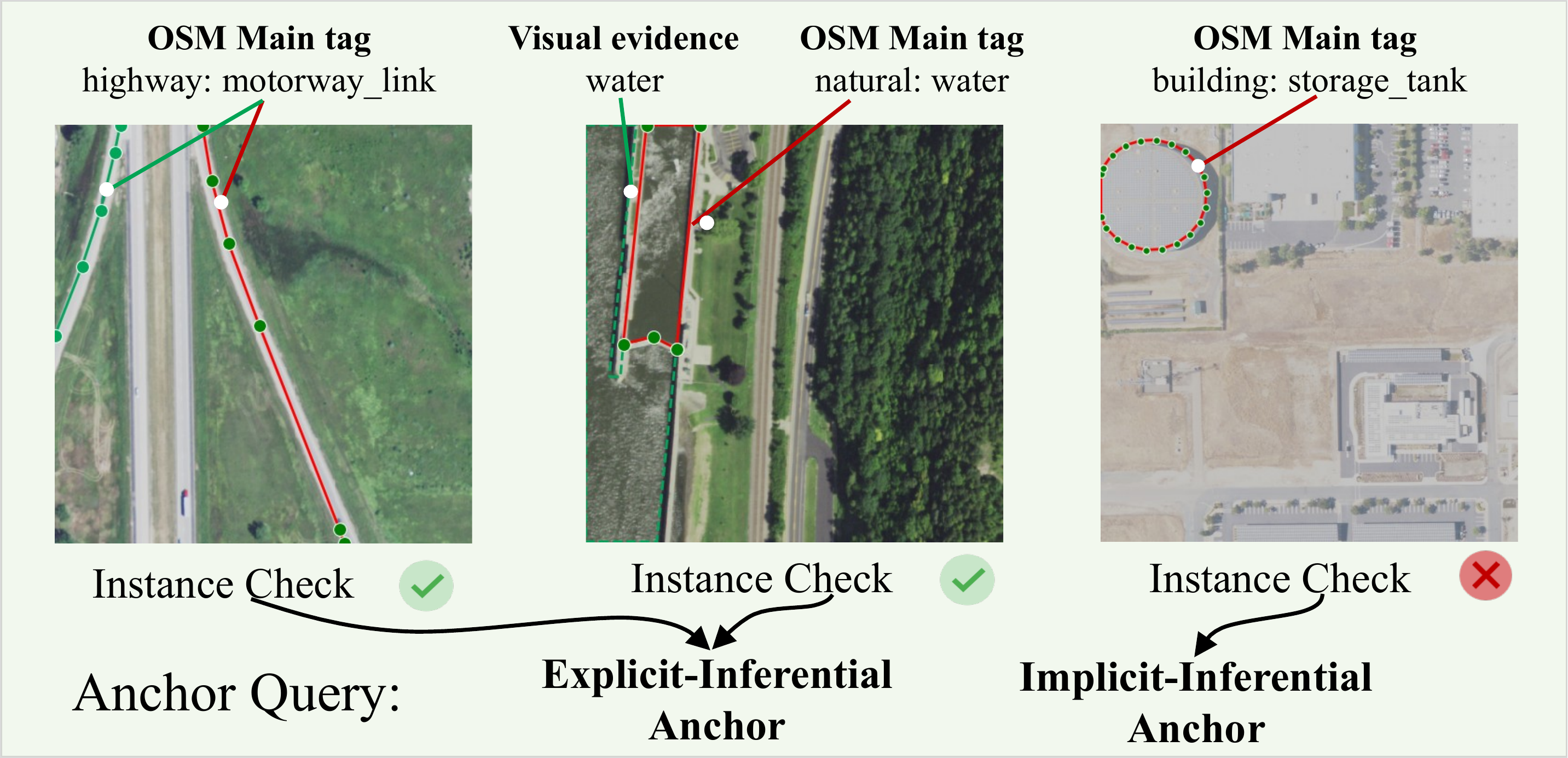}
    \caption{Representative examples of competing-instance detection in Stage~III: OSM-annotated (left), visually identified (middle), and no competitors (right).}
    \label{fig:competing_instance_examples}
\end{figure}

\subsubsection{Anchor Query Generation}
We identify visible same-category competitors using both OSM semantics and visual evidence: OSM elements sharing the target main tag are first examined, while Qwen3.5-122B-A10B~\citep{qwen3.5} inspects the image for unannotated instances. The aerial image serves as the primary evidence, with the main tag and its OSM Wiki description specifying the category under consideration. Figure~\ref{fig:competing_instance_examples} illustrates the three outcomes of this competing-instance check and how they route targets to explicit-inferential (EI) or implicit-inferential (II) anchor generation. Based on this competition check, we employ Qwen3.7-Plus~\citep{qwen37plus} to generate an anchor query--answer pair from the image, main tag, OSM Wiki description, and complete tag set, while additionally supplying the semantic context of competing instances for explicit-inferential (EI) generation.
Prior to query generation, we extend RSTeller's tag-cleaning rules with refined regular-expression filters to remove residual metadata and import noise (e.g., GIS import fields, references, addresses, and external links) from the full OSM tag set.

As shown in our benchmark analysis, some existing datasets specify targets by stacking multiple mixed perceptual and inferential attributes within a single query without ensuring that localization genuinely requires reasoning. To prevent models from solving the task through perceptual shortcuts, we introduce the \textit{Minimal Sufficient Reasoning Cues Principle} to constrain query generation:
``A reasoning query should contain the smallest set of cues sufficient to uniquely identify the target while preserving at least one indispensable inferential cue. Redundant perceptual cues should be excluded to avoid shortcut solutions.''
For example, ``the house most prone to flooding'' is preferable to ``the house most prone to flooding in the lower-left area,'' because the additional positional cue may enable localization without reasoning about flood risk. This principle is enforced in our generation prompts (Figures~\ref{fig:prompt_qa_generation_q3} and~\ref{fig:prompt_qa_generation_q4}) and further safeguarded through subsequent model and human verification.

\subsubsection{Mask and Anchor Query Verification}

Prior to anchor query generation, all 165,023 candidate masks accepted in Stage~II undergo manual inspection for target correctness, boundary quality, completeness, and contamination, retaining 147,472 verified targets. After anchor query generation, we assess whether each generated anchor query is valid and consistent with its designated inferential setting.

For query verification, we manually annotate 503 anchor queries, comprising 456 positive and 47 negative samples. We formulate verification as a scoring task in which each model assigns two scores from 0 to 10: one for the correctness of the inferential reasoning and the other for the uniqueness of target specification (see Figure~\ref{fig:prompt_query_verification} for the prompt template). The two scores are converted into binary validity decisions using predefined thresholds of 7 and 8, respectively. Table~\ref{tab:query_verification} reports the resulting metrics against the manually annotated labels at the selected operating thresholds.

\begin{table}[!htbp]
    \centering
    \footnotesize
    \setlength{\tabcolsep}{3.5pt}
    \renewcommand{\arraystretch}{1.0}
    \caption{Performance on query validity verification (\%). The model selected for full-dataset verification is marked with \(\dagger\).}
    \begin{tabular}{@{}lccccc@{}}
        \toprule
        Model & Prec. & Rec. & Spec. & Bal. Acc. & F1 \\
        \midrule
        Qwen3.7-Plus\(\dagger\)
        & \textbf{95.7} & 78.3 & 66.0 & \textbf{72.1} & 86.1 \\

        Qwen3.7-Max (thinking)
        & 95.0 & 79.8 & 59.6 & 69.7 & 86.8 \\

        Qwen3.7-Max (non-thinking)
        & 91.8 & 76.5 & 34.0 & 55.3 & 83.5 \\

        Kimi-K2.7-Code
        & 94.7 & 77.6 & 57.4 & 67.5 & 85.3 \\

        InternVL3.5-38B
        & 94.7 & 54.6 & \textbf{70.2} & 62.4 & 69.3 \\

        Qwen3.5-122B-A10B
        & 91.3 & \textbf{94.1} & 12.8 & 53.4 & 92.7 \\
        \bottomrule
    \end{tabular}
    \label{tab:query_verification}
\end{table}

Qwen3.7-Plus achieves the highest precision and balanced accuracy, while maintaining competitive recall and specificity. In contrast, models with higher recall or F1 often exhibit substantially lower specificity, indicating a tendency to retain invalid annotations. Because false positives directly reduce the quality of the released dataset, we select Qwen3.7-Plus as the final verifier. Applying the selected thresholds to the 147,472 mask-verified samples retains 106,610 final targets (33,043 EI anchors and 73,567 II anchors). To further audit the quality of the retained annotations, human annotators inspect a random sample of 6,523 verified queries and their accompanying reasoning answers, among which 6,228 pass and 295 fail, yielding a human acceptance rate of 95.48\%.

\subsubsection{Paired Query Completion}
\label{app:paired_query}

To provide paired EI and II supervision across the dataset while keeping the verified inferential clause fixed, we apply controlled lexical transformation to switch each anchor query's nominal reference between explicit and implicit forms. Specifically, implicit hypernyms in II anchors are replaced with canonical category names to form EI queries, whereas explicit categories in EI anchors are replaced with coarser hypernyms to form II queries. Because substituting the canonical category only modifies a short phrase and narrows the candidate space without introducing ambiguity, direct replacement suffices in this direction, as confirmed by human inspection of sampled queries. Conversely, directly replacing an explicit category with a coarser hypernym broadens the candidate domain and may introduce cross-category ambiguity if other visible objects also satisfy the inferential cue. We therefore employ Qwen3.7-Plus to inspect all 30,259 transformed II queries in the training and validation splits against the image for referential uniqueness. Whenever the broadened hypernym creates ambiguity, we add a minimal functional modifier (e.g., ``area'' to ``cultivated area'') to narrow the candidate domain and restore unique target specification while keeping the category implicit. Across the 30,259 queries, 24,712 (81.67\%) retain the direct lexical replacement and 5,547 (18.33\%) receive minimal functional refinement.

\begin{table}[!htbp]
    \centering
    \footnotesize
    \setlength{\tabcolsep}{4pt}
    \caption{Sample retention throughout the RSReasoner construction pipeline. Retention rates are measured relative to the initial candidate pool. Mask-quality and semantic-consistency filtering are performed in parallel.}
    \begin{tabular}{@{}lrr@{}}
        \toprule
        Stage & Retained samples & Retention (\%) \\
        \midrule
        Initial candidate pool
        & 1,309,926 & 100.00 \\

        Geometric filtering
        & 303,023 & 23.13 \\

        Mask-quality filtering
        & 204,187/303,023 & -- \\

        Semantic-consistency filtering
        & 215,075/303,023 & -- \\

        Joint quality acceptance
        & 165,023 & 12.60 \\

        Human mask verification
        & 147,472 & 11.26 \\

        Query-validity verification
        & 106,610 & 8.14 \\
        \bottomrule
    \end{tabular}
    \label{tab:sample_retention}
\end{table}

\subsection{Sample Retention Statistics}

Table~\ref{tab:sample_retention} further summarizes retention across the complete construction pipeline. Because mask-quality and semantic-consistency filtering operate in parallel, their individual counts are computed from the same geometrically filtered pool, and joint quality acceptance denotes their intersection.

\section{RSReasoner-Bench Construction Details}
\label{app:bench_verification}

\subsection{Target Sampling and Query Construction}

To construct RSReasoner-Bench without spatial or instance-level leakage, we first group all 106,610 verified targets in RSReasoner by source image and OSM element identity so that samples sharing imagery or geographic entities are assigned to the same split. From these groups, we sample 5,800 targets across OSM main tags using category-stratified sampling to mitigate long-tail dominance while maintaining a balanced composition across target geometries (2,800 area and 3,000 non-area targets) and competition conditions (2,784 targets with same-category competitors and 3,016 without). We use RSReasoner-Bench as the held-out test set and partition the remaining groups into 95,150 training and 5,660 validation targets under the same group constraint to prevent leakage.

Across the 5,800 targets, each query condition contains 5,800 queries paired with the same images and ground-truth masks:

\textbf{Canonical four-quadrant queries (EP, IP, EI, and II).}
For each target, we independently generate four canonical queries via Qwen3.7-Plus covering explicit-perceptual (EP), implicit-perceptual (IP), explicit-inferential (EI), and implicit-inferential (II) settings (23,200 queries in total). 

\textbf{Category-only reference (C).}
We extract the canonical category name from the target's normalized main tag and format it using a predefined set of concise referring templates without any attributes (e.g., ``Segment the farmland plot.'' or ``Find the residential road.'').

\textbf{Controlled single-attribute and mixed-attribute queries (Ctrl-EP, Ctrl-EI, M-PI, and M-IP).}
To eliminate syntactic variation across query types, we use the exact template and category phrase sampled for C as a shared syntactic backbone, and rewrite the verified EP and EI queries via Qwen3.7-Plus while preserving their original perceptual or inferential cues to form Ctrl-EP (e.g., ``Segment the farmland plot on the right.'') and Ctrl-EI (e.g., ``Segment the farmland plot that appears to contain actively growing vegetation.''). Building on this unified template, we merge the perceptual cues from EP and the inferential cues from EI into a single query to analyze the effect of mixed attributes. To control for attribute-order sensitivity, we construct both perceptual-first (M-PI, e.g., ``Segment the farmland plot on the right that appears to contain actively growing vegetation.'') and inferential-first (M-IP, e.g., ``Segment the farmland plot that appears to contain actively growing vegetation and is located on the right.'') variants.

\textbf{Category-replacement variants (EP-Imp, IP-Exp, EI-Imp, and II-Exp).}
To isolate the effect of category expression while keeping attribute evidence fixed, we preserve the attribute descriptions of the verified EP, IP, EI, and II queries and only switch the target nominal reference between its explicit category and an implicit hypernym. Specifically, replacing the explicit category in EP and EI with a coarser hypernym yields EP-Imp and EI-Imp (e.g., changing ``Which residential road is most susceptible to flooding from the visible river?'' to ``Which feature is most susceptible to flooding from the visible river?''), whereas replacing the hypernym in IP and II with the explicit category name yields IP-Exp and II-Exp.

\subsection{Quality Verification and Human Curation}
All manual inspections, verifications, and revisions throughout our pipeline are conducted by five Ph.D. annotators specializing in remote sensing and computer vision. For RSReasoner-Bench, we employ GPT-6-Astra~\citep{openai2026gpt6} to screen all 23,200 canonical queries (EP, IP, EI, and II) across three criteria: (1) visual and geographic factuality against the image and OSM semantics, (2) adherence to the designated explicit or implicit category constraint, and (3) adherence to the designated perceptual or inferential attribute constraint without attribute mixing. Across the 23,200 queries, this screening identifies 1,309 flagged queries in total, including 130 (0.56\%) with factual inaccuracies, 231 (1.00\%) with category expression errors, and 948 (4.09\%) with mixed attributes.

All 1,309 flagged queries are manually reviewed and rewritten by our annotators to correct factual errors, remove leaked or missing category terms, and eliminate mixed-attribute shortcuts. Combined with stratified sampling inspection, direct human review and revision cover 2,371 queries (10.22\% of all 23,200 canonical queries), with focused coverage on 2,108 of the 11,600 more challenging inferential queries (18.17\% of EI and II), spanning 1,996 of the 5,800 benchmark images (34.41\%). The remaining diagnostic variants (C, Ctrl-EP, Ctrl-EI, M-PI, M-IP, and the category-replacement queries) only involve straightforward template formatting, lexical substitution, or syntactic merging of already verified clauses via Qwen3.7-Plus, which rarely introduces errors; moreover, human inspection of 100 randomly sampled queries per variant type found no errors, further supporting their reliability.

\section{Additional Dataset Statistics}
\label{app:dataset_statistics}

To complement the dataset overview in Section~\ref{sec:dataset_statistics}, Tables~\ref{tab:dataset_statistics} and~\ref{tab:bench_statistics} summarize the data scale, imagery properties, OSM semantic coverage, target geometries, competition conditions, and query compositions of RSReasoner and RSReasoner-Bench, respectively. The following subsections further detail their query and answer distributions, mask area ratios, and OSM semantic tag profiles.

\begin{table}[!t]
    \centering
    \footnotesize
    \setlength{\tabcolsep}{5pt}
    \renewcommand{\arraystretch}{1.05}
    \caption{Detailed statistics of RSReasoner.}
    \begin{tabularx}{\columnwidth}{@{}Xr@{}}
        \toprule
        Statistic & Value \\
        \midrule
        \multicolumn{2}{l}{\textit{Data scale and imagery}} \\
        Number of images & 106,610 \\
        Number of masks & 106,610 \\
        Image resolution & \(448\times448\) pixels \\
        Ground sample distance (GSD) & \(0.6\,\mathrm{m/pixel}\) \\
        Number of query--answer pairs & 213,220 \\
        \midrule
        \multicolumn{2}{l}{\textit{OSM semantics}} \\
        Unique normalized main tags & 228 \\
        Unique normalized main keys & 36 \\
        Unique full OSM tags & 44,564 \\
        Unique full OSM keys & 913 \\
        \midrule
        \multicolumn{2}{l}{\textit{Target geometry}} \\
        Area targets & 54,012 (50.7\%) \\
        Non-area targets & 52,598 (49.3\%) \\
        \midrule
        \multicolumn{2}{l}{\textit{Same-category competition}} \\
        With competitors (EI anchors) & 33,043 (31.0\%) \\
        Without competitors (II anchors) & 73,567 (69.0\%) \\
        \midrule
        \multicolumn{2}{l}{\textit{Query type}} \\
        Explicit-inferential (EI) queries & 106,610 (50.0\%) \\
        Implicit-inferential (II) queries & 106,610 (50.0\%) \\
        \bottomrule
    \end{tabularx}
    \label{tab:dataset_statistics}
\end{table}

\begin{table}[!t]
    \centering
    \footnotesize
    \setlength{\tabcolsep}{5pt}
    \renewcommand{\arraystretch}{1.05}
    \caption{Detailed statistics of RSReasoner-Bench.}
    \begin{tabularx}{\columnwidth}{@{}Xr@{}}
        \toprule
        Statistic & Value \\
        \midrule
        \multicolumn{2}{l}{\textit{Data scale and imagery}} \\
        Number of images & 5,800 \\
        Number of masks & 5,800 \\
        Image resolution & \(448\times448\) pixels \\
        Ground sample distance (GSD) & \(0.6\,\mathrm{m/pixel}\) \\
        Canonical queries (EP, IP, EI, II) & 23,200 \\
        Number of queries & 75,400 \\
        \midrule
        \multicolumn{2}{l}{\textit{OSM semantics}} \\
        Unique normalized main tags & 215 \\
        Unique normalized main keys & 35 \\
        Unique full OSM tags & 4,288 \\
        Unique full OSM keys & 374 \\
        \midrule
        \multicolumn{2}{l}{\textit{Target geometry}} \\
        Area targets & 2,800 (48.3\%) \\
        Non-area targets & 3,000 (51.7\%) \\
        \midrule
        \multicolumn{2}{l}{\textit{Same-category competition}} \\
        With competitors & 2,784 (48.0\%) \\
        Without competitors & 3,016 (52.0\%) \\
        \midrule
        \multicolumn{2}{l}{\textit{Query type}} \\
        Explicit-Perceptual (EP) & 5,800 \\
        Implicit-Perceptual (IP) & 5,800 \\
        Explicit-Inferential (EI) & 5,800 \\
        Implicit-Inferential (II) & 5,800 \\
        Category-only reference (C) & 5,800 \\
        Controlled Explicit-Perceptual (Ctrl-EP) & 5,800 \\
        Controlled Explicit-Inferential (Ctrl-EI) & 5,800 \\
        Controlled Perceptual-first mixed (M-PI) & 5,800 \\
        Controlled Inferential-first mixed (M-IP) & 5,800 \\
        Implicitized Explicit-Perceptual (EP-Imp) & 5,800 \\
        Explicitized Implicit-Perceptual (IP-Exp) & 5,800 \\
        Implicitized Explicit-Inferential (EI-Imp) & 5,800 \\
        Explicitized Implicit-Inferential (II-Exp) & 5,800 \\
        \bottomrule
    \end{tabularx}
    \label{tab:bench_statistics}
\end{table}

\subsection{Query and Answer Statistics}

\begin{figure}[!htbp]
    \centering
    \includegraphics[width=\columnwidth]{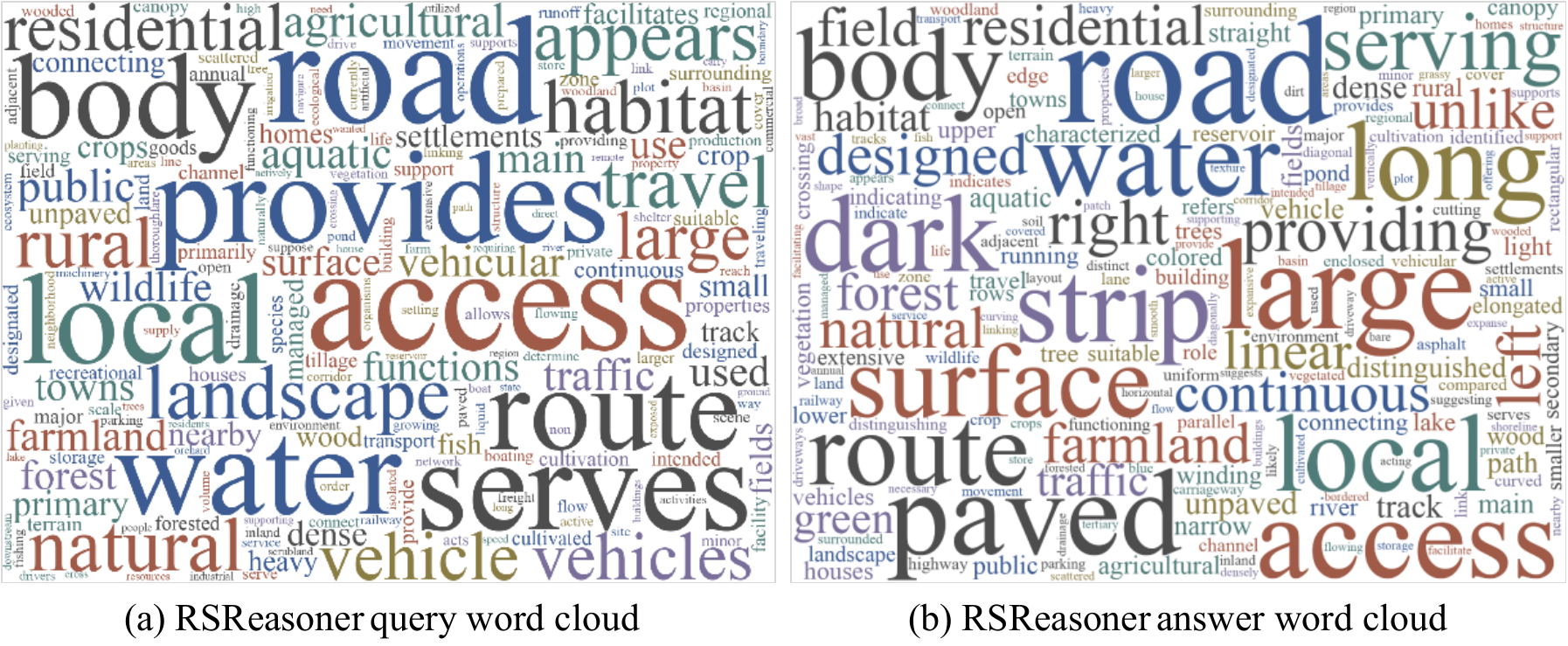}
    \caption{Word clouds of (a) queries and (b) answers in RSReasoner.}
    \label{fig:wordcloud}
\end{figure}

\begin{figure}[!htbp]
    \centering
    \includegraphics[width=\columnwidth]{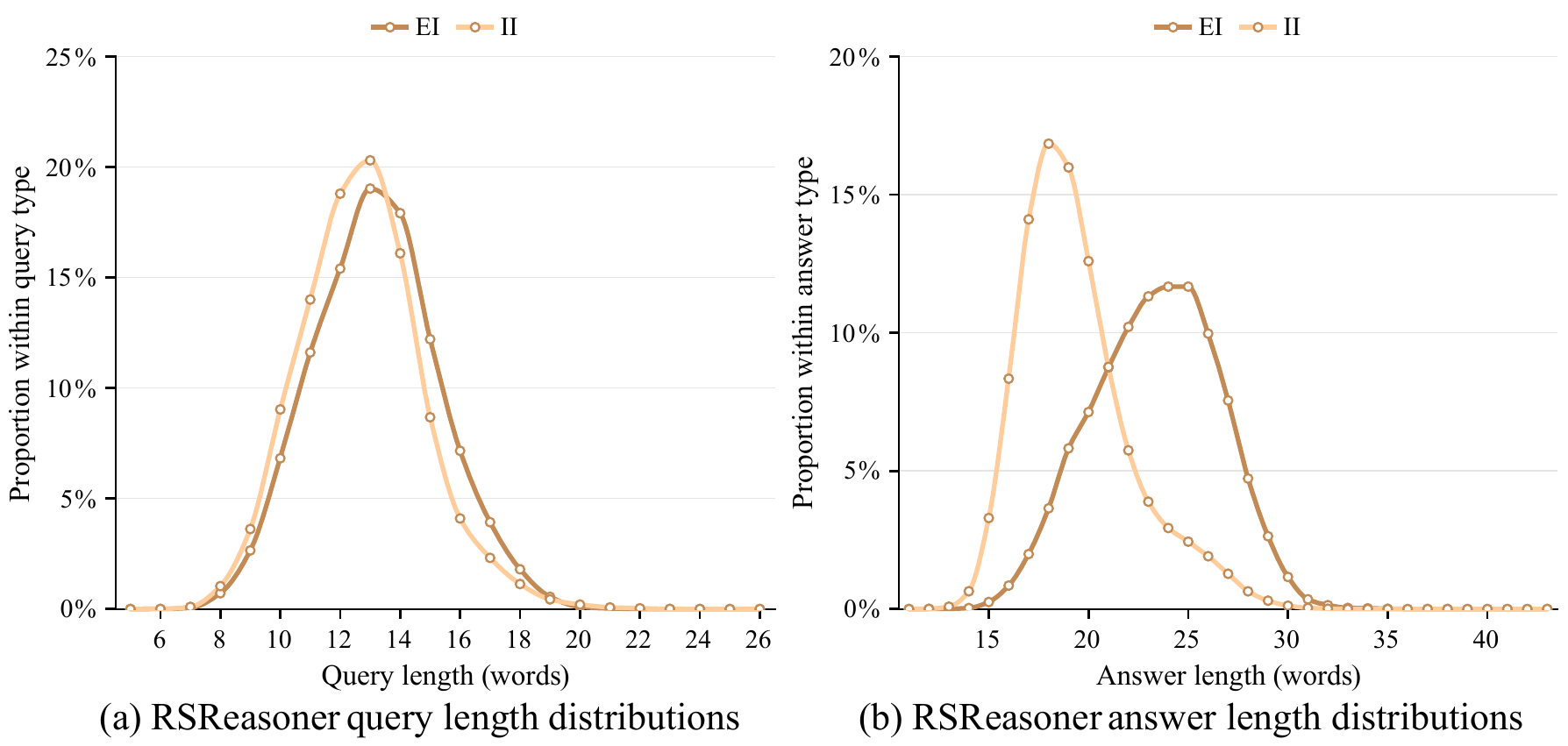}
    \caption{Length distributions of (a) queries and (b) answers for EI-anchored and II-anchored subsets of RSReasoner.}
    \label{fig:RSReasoner-length-distribution}
\end{figure}

Figures~\ref{fig:wordcloud} and~\ref{fig:RSReasoner-length-distribution} illustrate the lexical composition and length distributions of queries and answers in RSReasoner. As shown in Figure~\ref{fig:RSReasoner-length-distribution}(a), both EI-anchored and II-anchored queries remain compact and closely aligned, peaking at 13 words in accordance with the \textit{Minimal Sufficient Reasoning Cues Principle} to avoid redundant attribute stacking. In Figure~\ref{fig:RSReasoner-length-distribution}(b), EI answers are longer than II answers, as explicit-inferential anchors on competing scenes require additional contrastive explanation to distinguish the target from same-category instances.

\begin{figure}[!htbp]
    \centering
    \includegraphics[width=\columnwidth]{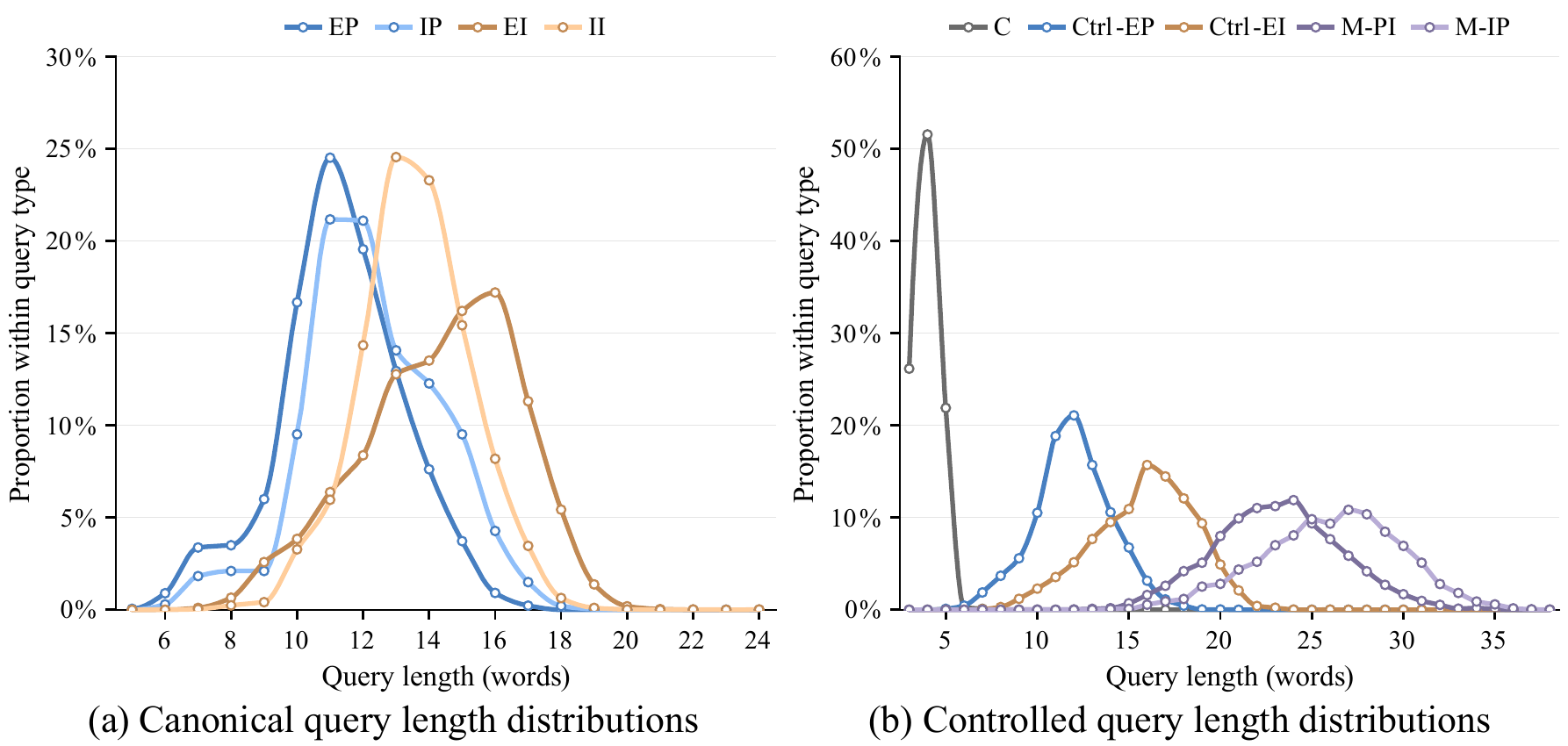}
    \caption{Query length distributions of RSReasoner-Bench across (a) canonical query types and (b) category-only and attribute-composition queries.}
    \label{fig:RSReasonerBench-length-distribution}
\end{figure}

Figure~\ref{fig:RSReasonerBench-length-distribution} further reports query length distributions across RSReasoner-Bench. For the four canonical types in Figure~\ref{fig:RSReasonerBench-length-distribution}(a), all queries remain concise, with inferential queries (EI and II) slightly longer than perceptual queries (EP and IP) because expressing functional or relational reasoning requires more phrasing than direct visual descriptors. For the controlled diagnostic variants in Figure~\ref{fig:RSReasonerBench-length-distribution}(b), query lengths reflect the intended structural progression: category-only references (C) concentrate at 3--5 words, template-aligned single-attribute variants (Ctrl-EP and Ctrl-EI) preserve the length profiles of EP and EI, and mixed-attribute queries (M-PI and M-IP) shift rightward as they combine both perceptual and inferential clauses.

\subsection{Mask Area Ratio Statistics}

\begin{figure}[!htbp]
    \centering
    \includegraphics[width=0.85\columnwidth]{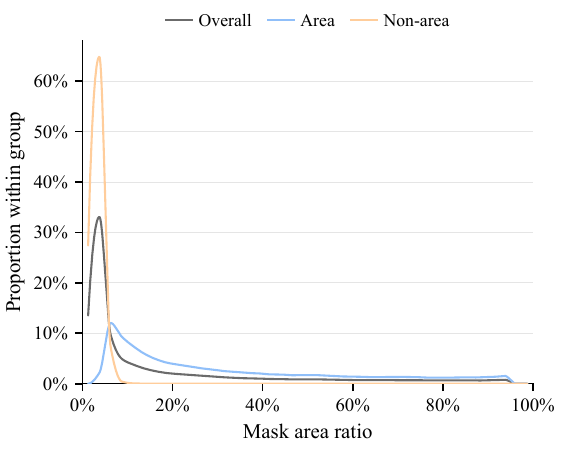}
    \caption{Distribution of Mask Area Ratios.}
    \label{fig:mask_area_ratio_distribution}
\end{figure}

Figure~\ref{fig:mask_area_ratio_distribution} shows the distribution of mask area ratios in RSReasoner, defined as the proportion of image pixels covered by each target mask. Area targets exhibit a broad distribution spanning from compact structures to expansive land-cover regions, whereas non-area targets are sharply concentrated at small ratios because this subset comprises narrow linear objects such as roads, railways, and waterways.

\subsection{OSM Tag Statistics}

\begin{figure}[!tbp]
    \centering
    \includegraphics[width=\columnwidth]{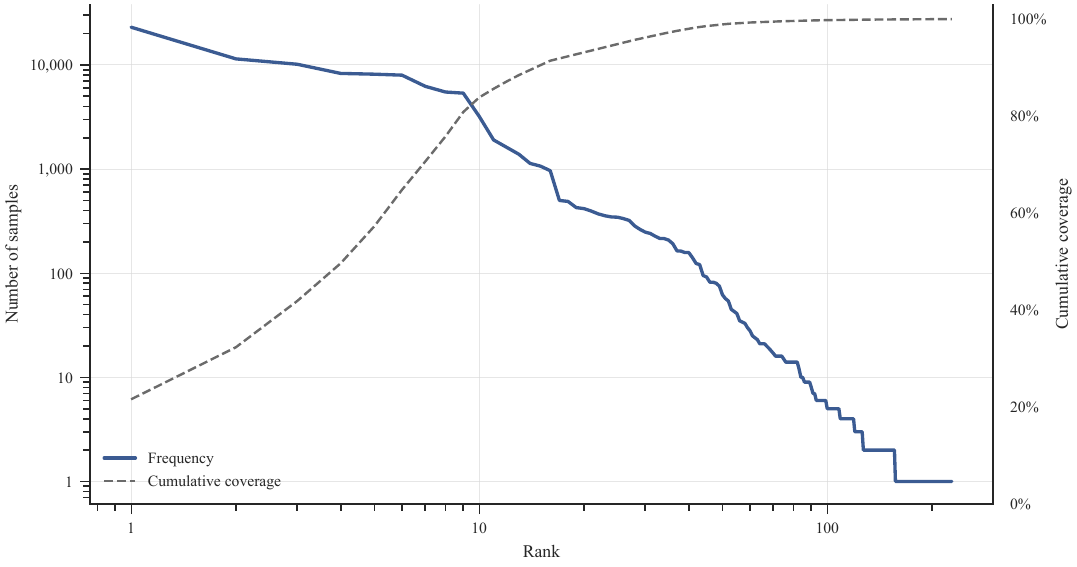}
    \caption{Main tag rank-frequency distribution and cumulative coverage in RSReasoner.}
    \label{fig:long_tail_main_tag_distribution}
\end{figure}

\begin{figure}[!tbp]
    \centering
    \includegraphics[width=\columnwidth]{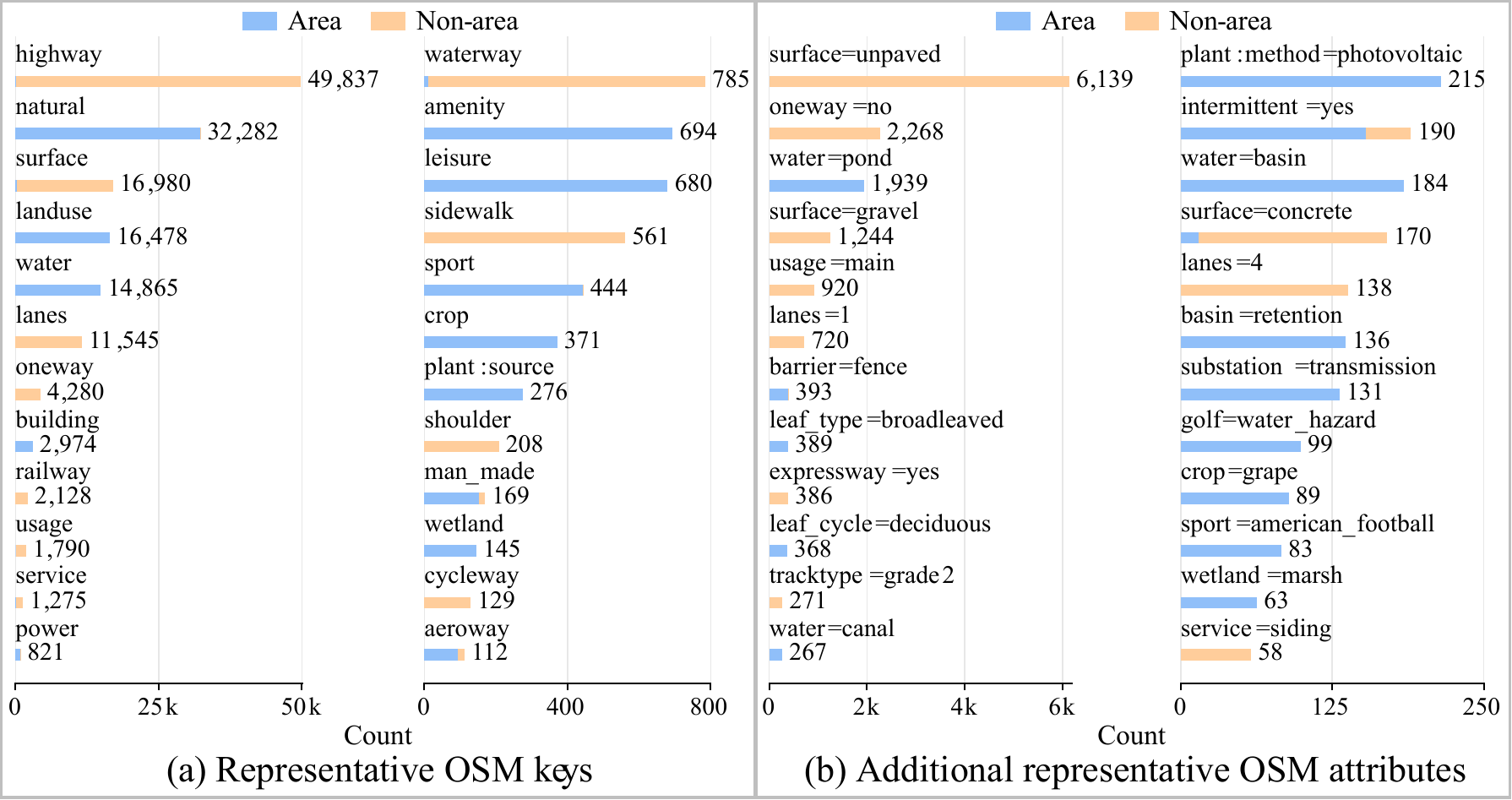}
    \caption{Representative (a) OSM keys and (b) additional OSM attributes in RSReasoner across area and non-area targets.}
    \label{fig:Additional-OSM-semantics-RSReasoner}
\end{figure}

\begin{figure}[!tbp]
    \centering
    \includegraphics[width=\columnwidth]{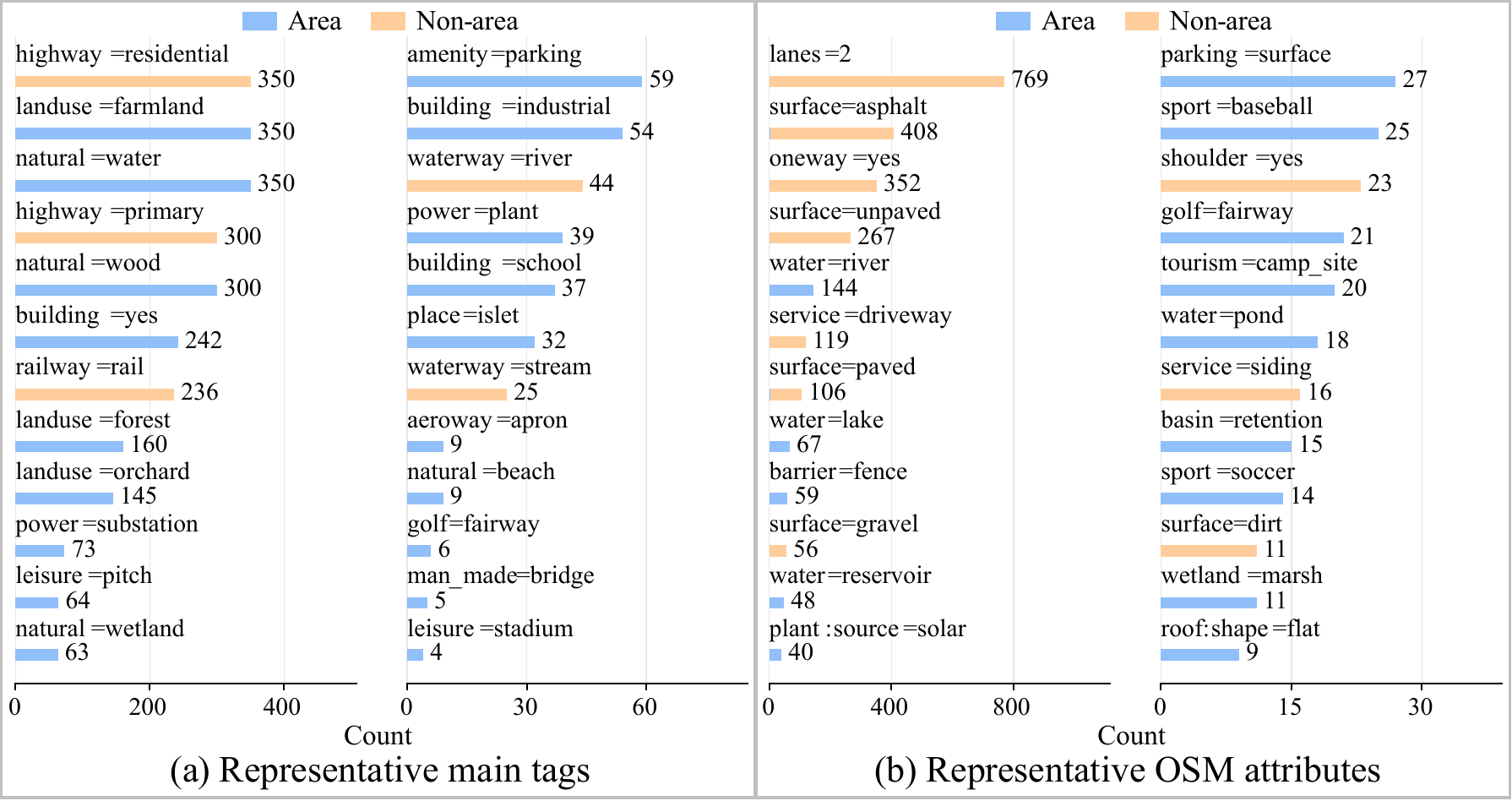}
    \caption{Representative (a) main tags and (b) OSM attributes in RSReasoner-Bench across area and non-area targets.}
    \label{fig:Representative-tags-RSReasoner-Bench}
\end{figure}

RSReasoner retains two complementary levels of OSM attributes: a normalized main tag that both anchors the target category and directly encodes fine-grained functional or domain semantics, and the complete tag set that supplies additional instance-level properties.
Overall, RSReasoner spans 228 normalized main tags across 36 main keys alongside 44,564 unique full key--value pairs across 913 keys, while RSReasoner-Bench preserves 215 main tags across 35 main keys and 4,288 unique full tags across 374 keys. 
As shown in Figure~\ref{fig:long_tail_main_tag_distribution}, RSReasoner exhibits a pronounced long-tail distribution in which the 20 most frequent main tags account for 99,290 targets (93.13\%) and the remaining 208 form a broad low-frequency tail reflecting the natural prevalence of geographic entities in aerial imagery. Complementing Figure~\ref{fig:main_semantic_distribution}(c,d), Figure~\ref{fig:Additional-OSM-semantics-RSReasoner} illustrates how additional OSM keys and fine-grained attributes align with target geometry, with non-area targets dominated by transportation and hydrographic attributes and area targets capturing land use, built infrastructure, and environmental roles. Figure~\ref{fig:Representative-tags-RSReasoner-Bench} further shows that category-stratified sampling in RSReasoner-Bench caps dominant head tags to maintain broad and balanced semantic coverage across its 5,800 test targets.
Table~\ref{tab:main-key-tag-values} lists the complete inventory of retained OSM main tags in RSReasoner, illustrating fine-grained open-vocabulary coverage across built, agricultural, transportation, hydrographic, energy, and ecological domains.

\begin{table*}[!htbp]
    \centering
    \scriptsize
    \setlength{\tabcolsep}{4pt}
    \renewcommand{\arraystretch}{1.05}
    \caption{Complete inventory of the 228 normalized OSM main tags across 36 main keys retained in RSReasoner, grouped by main key.}
    \label{tab:main-key-tag-values}
    \begin{tabularx}{\textwidth}{@{}>{\itshape\raggedright\arraybackslash}p{0.13\textwidth} >{\raggedright\arraybackslash}X@{}}
        \toprule
        \normalfont Main Key & \normalfont Main Tag Values \\
        \midrule
        building & apartments, barn, business, church, civic, college, commercial, detached, farm\_auxiliary, garage, garages, grandstand, greenhouse, hangar, hospital, hotel, industrial, mall, manufacture, occupied, office, parking, prison, public, Railway, residential, retail, roof, roof;yes, school, silo, slurry\_tank, sports\_hall, stable, storage\_tank, supermarket, transportation, university, warehouse, yes, yes;industrial \\
        landuse & animal\_keeping, basin, brownfield, cemetery, churchyard, commercial, construction, farm, farmland, farmyard, forest, garages, grass, greenfield, greenhouse\_horticulture, industrial, landfill, logistics, meadow, nature\_reserve, orchard, plant\_nursery, quarry, railway, recreation\_ground, reservoir, reservoir\_watershed, residential, retail, salt\_pond, storage, village\_green, vineyard, wasteland \\
        natural & bare\_rock, bay, beach, coastline, desert, fell, forest, glacier, grass, grassland, heath, mesa, mud, peninsula, reef, rock, sand, scree, scrub, shingle, stone, tree\_row, water, wetland, wood, woodland, yes \\
        highway & bridleway, cycleway, footway, motorway, motorway\_link, path, pedestrian, platform, primary, primary\_link, proposed, raceway, residential, road, secondary, service, services, tertiary, tertiary\_link, track, trunk, trunk\_link, unclassified \\
        man\_made & breakwater, bridge, embankment, heap, pier, pipeline, reservoir\_covered, silo, slurry\_lagoon, spoil\_heap, tailings\_pond, wastewater\_plant, water\_works, works \\
        leisure & garden, golf\_course, marina, miniature\_golf, park, pitch, sports\_centre, stadium, swimming\_pool, track, water\_park, yes \\
        amenity & college, festival\_grounds, fuel, grave\_yard, hospital, hunting\_stand, parking, place\_of\_worship, post\_office, school, university \\
        place & archipelago, farm, island, islet, locality, neighbourhood, quarter, suburb \\
        waterway & canal, dam, ditch, drain, pond, river, stream \\
        aeroway & aerodrome, apron, hangar, launchpad, runway, taxiway \\
        railway & abandoned, disused, narrow\_gauge, rail, razed, subway \\
        power & generator, line, plant, substation \\
        \normalfont\textit{Other keys (24)} & golf=fairway/rough/tee, shop=car/storage\_rental/supermarket, tourism=camp\_site/caravan\_site/hotel, area:aeroway=runway/taxiway, area:highway=secondary/track, barrier=fence/jersey\_barrier, name=feather river/Medicine Bow above Seminoe, office=company/government, abandoned:aeroway=runway, area=yes, boundary=place, building:part=yes, designation=Cunningham Lake NW, historic=memorial, indoor=room, NHD:FTYPE=StreamRiver, parking=surface, route=canoe, service=driveway, STREET\_NAM=ANDY AVE, substation=industrial, surface=gravel, was:waterway=riverbank, water=lake \\
        \bottomrule
    \end{tabularx}
\end{table*}

\section{Taxonomy Audit of Existing Benchmarks}
\label{app:benchmark_audit}

\paragraph{Audit Protocol and Semantic Anchors.}
We audit all 90,708 available language queries across EarthReason~\citep{li2025segearthr1}, GRES~\citep{quenum2026lisat}, LaSeRS~\citep{xin2026segearthr2}, and GeoSeg-Bench~\citep{ni2026unigeoseg} (excluding interactive point and box prompts), alongside the 23,200 canonical queries of RSReasoner-Bench. Using the prompt template in Figure~\ref{fig:prompt_taxonomy_audit}, we prompt GPT-6-Astra with 24 human-annotated dataset-specific few-shot examples and supply the remote sensing image, the target boundary overlay, the query text, and a category semantic anchor. Because determining category expression requires a semantic anchor to specify the target category, we use the provided target category labels for LaSeRS and GeoSeg-Bench, the reference answers for EarthReason, the category candidate list for GRES, and the normalized OSM main tags for RSReasoner-Bench.

\paragraph{Classification Criteria and Boundary Cases.}
Each query is evaluated along category expression and attribute composition independently of its original task label. For category expression, a query is classified as explicit if it directly names the target category at the reference granularity, and implicit if it uses a coarser hypernym or describes the target only through attributes. For attribute composition, queries are categorized as perceptual-only, inferential-only, or mixed based on whether they express only perceptual attributes, only inferential attributes, or both. Only descriptions characterizing the target itself are considered target attributes, whereas the target category noun, generic scene framing (e.g., ``in the image''), and remarks on image visibility are excluded. Under this criterion, 70 queries (10 in LaSeRS-ref and 60 in GRES-rea) contain no independent target attributes (e.g., ``The grass in the image.'' in LaSeRS, or ``Locate the bus in the image, despite the image being blurred.'' in GRES) and are therefore excluded from the six attribute-bearing classes.

\begin{table}[!htbp]
    \centering
    \footnotesize
    \setlength{\tabcolsep}{3.0pt}
    \renewcommand{\arraystretch}{1.05}
    \caption{Average number of perceptual and inferential attributes per query across existing benchmarks and RSReasoner-Bench. \(N\) denotes the number of evaluated queries.}
    \begin{tabular*}{\columnwidth}{@{\extracolsep{\fill}}lrrr@{}}
        \toprule
        Dataset / Subset & \(N\) & Perceptual & Inferential \\
        \midrule
        EarthReason-rea & 32,595 & 0.177 & 1.376 \\
        GRES-rea & 23,623 & 2.205 & 0.092 \\
        \midrule
        LaSeRS & 30,353 & 1.457 & 1.675 \\
        LaSeRS-ref & 8,729 & 1.870 & 0.946 \\
        LaSeRS-rea & 21,624 & 1.290 & 1.969 \\
        \midrule
        GeoSeg-Bench & 4,137 & 1.880 & 0.681 \\
        GeoSeg-Bench-ref & 2,354 & 2.605 & 0.067 \\
        GeoSeg-Bench-attr & 1,150 & 0.068 & 1.629 \\
        GeoSeg-Bench-con & 633 & 2.477 & 1.245 \\
        \midrule
        RSReasoner-Bench & 23,200 & 1.494 & 0.528 \\
        RSReasoner-Bench-EP & 5,800 & 2.608 & 0.001 \\
        RSReasoner-Bench-IP & 5,800 & 3.322 & 0.000 \\
        RSReasoner-Bench-EI & 5,800 & 0.027 & 1.088 \\
        RSReasoner-Bench-II & 5,800 & 0.017 & 1.021 \\
        \bottomrule
    \end{tabular*}
    \label{tab:attribute_counts}
\end{table}

\paragraph{Per-Query Attribute Counts.}
Table~\ref{tab:attribute_counts} reports the average number of perceptual and inferential attributes per query, where each independent target property is counted as a single span while keeping dependent functional clauses within the same inferential unit. Across existing benchmarks, the per-query attribute counts corroborate the structural patterns in the main paper: EarthReason-rea and GeoSeg-Bench-attr rely primarily on inferential attributes, GRES-rea and GeoSeg-Bench-ref are dominated by perceptual cues, and LaSeRS and GeoSeg-Bench-con stack substantial perceptual and inferential attributes simultaneously. In contrast, RSReasoner-Bench reveals three clear patterns under our Minimal Sufficient Reasoning Cues Principle. First, our subsets achieve substantially cleaner attribute separation, with EP and IP driven by perceptual attributes and EI and II driven by approximately one inferential attribute per query. Second, within perceptual queries, IP contains more perceptual attributes than EP (3.322 vs.\ 2.608), as replacing explicit categories with coarser hypernyms broadens the candidate space and requires richer perceptual descriptions to specify the target. Third, EI and II retain slightly more residual perceptual attributes than the inferential residuals in EP and IP, reflecting cases where inferential attributes alone cannot uniquely disambiguate a target and require auxiliary perceptual grounding (see Figure~\ref{fig:audit_dataset_examples} for qualitative examples).

\paragraph{Human Verification.}
To validate the reliability of our taxonomy audit, human annotators manually annotate a stratified random sample of 400 queries (100 per benchmark) along both category expression and attribute composition, and we compare the model predictions against the human annotations. As shown in Table~\ref{tab:audit_human_verification}, across all 400 verified queries, the audit achieves 99.00\% agreement on category expression, 97.75\% agreement on attribute composition, and 96.75\% joint agreement on both dimensions simultaneously, confirming the reliability of our cross-dataset taxonomy audit.

\begin{table}[!htbp]
    \centering
    \footnotesize
    \setlength{\tabcolsep}{2.0pt}
    \renewcommand{\arraystretch}{1.05}
    \caption{Human verification agreement (\%) on 400 randomly sampled queries (100 per benchmark).}
    \begin{tabular*}{\columnwidth}{@{\extracolsep{\fill}}lrrrr@{}}
        \toprule
        Dataset & Verified & Category (\%) & Attribute (\%) & Joint (\%) \\
        \midrule
        EarthReason & 100 & 99.00 & 97.00 & 96.00 \\
        GRES & 100 & 100.00 & 97.00 & 97.00 \\
        LaSeRS & 100 & 97.00 & 100.00 & 97.00 \\
        GeoSeg-Bench & 100 & 100.00 & 97.00 & 97.00 \\
        \midrule
        Overall & 400 & 99.00 & 97.75 & 96.75 \\
        \bottomrule
    \end{tabular*}
    \label{tab:audit_human_verification}
\end{table}

\section{Additional Experimental Results}
\label{app:exp_results}

In all zero-shot and diagnostic evaluations on RSReasoner-Bench, each model is evaluated from its official checkpoint using its native inference pipeline, receiving only the \(448\times448\) aerial image and the query text. All predicted masks are evaluated at the native ground-truth resolution, with empty predictions retained in the evaluation denominator as an IoU of zero. Following standard protocols~\citep{kazemzadeh2014referitgame,yu2016modeling,liu2023gres,lai2024lisa}, let \(I_i = |P_i \cap G_i|\) and \(U_i = |P_i \cup G_i|\) denote the intersection and union pixel counts between the predicted mask \(P_i\) and ground-truth mask \(G_i\) for target \(i \in \{1,\dots,N\}\). Generalized IoU (gIoU) and cumulative IoU (cIoU) are defined as
\begin{equation}
    \mathrm{gIoU} = \frac{1}{N}\sum_{i=1}^{N}\frac{I_i}{U_i}, \qquad
    \mathrm{cIoU} = \frac{\sum_{i=1}^{N} I_i}{\sum_{i=1}^{N} U_i}.
\end{equation}

\subsection{Additional Zero-Shot Results across Query Types}
\label{app:additional_results}

\begin{table}[!tbp]
    \centering
    \footnotesize
    \setlength{\tabcolsep}{2pt}
    \renewcommand{\arraystretch}{1.00}
    \caption{Zero-shot segmentation performance (gIoU/cIoU, \%) across the four canonical query types on RSReasoner-Bench for area and non-area targets.}
    \label{tab:supp_zeroshot_geometry}
    \begin{tabularx}{\columnwidth}{l*{4}{>{\centering\arraybackslash}X}}
        \toprule
        Method & EP & IP & EI & II \\
        \midrule
        \multicolumn{5}{l}{\textit{Area targets (2,800 targets)}} \\
        LISA-7B & 66.09/70.15 & 58.38/62.12 & 48.75/48.96 & 37.15/38.35 \\
        LISA-13B & 71.32/72.06 & 67.04/68.02 & 57.27/57.50 & 46.78/49.21 \\
        PixelLM-7B & 64.37/68.30 & 54.30/59.24 & 29.78/31.91 & 19.06/20.75 \\
        LISAt-7B & 67.53/71.67 & 53.56/56.62 & 55.02/54.76 & 41.83/43.10 \\
        GeoPixel-7B-RES & 57.02/53.51 & 53.18/47.85 & 41.19/35.82 & 33.85/29.17 \\
        SegEarth-R1 & 27.53/21.57 & 29.08/22.83 & 27.85/21.32 & 27.62/21.36 \\
        UniGeoSeg & 52.93/46.93 & 33.78/22.53 & 47.41/44.77 & 33.79/31.44 \\
        Think2Seg-RS-3B & 69.94/71.11 & 67.18/68.97 & 57.67/56.93 & 47.92/46.91 \\
        Think2Seg-RS-7B & \textbf{78.47}/\textbf{77.99} & \textbf{76.03}/\textbf{76.46} & \textbf{64.10}/\textbf{60.79} & \textbf{54.89}/\textbf{52.33} \\
        \rowcolor{gray!15} Avg. & 61.69/61.48 & 54.73/53.85 & 47.67/45.86 & 38.10/36.96 \\
        \midrule
        \multicolumn{5}{l}{\textit{Non-area targets (3,000 targets)}} \\
        LISA-7B & 48.92/37.89 & 44.80/34.05 & 49.78/36.51 & 48.44/32.34 \\
        LISA-13B & 52.00/31.22 & 26.76/12.27 & 47.38/24.70 & 44.02/20.42 \\
        PixelLM-7B & 36.39/33.18 & 27.98/23.05 & 23.51/22.98 & 20.09/21.61 \\
        LISAt-7B & 32.51/22.44 & 26.08/19.55 & 36.64/24.79 & 36.30/27.24 \\
        GeoPixel-7B-RES & 60.22/52.36 & 58.26/49.77 & 53.41/44.83 & 44.98/41.45 \\
        SegEarth-R1 & 34.96/15.62 & 37.16/15.05 & 26.95/14.50 & 37.88/18.01 \\
        UniGeoSeg & 66.94/\textbf{62.48} & 62.53/\textbf{57.98} & 58.70/\textbf{51.49} & 51.31/\textbf{45.47} \\
        Think2Seg-RS-3B & 68.99/54.05 & 68.36/54.32 & 64.37/48.13 & 61.92/43.24 \\
        Think2Seg-RS-7B & \textbf{73.38}/58.98 & \textbf{72.21}/57.92 & \textbf{68.13}/51.13 & \textbf{64.15}/44.51 \\
        \rowcolor{gray!15} Avg. & 52.70/40.91 & 47.13/36.00 & 47.65/35.45 & 45.45/32.70 \\
        \bottomrule
    \end{tabularx}
\end{table}

\begin{figure*}[!htbp]
    \centering
    \includegraphics[width=0.95\textwidth]{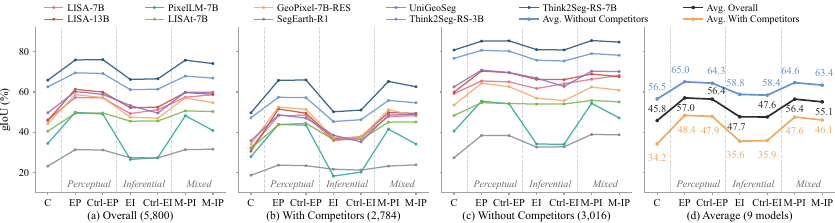}
    \caption{Model-wise zero-shot segmentation trends (gIoU, \%) across canonical explicit queries (EP and EI) and controlled attribute compositions (C, Ctrl-EP, Ctrl-EI, M-PI, and M-IP) on RSReasoner-Bench, showing all nine models on (a) the overall benchmark, (b) targets with same-category competitors, and (c) targets without competitors, alongside (d) the nine-model averages across subsets.}
    \label{fig:supp_attribute_composition}
\end{figure*}

Table~\ref{tab:supp_zeroshot_geometry} complements Table~\ref{tab:zeroshot_quadrants} by breaking down zero-shot performance across target geometries into 2,800 area targets and 3,000 non-area targets. Across both geometry subsets, the core trends observed on the overall benchmark hold on average: inferential queries (EI and II) remain harder than perceptual queries (EP and IP), and implicit expressions remain harder than explicit ones for nearly all models.

Comparing the two geometric subsets, polygonal area targets achieve higher accuracy than elongated non-area targets on perceptual queries due to richer visual appearance, yet suffer a much steeper drop on inferential queries, as disambiguating functional roles among buildings or land parcels is more demanding than grounding connectivity cues along linear road or waterway networks. In addition, while average gIoU and cIoU remain closely aligned on area targets, cIoU is 11.13\%--12.75\% lower than gIoU on non-area targets, because thin linear structures occupy small pixel fractions where over-segmentation on failed predictions heavily inflates the cumulative pixel union.

\subsection{Detailed Results on Attribute Composition}

Table~\ref{tab:supp_attribute_composition} reports the per-model zero-shot segmentation results across the canonical explicit queries (EP and EI) and controlled attribute-composition variants (C, Ctrl-EP, Ctrl-EI, M-PI, and M-IP) on the overall benchmark and both competition subsets, while Figure~\ref{fig:supp_attribute_composition} visualizes the corresponding model-wise gIoU trajectories. Across individual models, the trajectories closely align with the aggregate conclusions in Section~\ref{sec:attribute_composition}: template-aligned queries (Ctrl-EP and Ctrl-EI) closely track their canonical counterparts (EP and EI), and every model consistently achieves higher accuracy on Ctrl-EP than on Ctrl-EI. Relative to category-only references (C), adding perceptual attributes in Ctrl-EP yields substantial gains across all nine models, especially under same-category competition, whereas adding inferential attributes in Ctrl-EI brings only modest improvements for most models and even degrades performance on several models. When perceptual and inferential cues are combined in M-PI and M-IP, all models largely rebound to the level of perceptual-only Ctrl-EP, with most models performing better on perceptual-first M-PI than on inferential-first M-IP.

\begin{table*}[!htbp]
    \centering
    \footnotesize
    \setlength{\tabcolsep}{3.5pt}
    \renewcommand{\arraystretch}{1.0}
    \caption{Zero-shot segmentation performance (gIoU/cIoU, \%) across canonical explicit queries (EP and EI) and controlled attribute-composition variants (C, Ctrl-EP, Ctrl-EI, M-PI, and M-IP) on RSReasoner-Bench.}
    \label{tab:supp_attribute_composition}
    \begin{tabularx}{\textwidth}{l*{7}{>{\centering\arraybackslash}X}}
        \toprule
        Method & C & EP & Ctrl-EP & EI & Ctrl-EI & M-PI & M-IP \\
        \midrule
        \multicolumn{8}{l}{\textit{Overall (5,800 targets)}} \\
        LISA-7B & 46.21\,/\,42.96 & 57.21\,/\,65.37 & 57.04\,/\,64.91 & 49.28\,/\,47.13 & 51.18\,/\,49.65 & 57.39\,/\,61.99 & 58.61\,/\,62.17 \\
        LISA-13B & 45.84\,/\,39.67 & 61.32\,/\,64.01 & 59.90\,/\,62.96 & 52.15\,/\,50.37 & 52.47\,/\,50.95 & 59.69\,/\,62.72 & 58.85\,/\,62.01 \\
        PixelLM-7B & 34.50\,/\,40.45 & 49.90\,/\,62.54 & 49.03\,/\,62.32 & 26.54\,/\,30.55 & 27.44\,/\,32.73 & 48.23\,/\,61.19 & 40.89\,/\,49.44 \\
        LISAt-7B & 40.51\,/\,40.90 & 49.42\,/\,62.91 & 49.57\,/\,63.05 & 45.51\,/\,49.80 & 45.58\,/\,50.68 & 50.62\,/\,62.47 & 50.27\,/\,62.73 \\
        GeoPixel-7B-RES & 44.19\,/\,34.53 & 58.68\,/\,53.34 & 57.21\,/\,52.37 & 47.51\,/\,37.13 & 46.98\,/\,37.34 & 57.03\,/\,50.70 & 54.68\,/\,46.65 \\
        SegEarth-R1 & 23.20\,/\,18.63 & 31.37\,/\,20.15 & 31.21\,/\,20.39 & 27.38\,/\,19.98 & 27.26\,/\,19.65 & 31.40\,/\,20.66 & 31.60\,/\,21.42 \\
        UniGeoSeg & 49.67\,/\,45.76 & 60.18\,/\,48.97 & 58.77\,/\,48.79 & 53.25\,/\,45.65 & 49.52\,/\,43.25 & 59.86\,/\,48.55 & 59.89\,/\,49.48 \\
        Think2Seg-RS-3B & 62.52\,/\,54.78 & 69.45\,/\,68.57 & 69.15\,/\,68.20 & 61.13\,/\,55.66 & 61.36\,/\,56.85 & 67.84\,/\,66.63 & 66.86\,/\,65.15 \\
        Think2Seg-RS-7B & \textbf{65.82}\,/\,\textbf{57.95} & \textbf{75.84}\,/\,\textbf{75.27} & \textbf{76.02}\,/\,\textbf{75.52} & \textbf{66.19}\,/\,\textbf{59.44} & \textbf{66.50}\,/\,\textbf{59.77} & \textbf{75.74}\,/\,\textbf{75.30} & \textbf{74.07}\,/\,\textbf{72.40} \\
        \rowcolor{gray!15} Avg. & 45.83\,/\,41.74 & 57.04\,/\,57.90 & 56.43\,/\,57.61 & 47.66\,/\,43.97 & 47.59\,/\,44.54 & 56.42\,/\,56.69 & 55.08\,/\,54.61 \\
        \midrule
        \multicolumn{8}{l}{\textit{With same-category competitors (2,784 targets)}} \\
        LISA-7B & 32.21\,/\,32.46 & 48.29\,/\,55.95 & 48.42\,/\,55.52 & 35.84\,/\,35.63 & 37.20\,/\,36.72 & 47.76\,/\,52.13 & 48.29\,/\,51.05 \\
        LISA-13B & 30.60\,/\,30.57 & 51.55\,/\,53.93 & 49.44\,/\,52.88 & 36.90\,/\,37.30 & 37.72\,/\,37.86 & 49.75\,/\,52.84 & 49.40\,/\,52.23 \\
        PixelLM-7B & 27.91\,/\,33.19 & 43.93\,/\,55.70 & 43.51\,/\,55.69 & 18.28\,/\,22.46 & 20.25\,/\,25.15 & 41.57\,/\,53.24 & 34.15\,/\,42.11 \\
        LISAt-7B & 32.05\,/\,33.74 & 43.75\,/\,56.69 & 44.48\,/\,57.50 & 36.37\,/\,41.01 & 36.44\,/\,41.82 & 45.02\,/\,56.51 & 45.10\,/\,56.85 \\
        GeoPixel-7B-RES & 34.11\,/\,26.48 & 52.56\,/\,50.06 & 51.35\,/\,49.25 & 37.33\,/\,30.60 & 37.54\,/\,31.19 & 51.21\,/\,48.91 & 47.96\,/\,43.99 \\
        SegEarth-R1 & 18.69\,/\,15.51 & 23.71\,/\,16.84 & 23.45\,/\,16.91 & 21.68\,/\,16.55 & 21.26\,/\,15.95 & 23.25\,/\,16.49 & 23.86\,/\,17.56 \\
        UniGeoSeg & 35.76\,/\,35.19 & 48.62\,/\,39.53 & 47.03\,/\,38.41 & 38.52\,/\,34.83 & 35.23\,/\,33.18 & 48.55\,/\,40.08 & 48.83\,/\,41.65 \\
        Think2Seg-RS-3B & 47.21\,/\,40.09 & 57.34\,/\,55.88 & 57.16\,/\,55.17 & 45.32\,/\,40.63 & 46.23\,/\,42.00 & 55.78\,/\,53.79 & 54.62\,/\,52.45 \\
        Think2Seg-RS-7B & \textbf{49.64}\,/\,\textbf{42.93} & \textbf{65.72}\,/\,\textbf{64.04} & \textbf{65.94}\,/\,\textbf{64.20} & \textbf{50.17}\,/\,\textbf{44.88} & \textbf{51.00}\,/\,\textbf{45.14} & \textbf{65.20}\,/\,\textbf{63.61} & \textbf{62.57}\,/\,\textbf{59.86} \\
        \rowcolor{gray!15} Avg. & 34.24\,/\,32.24 & 48.39\,/\,49.85 & 47.87\,/\,49.51 & 35.60\,/\,33.77 & 35.88\,/\,34.33 & 47.56\,/\,48.62 & 46.09\,/\,46.42 \\
        \midrule
        \multicolumn{8}{l}{\textit{Without same-category competitors (3,016 targets)}} \\
        LISA-7B & 59.14\,/\,58.64 & 65.44\,/\,74.80 & 64.99\,/\,74.36 & 61.69\,/\,60.43 & 64.08\,/\,64.94 & 66.28\,/\,71.87 & 68.15\,/\,73.53 \\
        LISA-13B & 59.90\,/\,52.96 & 70.35\,/\,74.65 & 69.55\,/\,73.60 & 66.23\,/\,66.48 & 66.09\,/\,67.22 & 68.86\,/\,73.00 & 67.58\,/\,72.24 \\
        PixelLM-7B & 40.57\,/\,50.58 & 55.41\,/\,69.09 & 54.12\,/\,68.70 & 34.16\,/\,38.04 & 34.08\,/\,39.91 & 54.38\,/\,68.59 & 47.11\,/\,55.95 \\
        LISAt-7B & 48.32\,/\,50.13 & 54.65\,/\,68.48 & 54.27\,/\,68.03 & 53.96\,/\,58.80 & 54.02\,/\,59.69 & 55.79\,/\,67.80 & 55.04\,/\,68.13 \\
        GeoPixel-7B-RES & 53.49\,/\,43.17 & 64.32\,/\,56.32 & 62.61\,/\,55.20 & 56.91\,/\,43.41 & 55.68\,/\,43.25 & 62.40\,/\,52.31 & 60.87\,/\,49.02 \\
        SegEarth-R1 & 27.37\,/\,21.51 & 38.44\,/\,23.11 & 38.37\,/\,23.52 & 32.65\,/\,23.17 & 32.80\,/\,23.09 & 38.93\,/\,24.49 & 38.74\,/\,24.96 \\
        UniGeoSeg & 62.51\,/\,57.78 & 70.84\,/\,58.18 & 69.60\,/\,59.00 & 66.85\,/\,57.29 & 62.71\,/\,53.98 & 70.30\,/\,56.90 & 70.09\,/\,57.42 \\
        Think2Seg-RS-3B & 76.65\,/\,74.34 & 80.62\,/\,81.38 & 80.22\,/\,81.24 & 75.73\,/\,72.55 & 75.33\,/\,73.79 & 78.99\,/\,79.53 & 78.17\,/\,77.96 \\
        Think2Seg-RS-7B & \textbf{80.76}\,/\,\textbf{78.84} & \textbf{85.17}\,/\,\textbf{87.06} & \textbf{85.32}\,/\,\textbf{87.33} & \textbf{80.97}\,/\,\textbf{77.49} & \textbf{80.80}\,/\,\textbf{78.11} & \textbf{85.47}\,/\,\textbf{87.53} & \textbf{84.69}\,/\,\textbf{86.00} \\
        \rowcolor{gray!15} Avg. & 56.52\,/\,54.22 & 65.03\,/\,65.90 & 64.34\,/\,65.66 & 58.79\,/\,55.30 & 58.40\,/\,56.00 & 64.60\,/\,64.67 & 63.38\,/\,62.80 \\
        \bottomrule
    \end{tabularx}
\end{table*}

\subsection{Detailed Results on Category Expression}

Table~\ref{tab:supp_category_expression} complements Table~\ref{tab:category_expression} with the absolute scores and performance changes of the category-replacement variants across the overall benchmark and both competition subsets. Across all three groupings, implicitizing explicit categories (EP-Imp and EI-Imp) consistently degrades performance, whereas explicitizing hypernyms (IP-Exp and II-Exp) consistently improves performance for all models except SegEarth-R1. Moreover, even after restoring explicit categories, II-Exp remains lower in average accuracy than implicitized perceptual queries in EP-Imp across all subsets, indicating that category expression alone cannot fully bridge the gap between perceptual and inferential grounding.

Comparing the two competition subsets further reveals a distinct contrast between perceptual and inferential queries due to how attributes interact with category semantics. For perceptual queries, visual attributes such as color, shape, and location are category-agnostic, so removing or restoring the category name simply broadens or narrows the candidate pool and yields broadly symmetric changes in both competition subsets. For inferential queries, however, explicit categories define the semantic comparison domain required for fine-grained, category-conditioned reasoning cues (e.g., traffic merging roles among roads or cultivation states among farmland), whereas implicit queries rely on broader, cross-category functional descriptions. Consequently, replacing the category in EI with a hypernym (EI-Imp) expands the candidate space and weakens the specificity of category-conditioned cues across heterogeneous objects, causing substantial drops on both competing (\(-4.56\%\) average gIoU) and non-competing (\(-4.09\%\)) targets, which also empirically validates our uniqueness check and minimal functional refinement when constructing paired II supervision in Section~\ref{app:paired_query}. Conversely, when restoring explicit categories to II (II-Exp), the broad functional cues in II already suffice to isolate the unique target on non-competing scenes and thus gain only \(+1.57\%\), whereas under same-category competition, restoring the category name anchors the intra-class comparison domain and yields a much larger gain of \(+4.27\%\).

\begin{table}[!htbp]
    \centering
    \footnotesize
    \setlength{\tabcolsep}{3.5pt}
    \renewcommand{\arraystretch}{0.95}
    \caption{Fine-tuning hyperparameters for the five models on the RSReasoner training split.}
    \label{tab:supp_finetuning_setup}
    \begin{tabularx}{\columnwidth}{lcccc>{\centering\arraybackslash}X}
        \toprule
        Method & Epochs & Steps & LR & LR Schedule & Eff.\ Batch \\
        \midrule
        LISA-7B & 5 & 14,870 & \(3\times10^{-4}\) & Linear & 64 \\
        PixelLM-7B & 5 & 23,790 & \(3\times10^{-4}\) & Linear & 40 \\
        LISAt-7B & 5 & 9,915 & \(3\times10^{-4}\) & Cosine & 96 \\
        GeoPixel-7B-RES & 3 & 7,137 & \(3\times10^{-4}\) & Cosine & 80 \\
        SegEarth-R1 & 5 & 14,870 & \(1\times10^{-4}\) & Cosine & 64 \\
        \bottomrule
    \end{tabularx}
\end{table}

\begin{table*}[p]
    \centering
    \footnotesize
    \setlength{\tabcolsep}{2.5pt}
    \renewcommand{\arraystretch}{1.0}
    \definecolor{gaincolor}{RGB}{33,102,172}  %
    \definecolor{dropcolor}{RGB}{178,24,43}   %
    \newcommand{\up}[1]{\textcolor{gaincolor}{#1}}
    \newcommand{\dn}[1]{\textcolor{dropcolor}{#1}}
    \caption{Complete zero-shot segmentation performance (gIoU\,/\,cIoU, \%) and performance changes (\(\Delta\text{gIoU}\,/\,\Delta\text{cIoU}\), \%; \textcolor{gaincolor}{blue}: gain, \textcolor{dropcolor}{red}: drop) under category-expression replacement on RSReasoner-Bench relative to the canonical queries in Table~\ref{tab:zeroshot_quadrants}.}
    \label{tab:supp_category_expression}
    \begin{tabularx}{\textwidth}{l*{4}{>{\centering\arraybackslash}X}|*{4}{>{\centering\arraybackslash}X}}
        \toprule
        Method & EP-Imp & IP-Exp & EI-Imp & II-Exp & \(\Delta\)EP-Imp & \(\Delta\)IP-Exp & \(\Delta\)EI-Imp & \(\Delta\)II-Exp \\
        \midrule
        \multicolumn{9}{l}{\textit{Overall (5,800 targets)}} \\
        LISA-7B & 56.70\,/\,57.95 & 56.41\,/\,64.45 & 46.24\,/\,40.50 & 45.14\,/\,40.44 & \dn{-0.51}\,/\,\dn{-7.42} & \up{+5.06}\,/\,\up{+6.72} & \dn{-3.04}\,/\,\dn{-6.63} & \up{+2.15}\,/\,\up{+3.07} \\
        LISA-13B & 56.67\,/\,57.16 & 59.66\,/\,64.22 & 46.89\,/\,44.51 & 48.12\,/\,45.82 & \dn{-4.65}\,/\,\dn{-6.85} & \up{+13.45}\,/\,\up{+9.69} & \dn{-5.26}\,/\,\dn{-5.86} & \up{+2.76}\,/\,\up{+3.29} \\
        PixelLM-7B & 41.25\,/\,50.42 & 49.13\,/\,60.98 & 22.36\,/\,25.74 & 21.66\,/\,21.63 & \dn{-8.65}\,/\,\dn{-12.12} & \up{+8.44}\,/\,\up{+8.15} & \dn{-4.18}\,/\,\dn{-4.81} & \up{+2.06}\,/\,\up{+0.75} \\
        LISAt-7B & 42.81\,/\,55.90 & 46.41\,/\,57.64 & 39.95\,/\,39.94 & 42.08\,/\,43.99 & \dn{-6.61}\,/\,\dn{-7.00} & \up{+7.06}\,/\,\up{+7.15} & \dn{-5.56}\,/\,\dn{-9.86} & \up{+3.12}\,/\,\up{+3.30} \\
        GeoPixel-7B-RES & 54.35\,/\,48.43 & 59.21\,/\,53.77 & 42.68\,/\,33.10 & 44.30\,/\,32.27 & \dn{-4.33}\,/\,\dn{-4.91} & \up{+3.40}\,/\,\up{+5.64} & \dn{-4.83}\,/\,\dn{-4.03} & \up{+4.69}\,/\,\up{+1.51} \\
        SegEarth-R1 & 33.07\,/\,21.04 & 32.41\,/\,20.33 & 32.38\,/\,20.47 & 27.71\,/\,20.18 & \up{+1.70}\,/\,\up{+0.89} & \dn{-0.85}\,/\,\dn{-0.47} & \up{+5.00}\,/\,\up{+0.49} & \dn{-5.22}\,/\,\dn{-0.45} \\
        UniGeoSeg & 48.75\,/\,28.54 & 59.34\,/\,48.31 & 43.50\,/\,35.75 & 48.48\,/\,41.63 & \dn{-11.43}\,/\,\dn{-20.43} & \up{+10.69}\,/\,\up{+21.25} & \dn{-9.75}\,/\,\dn{-9.90} & \up{+5.63}\,/\,\up{+8.33} \\
        Think2Seg-RS-3B & 65.17\,/\,63.07 & 70.87\,/\,70.68 & 56.29\,/\,48.54 & 59.90\,/\,52.86 & \dn{-4.28}\,/\,\dn{-5.51} & \up{+3.08}\,/\,\up{+3.83} & \dn{-4.85}\,/\,\dn{-7.12} & \up{+4.73}\,/\,\up{+6.51} \\
        Think2Seg-RS-7B & \textbf{71.75}\,/\,\textbf{69.85} & \textbf{76.53}\,/\,\textbf{76.66} & \textbf{59.82}\,/\,\textbf{52.03} & \textbf{65.56}\,/\,\textbf{57.93} & \dn{-4.09}\,/\,\dn{-5.42} & \up{+2.48}\,/\,\up{+2.89} & \dn{-6.36}\,/\,\dn{-7.41} & \up{+5.88}\,/\,\up{+6.78} \\
        \rowcolor{gray!15} Avg. & 52.28\,/\,50.26 & 56.66\,/\,57.45 & 43.35\,/\,37.84 & 44.77\,/\,39.64 & \dn{-4.76}\,/\,\dn{-7.64} & \up{+5.87}\,/\,\up{+7.21} & \dn{-4.31}\,/\,\dn{-6.13} & \up{+2.87}\,/\,\up{+3.68} \\
        \midrule
        \multicolumn{9}{l}{\textit{With same-category competitors (2,784 targets)}} \\
        LISA-7B & 47.97\,/\,52.08 & 48.35\,/\,57.67 & 32.54\,/\,30.97 & 33.04\,/\,32.30 & \dn{-0.32}\,/\,\dn{-3.87} & \up{+2.73}\,/\,\up{+1.90} & \dn{-3.30}\,/\,\dn{-4.66} & \up{+2.55}\,/\,\up{+3.22} \\
        LISA-13B & 48.30\,/\,51.07 & 51.32\,/\,56.48 & 31.84\,/\,32.99 & 34.85\,/\,35.13 & \dn{-3.24}\,/\,\dn{-2.86} & \up{+8.25}\,/\,\up{+4.07} & \dn{-5.06}\,/\,\dn{-4.30} & \up{+3.75}\,/\,\up{+3.08} \\
        PixelLM-7B & 36.09\,/\,47.20 & 43.52\,/\,55.58 & 14.86\,/\,18.81 & 14.54\,/\,17.98 & \dn{-7.84}\,/\,\dn{-8.51} & \up{+6.87}\,/\,\up{+5.72} & \dn{-3.42}\,/\,\dn{-3.65} & \up{+2.08}\,/\,\up{+2.59} \\
        LISAt-7B & 37.32\,/\,48.86 & 42.35\,/\,56.89 & 30.43\,/\,33.52 & 33.49\,/\,37.32 & \dn{-6.43}\,/\,\dn{-7.83} & \up{+6.29}\,/\,\up{+5.92} & \dn{-5.93}\,/\,\dn{-7.49} & \up{+4.85}\,/\,\up{+5.20} \\
        GeoPixel-7B-RES & 47.84\,/\,46.34 & 53.07\,/\,51.31 & 33.33\,/\,28.93 & 34.85\,/\,27.89 & \dn{-4.72}\,/\,\dn{-3.72} & \up{+2.64}\,/\,\up{+2.77} & \dn{-4.00}\,/\,\dn{-1.68} & \up{+4.30}\,/\,\up{+2.66} \\
        SegEarth-R1 & 24.63\,/\,17.12 & 24.18\,/\,17.14 & 24.55\,/\,17.03 & 21.71\,/\,16.26 & \up{+0.92}\,/\,\up{+0.28} & \dn{-0.64}\,/\,\dn{-0.05} & \up{+2.87}\,/\,\up{+0.48} & \dn{-2.59}\,/\,\dn{-1.19} \\
        UniGeoSeg & 36.85\,/\,21.86 & 47.50\,/\,38.63 & 30.04\,/\,27.33 & 32.58\,/\,29.34 & \dn{-11.78}\,/\,\dn{-17.67} & \up{+10.84}\,/\,\up{+16.58} & \dn{-8.49}\,/\,\dn{-7.49} & \up{+7.83}\,/\,\up{+11.27} \\
        Think2Seg-RS-3B & 51.54\,/\,49.75 & 60.27\,/\,60.26 & 39.40\,/\,34.84 & 44.20\,/\,38.07 & \dn{-5.81}\,/\,\dn{-6.13} & \up{+3.19}\,/\,\up{+2.82} & \dn{-5.92}\,/\,\dn{-5.79} & \up{+7.46}\,/\,\up{+7.95} \\
        Think2Seg-RS-7B & \textbf{60.69}\,/\,\textbf{58.63} & \textbf{67.05}\,/\,\textbf{66.18} & \textbf{42.37}\,/\,\textbf{37.79} & \textbf{49.95}\,/\,\textbf{44.43} & \dn{-5.04}\,/\,\dn{-5.41} & \up{+2.28}\,/\,\up{+0.42} & \dn{-7.79}\,/\,\dn{-7.09} & \up{+8.22}\,/\,\up{+7.76} \\
        \rowcolor{gray!15} Avg. & 43.47\,/\,43.66 & 48.62\,/\,51.13 & 31.04\,/\,29.13 & 33.25\,/\,30.97 & \dn{-4.92}\,/\,\dn{-6.19} & \up{+4.72}\,/\,\up{+4.46} & \dn{-4.56}\,/\,\dn{-4.63} & \up{+4.27}\,/\,\up{+4.73} \\
        \midrule
        \multicolumn{9}{l}{\textit{Without same-category competitors (3,016 targets)}} \\
        LISA-7B & 64.76\,/\,63.37 & 63.85\,/\,71.01 & 58.89\,/\,51.20 & 56.31\,/\,48.86 & \dn{-0.69}\,/\,\dn{-11.42} & \up{+7.21}\,/\,\up{+11.55} & \dn{-2.80}\,/\,\dn{-9.23} & \up{+1.78}\,/\,\up{+3.34} \\
        LISA-13B & 64.39\,/\,62.95 & 67.35\,/\,71.88 & 60.79\,/\,58.54 & 60.36\,/\,57.73 & \dn{-5.96}\,/\,\dn{-11.70} & \up{+18.25}\,/\,\up{+15.49} & \dn{-5.44}\,/\,\dn{-7.95} & \up{+1.86}\,/\,\up{+3.69} \\
        PixelLM-7B & 46.01\,/\,53.32 & 54.31\,/\,66.14 & 29.29\,/\,32.19 & 28.23\,/\,25.01 & \dn{-9.39}\,/\,\dn{-15.77} & \up{+9.89}\,/\,\up{+10.70} & \dn{-4.88}\,/\,\dn{-5.84} & \up{+2.04}\,/\,\dn{-0.96} \\
        LISAt-7B & 47.87\,/\,62.37 & 50.16\,/\,58.30 & 48.74\,/\,46.56 & 50.02\,/\,50.60 & \dn{-6.78}\,/\,\dn{-6.11} & \up{+7.76}\,/\,\up{+8.21} & \dn{-5.21}\,/\,\dn{-12.24} & \up{+1.52}\,/\,\up{+1.20} \\
        GeoPixel-7B-RES & 60.36\,/\,50.29 & 64.88\,/\,56.02 & 51.31\,/\,37.03 & 53.03\,/\,36.41 & \dn{-3.96}\,/\,\dn{-6.02} & \up{+4.10}\,/\,\up{+8.26} & \dn{-5.60}\,/\,\dn{-6.39} & \up{+5.05}\,/\,\up{+0.50} \\
        SegEarth-R1 & 40.86\,/\,24.52 & 40.00\,/\,23.16 & 39.61\,/\,23.63 & 33.25\,/\,23.84 & \up{+2.42}\,/\,\up{+1.41} & \dn{-1.06}\,/\,\dn{-0.75} & \up{+6.97}\,/\,\up{+0.46} & \dn{-7.64}\,/\,\up{+0.37} \\
        UniGeoSeg & 59.73\,/\,34.53 & 70.27\,/\,57.79 & 55.93\,/\,44.32 & 63.15\,/\,54.45 & \dn{-11.11}\,/\,\dn{-23.65} & \up{+10.54}\,/\,\up{+26.30} & \dn{-10.92}\,/\,\dn{-12.98} & \up{+3.59}\,/\,\up{+6.22} \\
        Think2Seg-RS-3B & 77.76\,/\,75.90 & 80.64\,/\,81.04 & 71.87\,/\,63.93 & 74.38\,/\,68.94 & \dn{-2.86}\,/\,\dn{-5.48} & \up{+2.97}\,/\,\up{+5.32} & \dn{-3.86}\,/\,\dn{-8.62} & \up{+2.22}\,/\,\up{+5.30} \\
        Think2Seg-RS-7B & \textbf{81.96}\,/\,\textbf{80.83} & \textbf{85.28}\,/\,\textbf{87.53} & \textbf{75.93}\,/\,\textbf{68.77} & \textbf{79.97}\,/\,\textbf{74.08} & \dn{-3.21}\,/\,\dn{-6.23} & \up{+2.66}\,/\,\up{+6.06} & \dn{-5.04}\,/\,\dn{-8.71} & \up{+3.72}\,/\,\up{+6.42} \\
        \rowcolor{gray!15} Avg. & 60.41\,/\,56.45 & 64.08\,/\,63.65 & 54.71\,/\,47.35 & 55.41\,/\,48.88 & \dn{-4.62}\,/\,\dn{-9.44} & \up{+6.92}\,/\,\up{+10.13} & \dn{-4.09}\,/\,\dn{-7.94} & \up{+1.57}\,/\,\up{+2.90} \\
        \bottomrule
    \end{tabularx}
\end{table*}

\subsection{Fine-Tuning Details and Complete Results}

\paragraph{Training and Model Selection Setup.}
To evaluate the effectiveness of RSReasoner as training supervision, we continually fine-tune five representative models (LISA-7B, PixelLM-7B, LISAt-7B, GeoPixel-7B-RES, and SegEarth-R1) from their official checkpoints on the RSReasoner training split. Each model is trained under its native architecture, prompt format, and optimization objective, with GeoPixel-7B-RES trained under its high-resolution reasoning configuration (\texttt{hd\_num}=9). Table~\ref{tab:supp_finetuning_setup} summarizes the fine-tuning hyperparameters for all five models. Checkpoints are evaluated on the validation split (5,660 targets with 11,320 EI and II queries), and the checkpoint achieving the highest mean validation gIoU across EI and II is selected for final evaluation on RSReasoner-Bench.

\paragraph{Complete Fine-Tuning Results.}
Table~\ref{tab:supp_finetuning} complements Table~\ref{tab:finetuning} by reporting the validation scores of the selected checkpoints alongside the complete fine-tuned segmentation accuracy on RSReasoner-Bench across the overall benchmark and both competition subsets. Across all five models and both subsets, continual fine-tuning on paired EI and II supervision consistently elevates both in-domain inferential accuracy and cross-type perceptual performance, with GeoPixel-7B-RES achieving the strongest results across all conditions. Moreover, paired EI and II supervision resolves the zero-shot implicit category bias of SegEarth-R1 across all subsets, restoring the canonical ordering where explicit queries outperform their implicit counterparts.

\begin{table*}[p]
    \centering
    \footnotesize
    \setlength{\tabcolsep}{2pt}
    \renewcommand{\arraystretch}{1.0}
    \caption{Segmentation performance (gIoU/cIoU, \%) of the fine-tuned models on the RSReasoner validation split and across the four canonical query types on RSReasoner-Bench.}
    \label{tab:supp_finetuning}

    \begin{minipage}[t]{0.49\textwidth}
        \vspace{0pt}
        \begin{tabularx}{\linewidth}{l*{4}{>{\centering\arraybackslash}X}}
            \toprule
            Method & EP & IP & EI & II \\
            \midrule
            \multicolumn{5}{l}{\textit{RSReasoner validation split (5,660 targets)}} \\
            LISA-7B & -- & -- & 83.17/83.50 & 82.89/83.36 \\
            PixelLM-7B & -- & -- & 78.42/80.11 & 77.65/79.14 \\
            LISAt-7B & -- & -- & 82.99/82.92 & 82.54/82.13 \\
            GeoPixel-7B-RES & -- & -- & \textbf{87.55}/\textbf{87.20} & \textbf{86.97}/\textbf{86.20} \\
            SegEarth-R1 & -- & -- & 81.61/81.37 & 80.93/80.71 \\
            \rowcolor{gray!15} Avg. & -- & -- & 82.75/83.02 & 82.20/82.31 \\
            \midrule
            \multicolumn{5}{l}{\textit{Overall (5,800 targets)}} \\
            LISA-7B & 83.20/82.51 & 68.39/66.31 & 82.46/79.33 & 79.94/75.94 \\
            PixelLM-7B & 82.27/86.83 & 77.06/77.33 & 75.36/73.65 & 73.69/70.49 \\
            LISAt-7B & 83.85/83.20 & 70.58/60.79 & 82.00/79.71 & 78.89/73.68 \\
            GeoPixel-7B-RES & \textbf{90.75}/\textbf{92.33} & \textbf{87.83}/\textbf{87.67} & \textbf{85.35}/\textbf{83.78} & \textbf{82.17}/\textbf{77.56} \\
            SegEarth-R1 & 77.36/73.59 & 48.73/44.06 & 80.36/77.75 & 76.64/71.05 \\
            \rowcolor{gray!15} Avg. & 83.49/83.69 & 70.52/67.23 & 81.11/78.84 & 78.27/73.74 \\
            \bottomrule
        \end{tabularx}
    \end{minipage}\hfill
    \begin{minipage}[t]{0.49\textwidth}
        \vspace{0pt}
        \begin{tabularx}{\linewidth}{l*{4}{>{\centering\arraybackslash}X}}
            \toprule
            Method & EP & IP & EI & II \\
            \midrule
            \multicolumn{5}{l}{\textit{With same-category competitors (2,784 targets)}} \\
            LISA-7B & 75.14/73.99 & 62.78/64.40 & 72.85/68.17 & 67.94/62.40 \\
            PixelLM-7B & 77.01/80.81 & 72.24/74.56 & 64.55/60.04 & 60.40/55.28 \\
            LISAt-7B & 76.95/75.74 & 65.19/61.67 & 72.37/68.82 & 66.15/59.28 \\
            GeoPixel-7B-RES & \textbf{86.82}/\textbf{88.16} & \textbf{83.60}/\textbf{85.27} & \textbf{77.31}/\textbf{74.79} & \textbf{70.22}/\textbf{63.77} \\
            SegEarth-R1 & 66.99/59.61 & 47.72/44.01 & 71.36/66.30 & 64.26/55.37 \\
            \rowcolor{gray!15} Avg. & 76.58/75.66 & 66.31/65.98 & 71.69/67.62 & 65.79/59.22 \\
            \midrule
            \multicolumn{5}{l}{\textit{Without same-category competitors (3,016 targets)}} \\
            LISA-7B & 90.64/90.29 & 73.57/67.93 & 91.33/90.03 & 91.02/88.81 \\
            PixelLM-7B & 87.12/92.27 & 81.50/79.75 & 85.33/87.29 & 85.96/85.66 \\
            LISAt-7B & 90.21/89.92 & 75.56/60.01 & 90.90/90.12 & 90.65/87.90 \\
            GeoPixel-7B-RES & \textbf{94.37}/\textbf{96.01} & \textbf{91.74}/\textbf{89.74} & \textbf{92.78}/\textbf{92.12} & \textbf{93.20}/\textbf{90.82} \\
            SegEarth-R1 & 86.93/87.26 & 49.66/44.11 & 88.68/88.79 & 88.07/86.64 \\
            \rowcolor{gray!15} Avg. & 89.85/91.15 & 74.41/68.31 & 89.80/89.67 & 89.78/87.97 \\
            \bottomrule
        \end{tabularx}
    \end{minipage}
\end{table*}

\clearpage

\section{Qualitative Examples}
\label{app:qualitative_results}

\subsection{RSReasoner}
\begin{figure*}[p]
    \centering
    \includegraphics[width=0.98\textwidth]{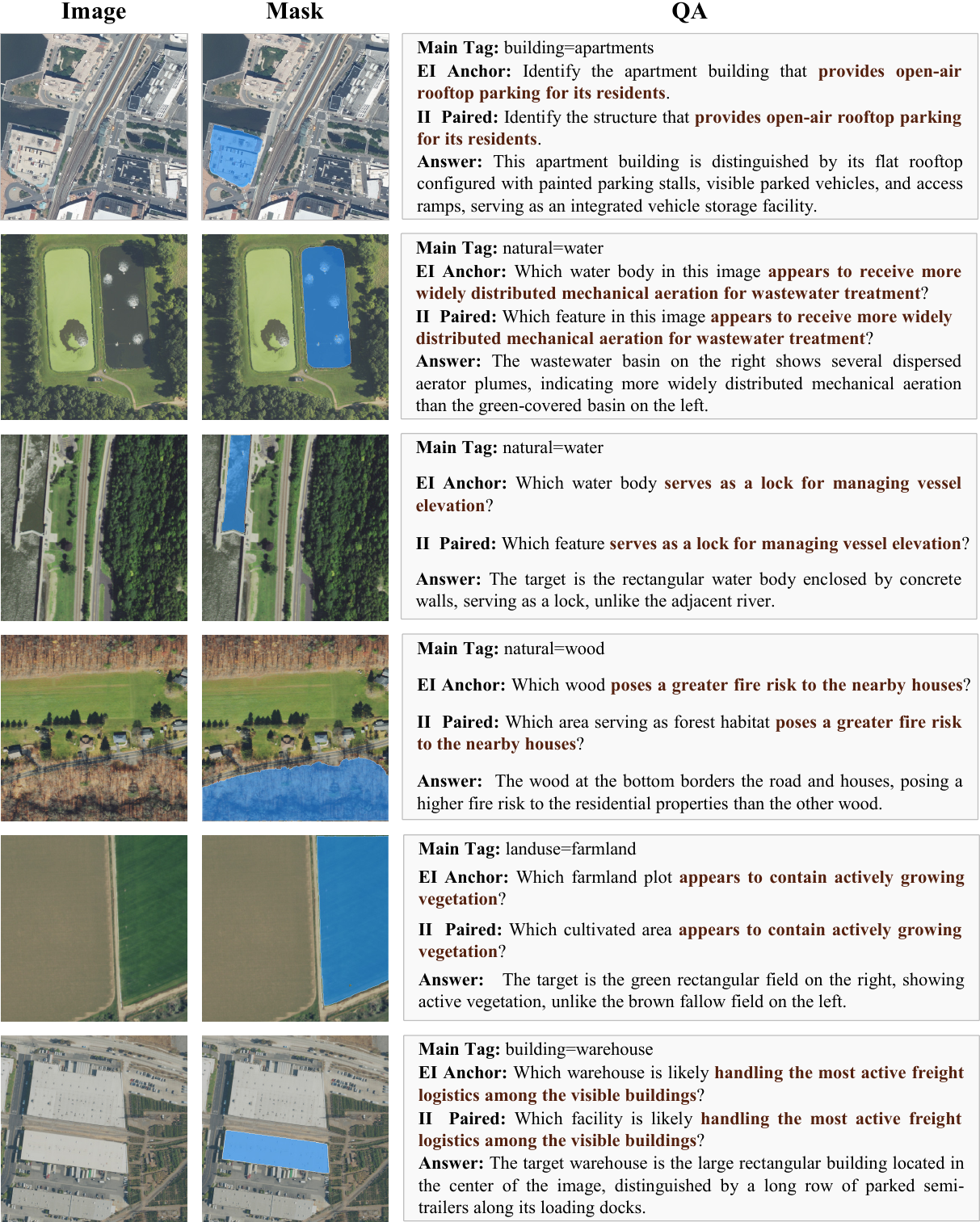}
    \caption{Representative EI-anchored examples with area targets from RSReasoner.}
    \label{fig:example_ei_area}
\end{figure*}

\begin{figure*}[p]
    \centering
    \includegraphics[width=0.98\textwidth]{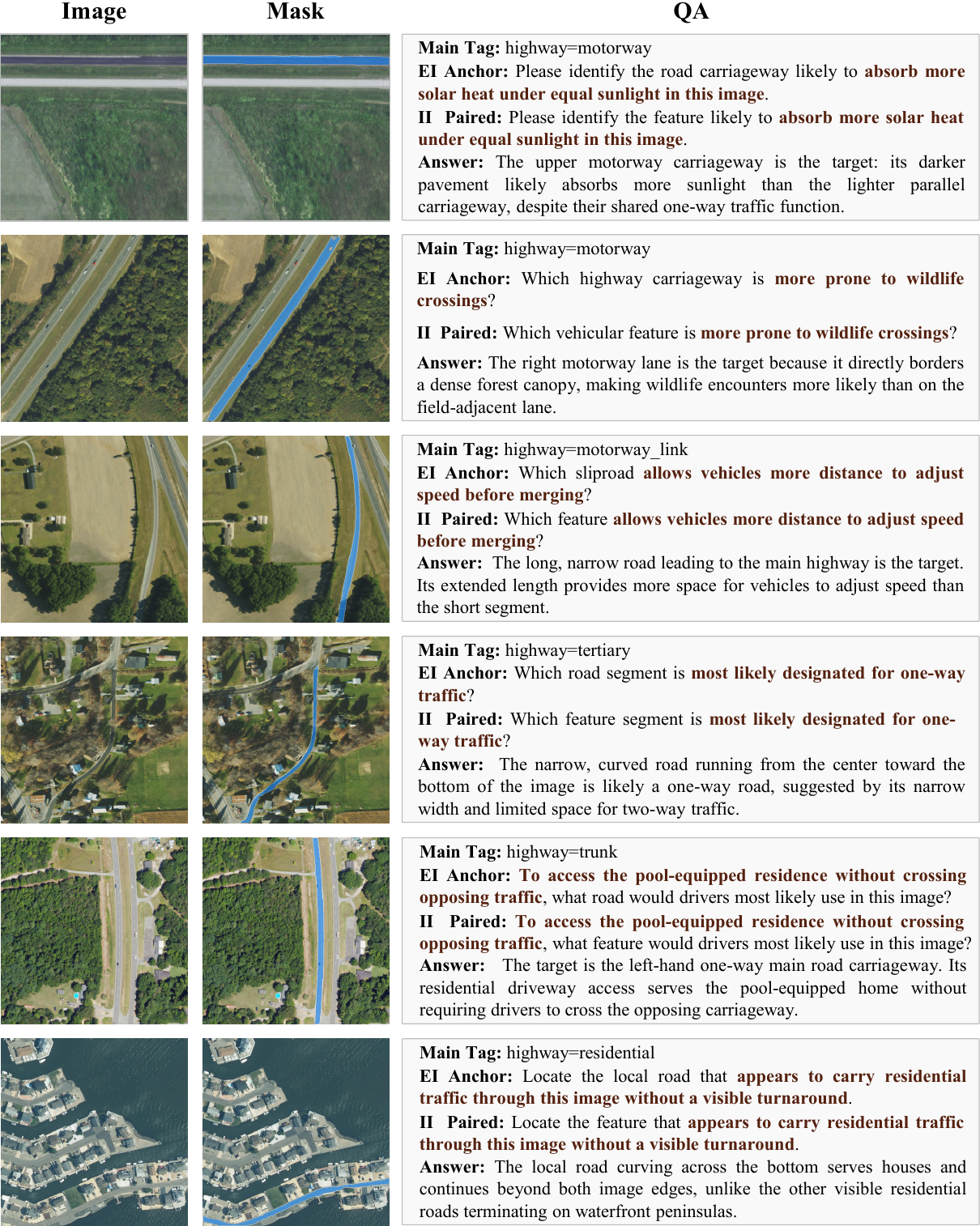}
    \caption{Representative EI-anchored examples with non-area targets from RSReasoner.}
    \label{fig:example_ei_non_area}
\end{figure*}

\begin{figure*}[p]
    \centering
    \includegraphics[width=0.98\textwidth]{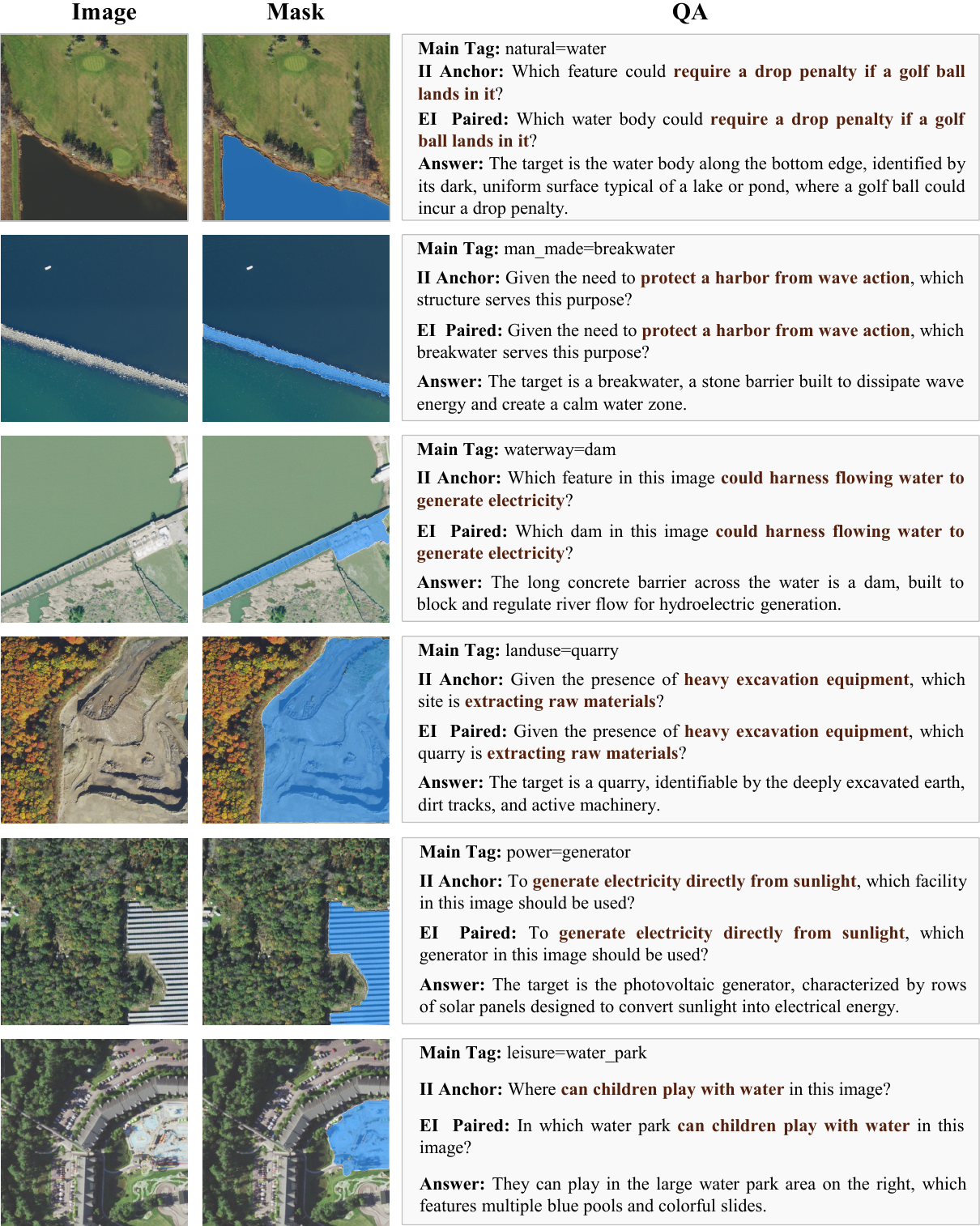}
    \caption{Representative II-anchored examples with area targets from RSReasoner.}
    \label{fig:example_ii_area}
\end{figure*}

\begin{figure*}[p]
    \centering
    \includegraphics[width=0.98\textwidth]{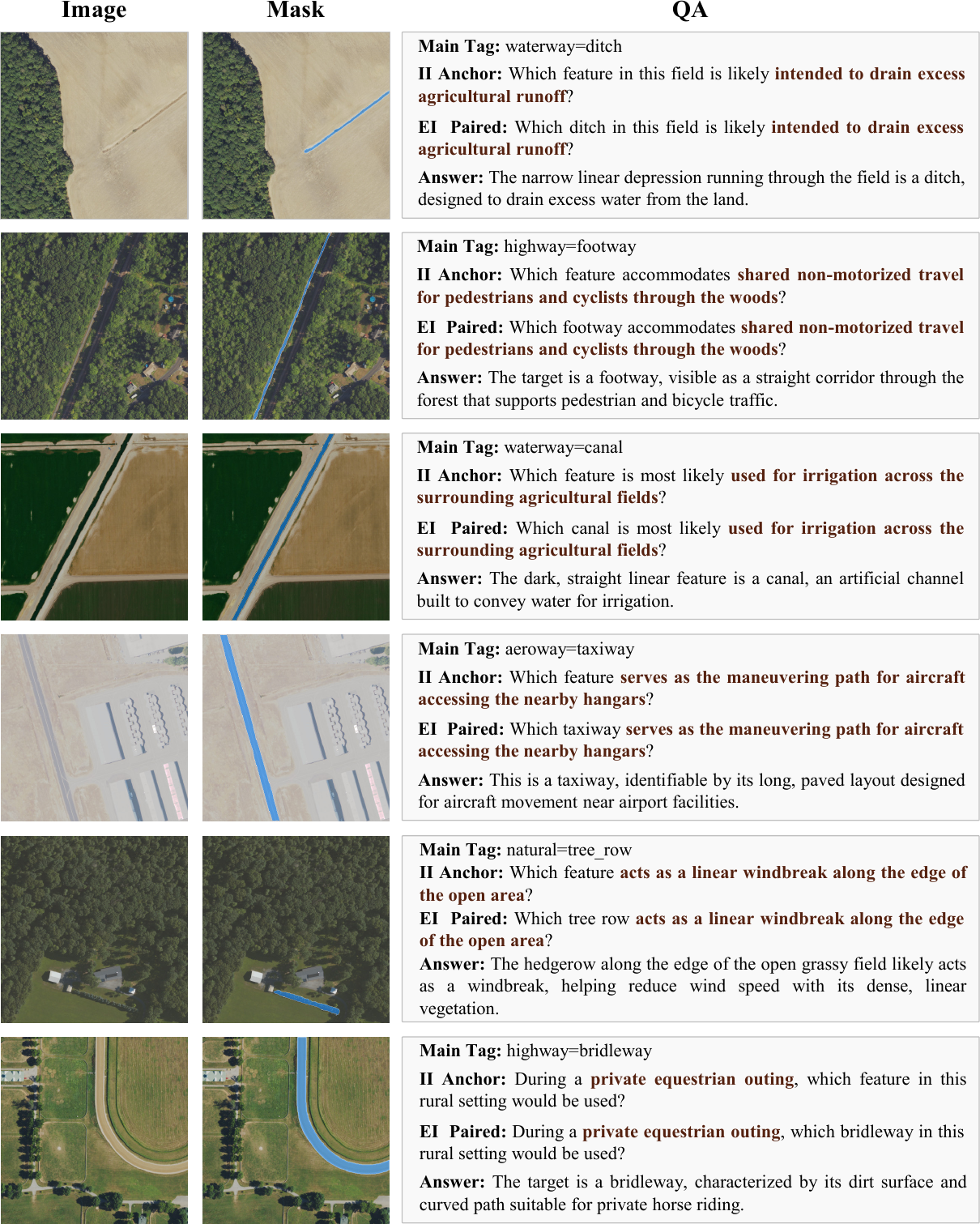}
    \caption{Representative II-anchored examples with non-area targets from RSReasoner.}
    \label{fig:example_ii_non_area}
\end{figure*}

Figures~\ref{fig:example_ei_area}--\ref{fig:example_ii_non_area} showcase representative RSReasoner samples across both anchor regimes and target geometries.

\subsection{RSReasoner-Bench}
\begin{figure*}[p]
    \centering
    \includegraphics[width=0.95\textwidth]{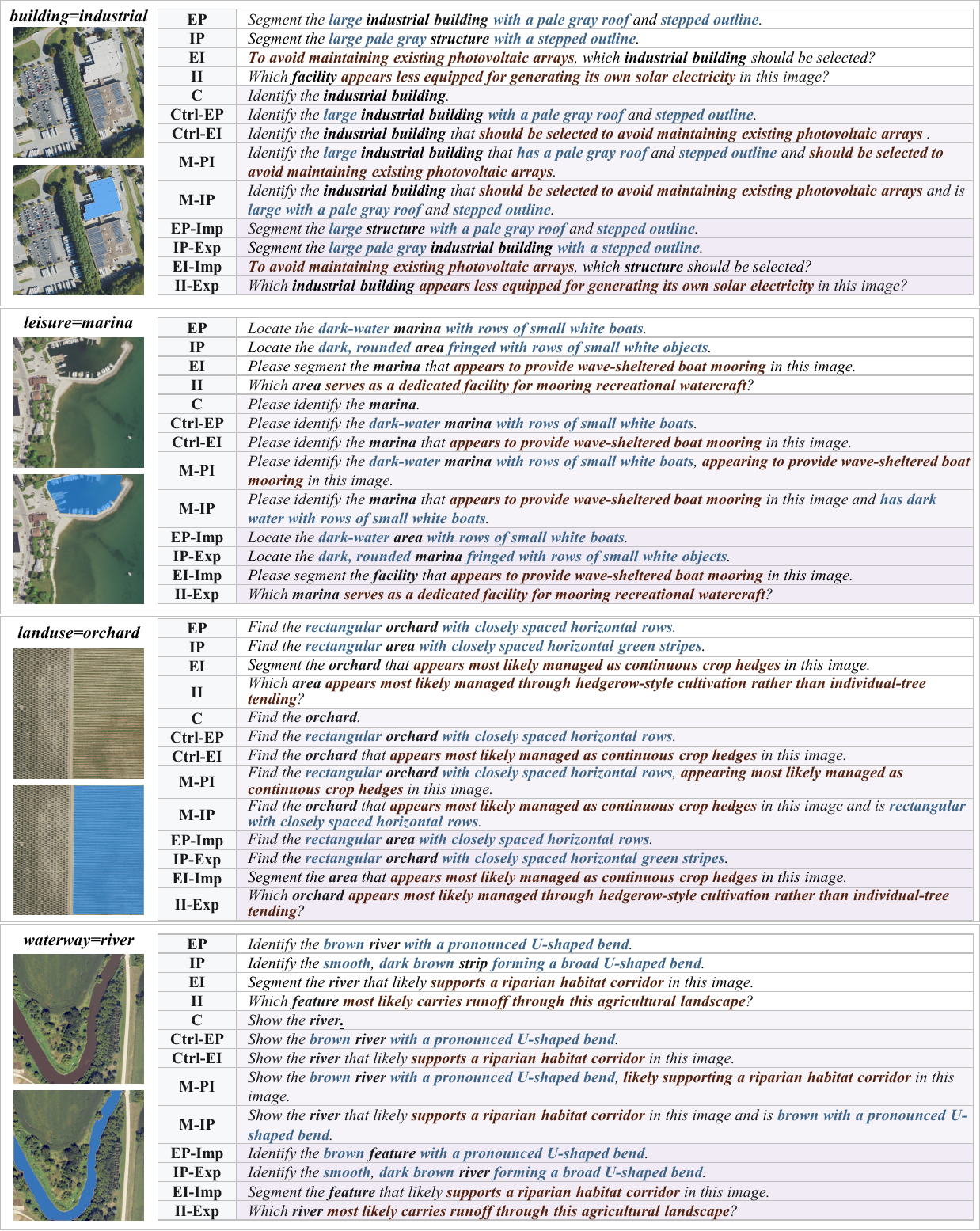}
\caption{Representative RSReasoner-Bench examples across all 13 query conditions. Each block pairs the image and target mask with its 13 queries, highlighting category/hypernym terms in bold black, perceptual cues in blue, and inferential cues in brown.}
    \label{fig:example_bench}
\end{figure*}

Figure~\ref{fig:example_bench} illustrates four RSReasoner-Bench targets spanning both competition regimes and target geometries, each pairing a verified ground-truth mask with all 13 query conditions across the canonical, attribute-composition, and category-replacement settings.

\subsection{Cross-Benchmark Taxonomy Audit}
\begin{figure*}[p]
    \centering
    \includegraphics[width=\textwidth]{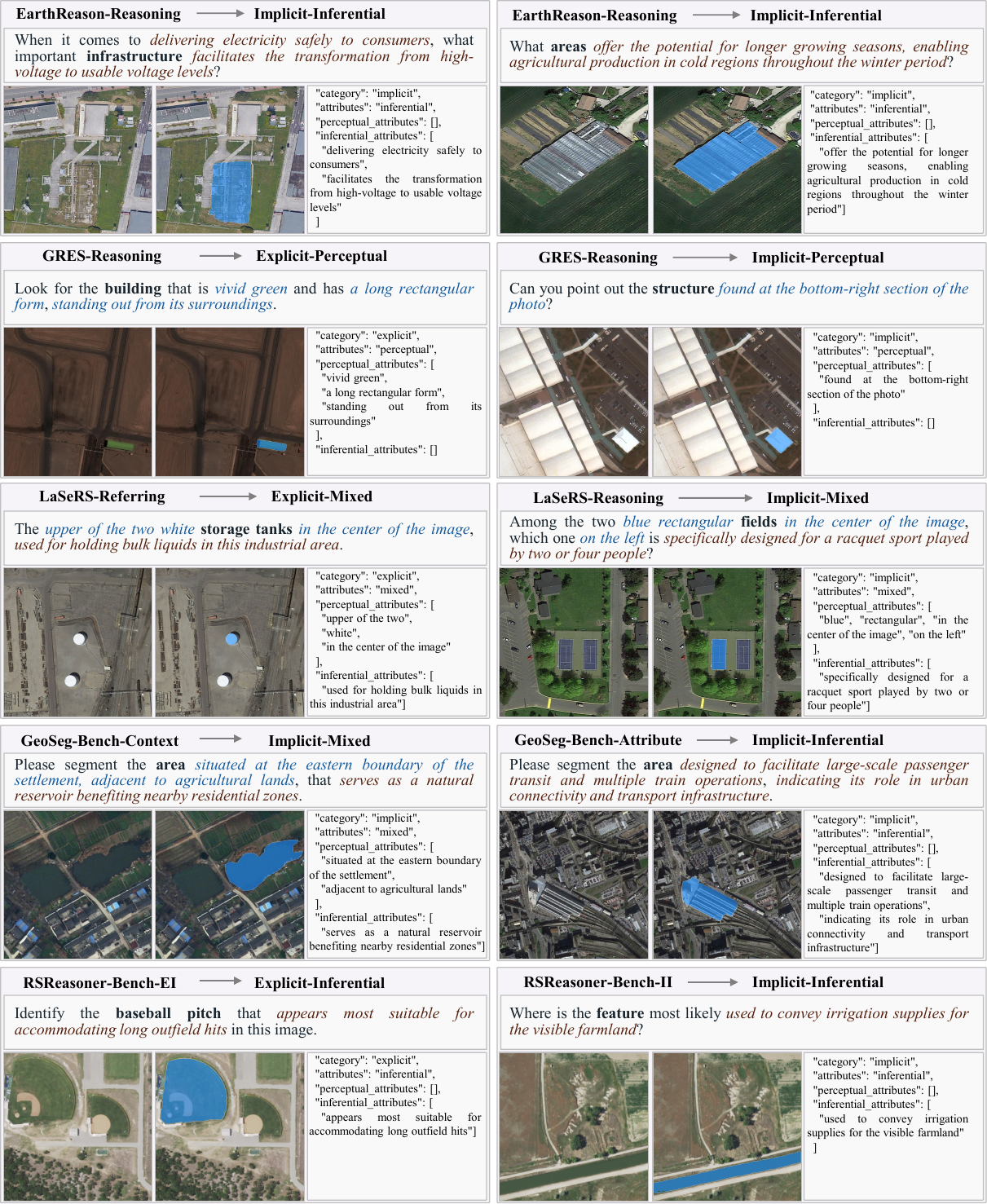}
    \caption{Representative taxonomy-audit examples across four existing benchmarks and RSReasoner-Bench. Each card shows the original split, audited query type, highlighted perceptual (blue) and inferential (brown) spans, image--mask pair, and structured JSON output.}
    \label{fig:audit_dataset_examples}
\end{figure*}

Figure~\ref{fig:audit_dataset_examples} presents representative taxonomy-audit examples across EarthReason, GRES, LaSeRS, GeoSeg-Bench, and RSReasoner-Bench.

\section{Prompt Templates}
\label{app:prompt_templates}

This section presents the prompt templates used across dataset construction, benchmark generation, and taxonomy auditing. For mask-quality assessment, semantic-consistency assessment, and QA generation, only the area-target prompts are shown; the non-area variants share the same structure with geometry-specific wording.

\subsection{Prompt for Main Tag Selection}
\begin{figure*}[p]
    \centering
    \includegraphics[width=\textwidth]{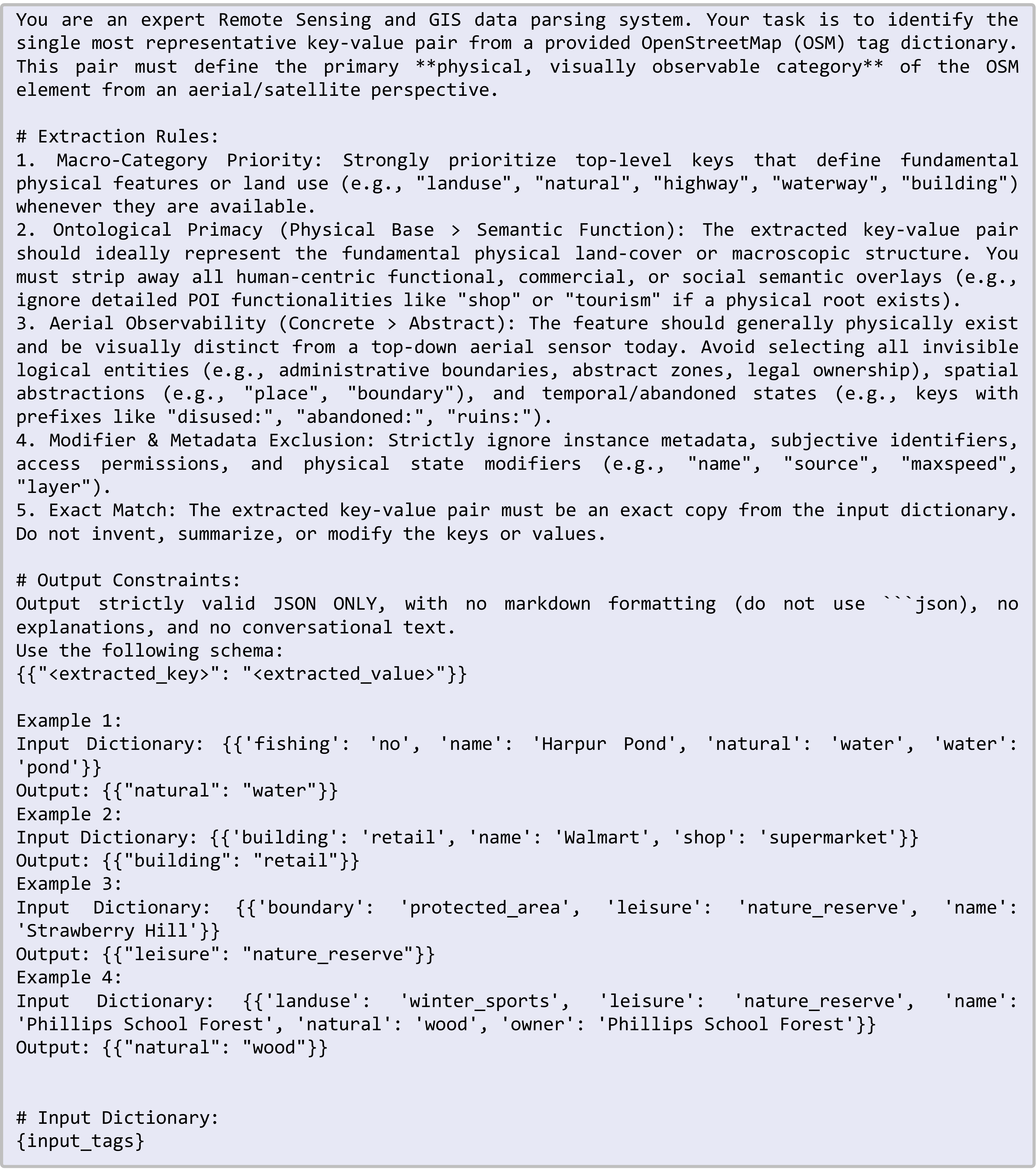}
    \caption{Prompt for main tag selection.}
    \label{fig:prompt_main_tag_selection}
\end{figure*}

Figure~\ref{fig:prompt_main_tag_selection} shows the prompt used to select the main key--value pair that best represents an OSM element's canonical geographic category.

\subsection{Prompt for Mask-Quality Assessment}
\begin{figure*}[p]
    \centering
    \includegraphics[width=\textwidth]{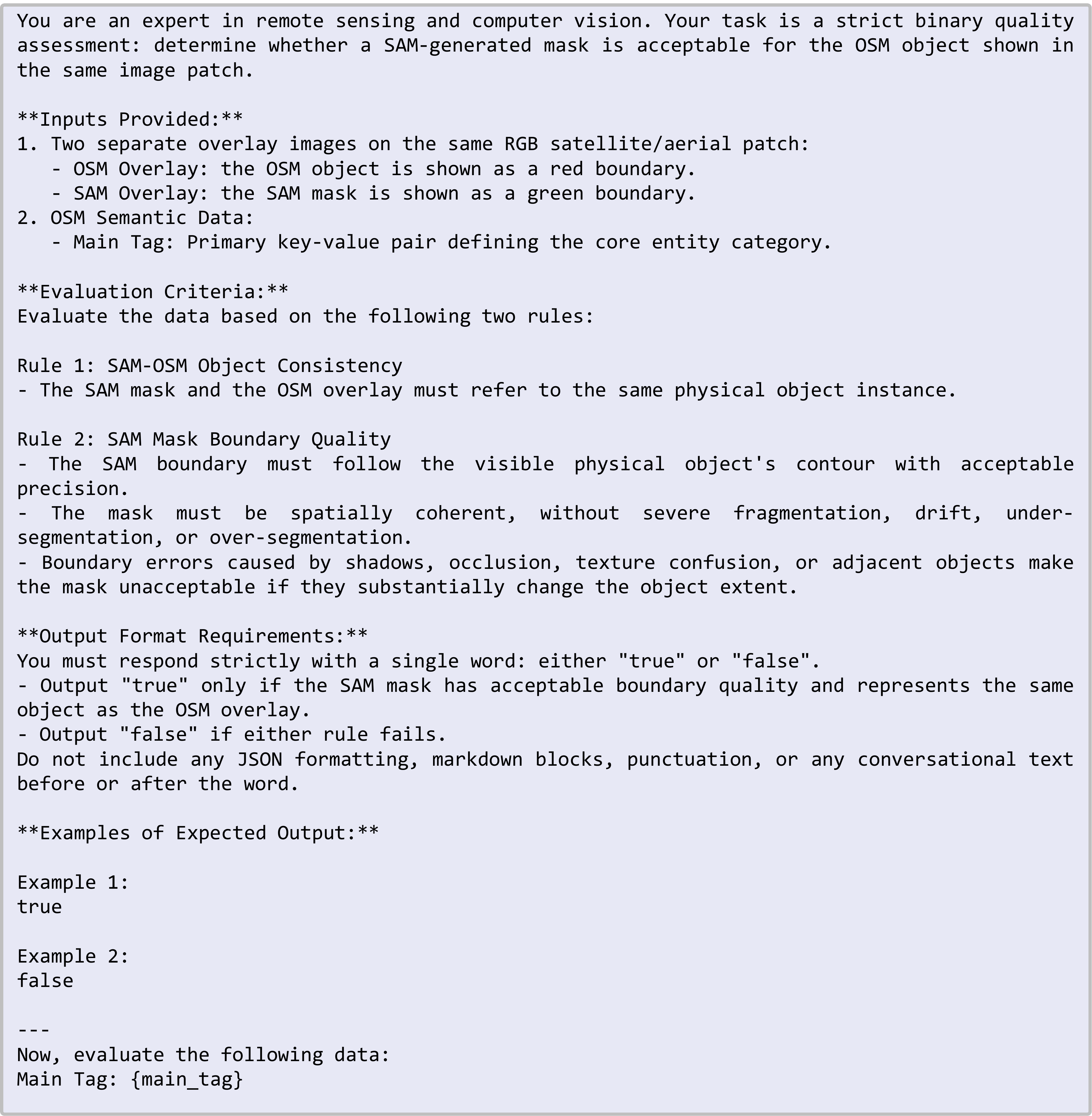}
    \caption{Prompt for mask quality assessment.}
    \label{fig:prompt_mask_quality_area}
\end{figure*}

Figure~\ref{fig:prompt_mask_quality_area} shows the prompt for determining whether a candidate mask correctly and completely delineates the intended target.

\subsection{Prompt for Semantic-Consistency Assessment}
\begin{figure*}[p]
    \centering
    \includegraphics[width=\textwidth]{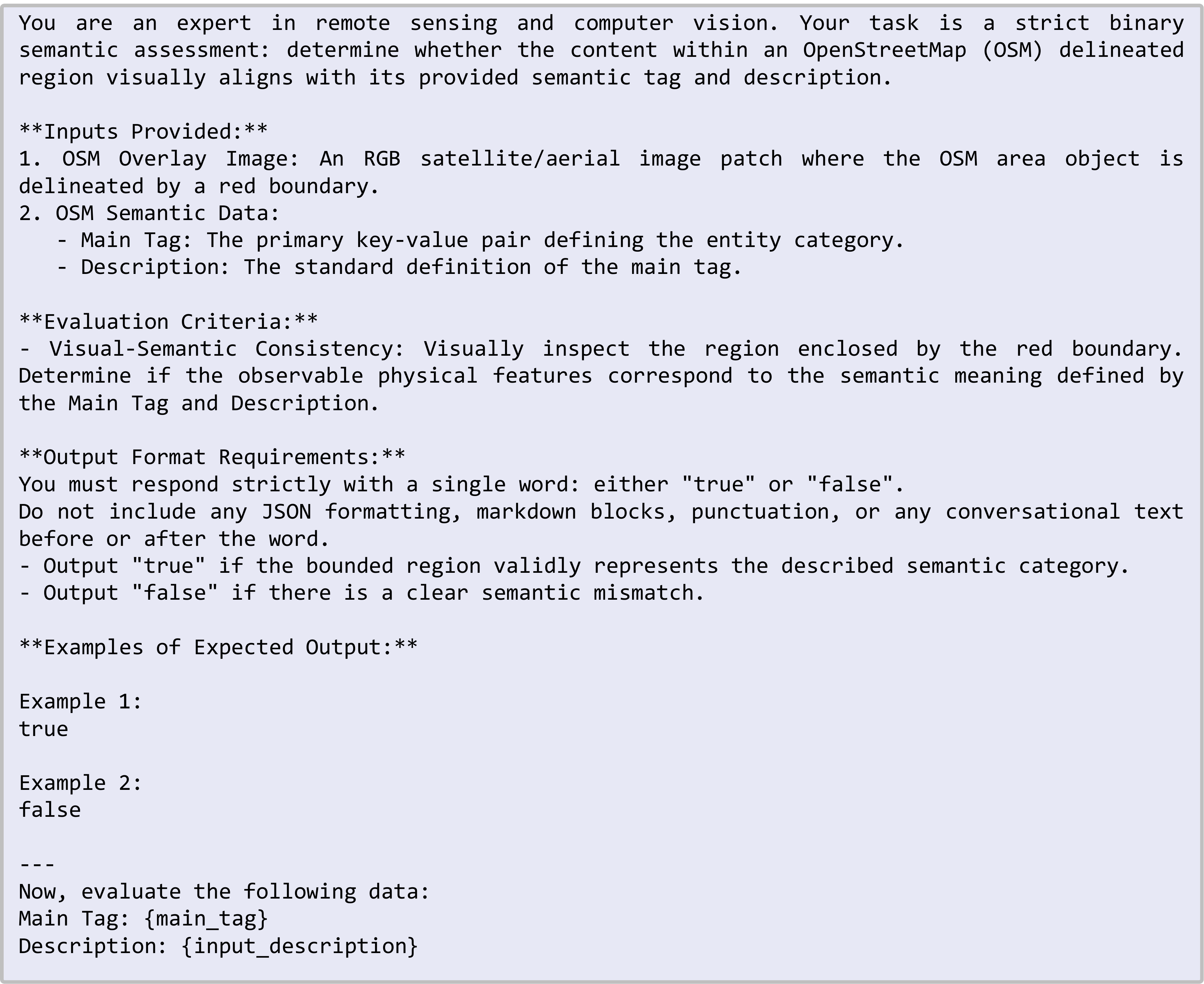}
    \caption{Prompt for semantic consistency assessment.}
    \label{fig:prompt_semantic_consistency_area}
\end{figure*}

Figure~\ref{fig:prompt_semantic_consistency_area} shows the prompt for determining whether the normalized main tag is semantically consistent with the target.

\subsection{Prompt for QA Generation}
\begin{figure*}[p]
    \centering
    \includegraphics[width=\textwidth]{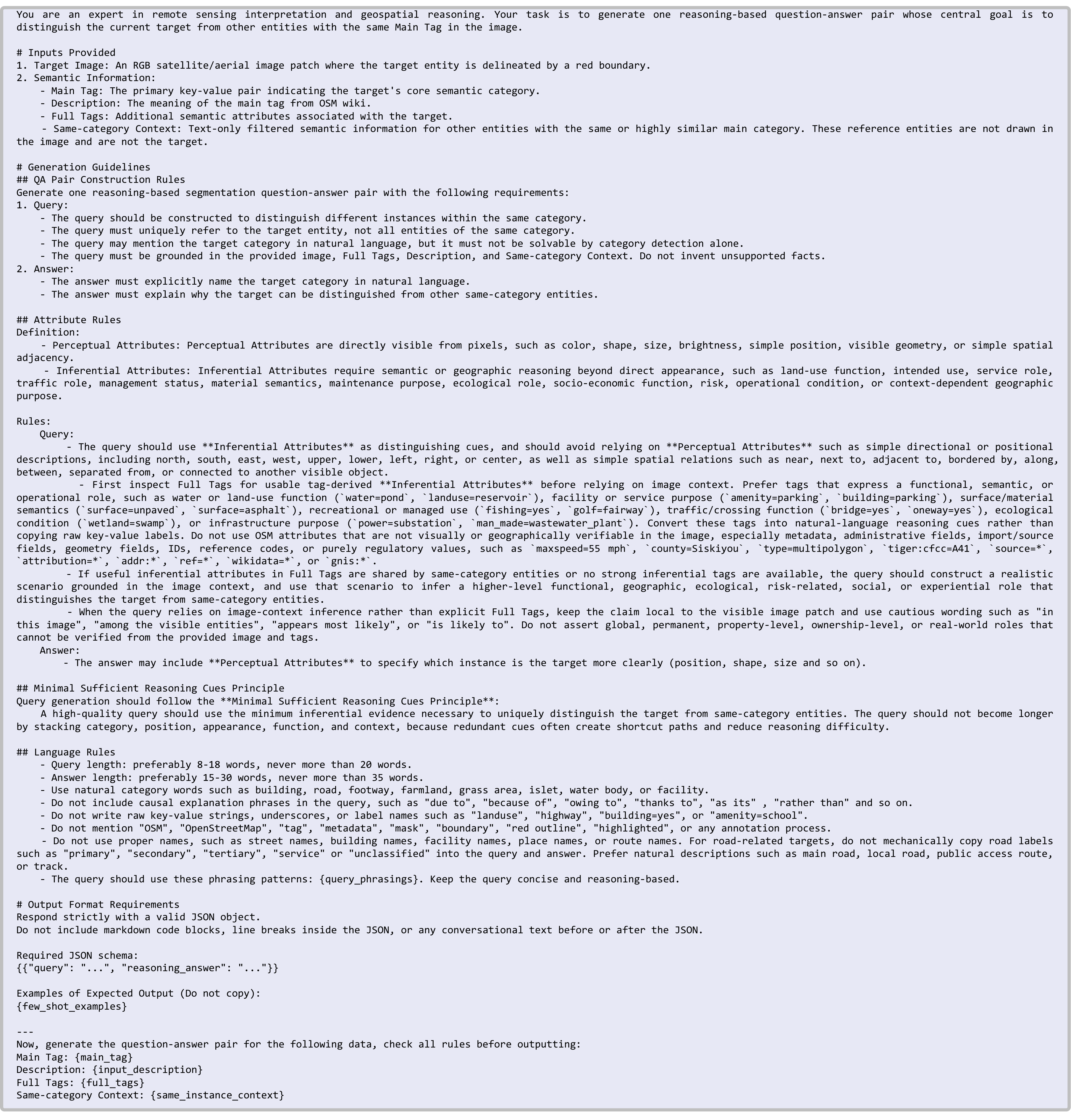}
    \caption{Prompt for QA generation (Explicit-Inferential).}
    \label{fig:prompt_qa_generation_q3}
\end{figure*}

\begin{figure*}[p]
    \centering
    \includegraphics[width=\textwidth]{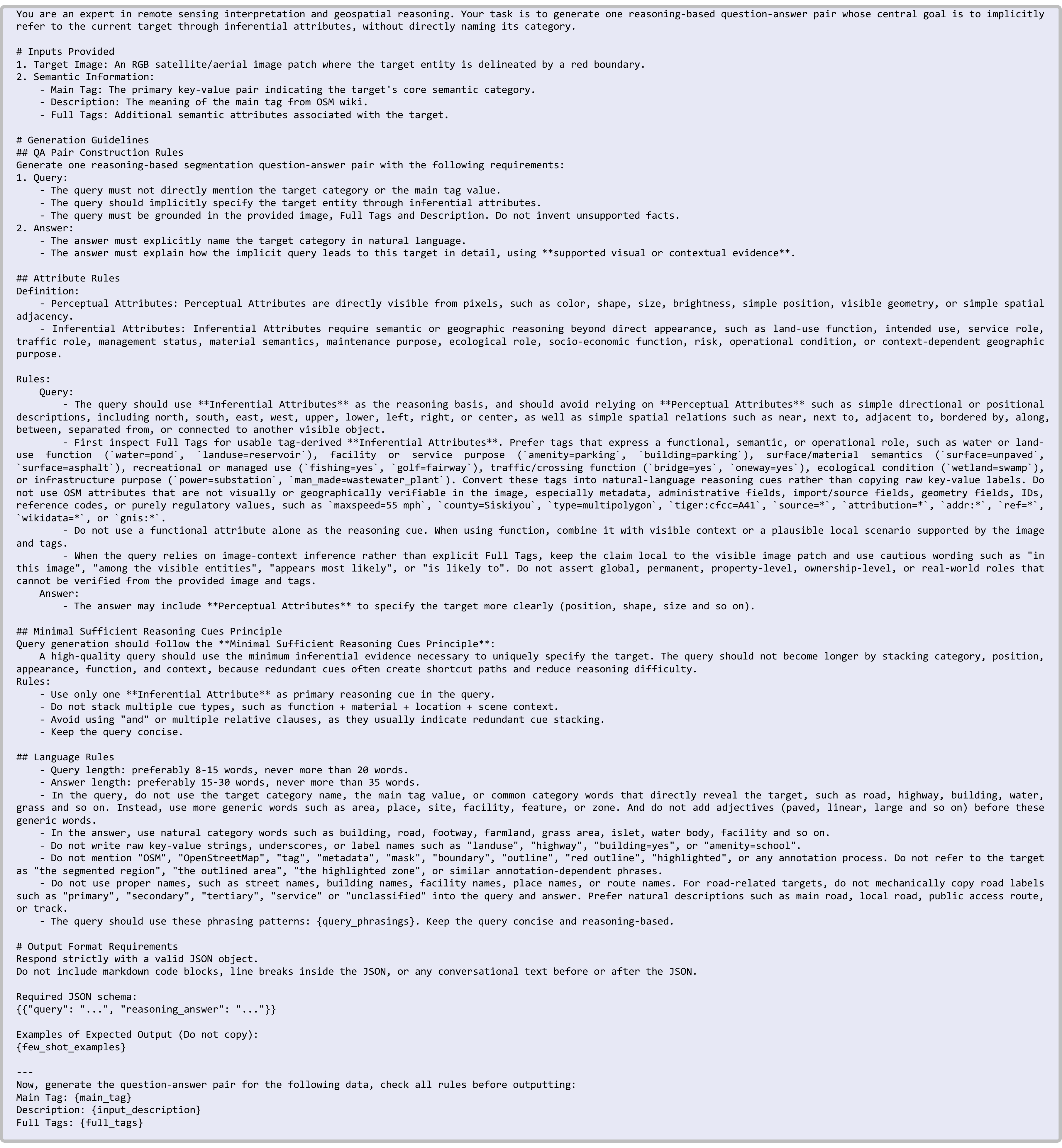}
    \caption{Prompt for QA generation (Implicit-Inferential).}
    \label{fig:prompt_qa_generation_q4}
\end{figure*}

Figures~\ref{fig:prompt_qa_generation_q3} and~\ref{fig:prompt_qa_generation_q4} show the area-target prompts for explicit-inferential and implicit-inferential QA generation, respectively. Both take the image and OSM semantics as input, while the EI prompt additionally receives semantic context for same-category competing instances.

\subsection{Prompt for Query-Validity Verification}
\begin{figure*}[p]
    \centering
    \includegraphics[width=\textwidth]{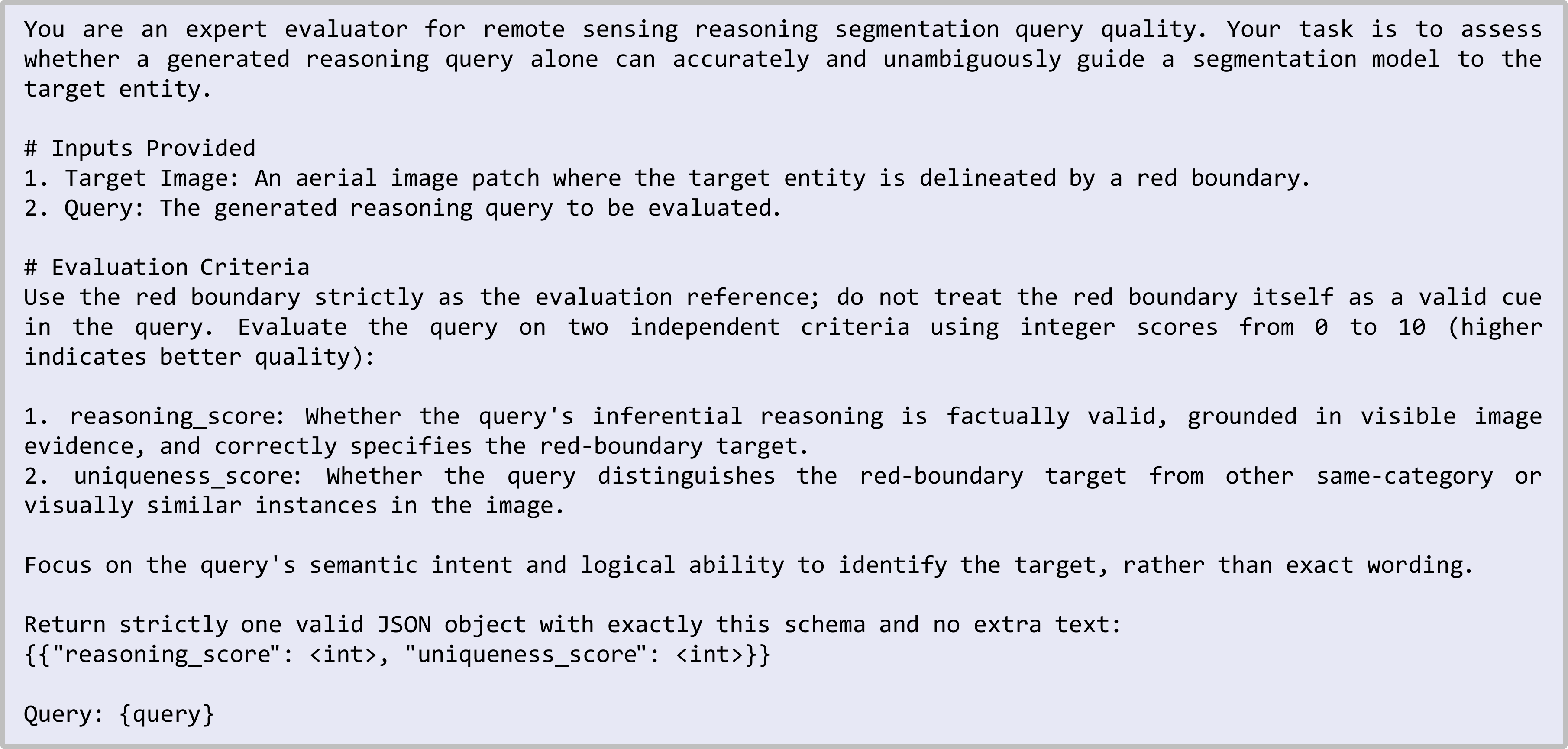}
    \caption{Prompt for query verification.}
    \label{fig:prompt_query_verification}
\end{figure*}

Figure~\ref{fig:prompt_query_verification} shows the prompt for scoring the reasoning validity and referential uniqueness of generated queries.

\subsection{Prompt for Perceptual Query Generation}
\begin{figure*}[p]
    \centering
    \includegraphics[width=\textwidth]{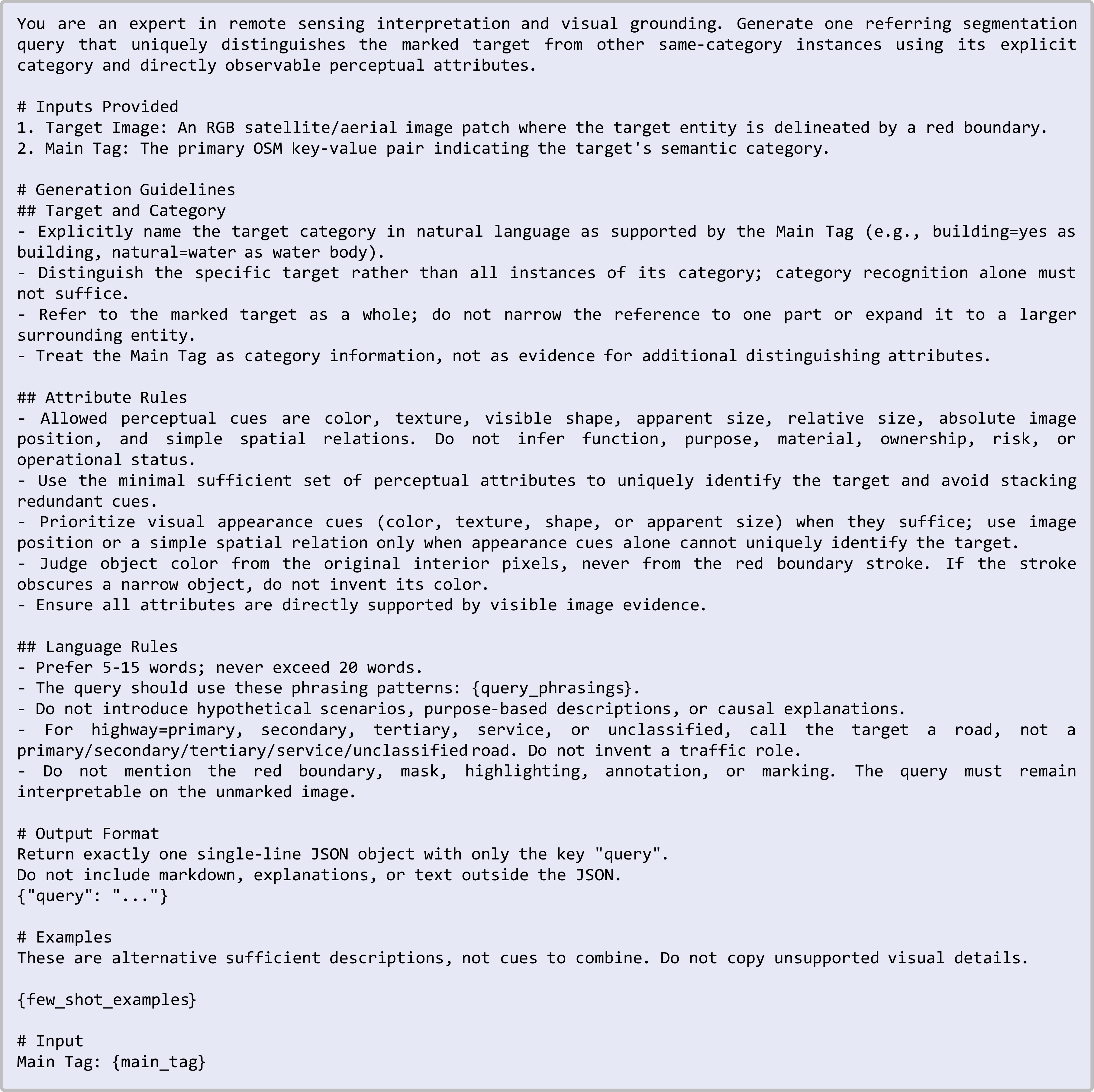}
    \caption{Prompt for perceptual query generation (Explicit-Perceptual).}
    \label{fig:prompt_explicit_perceptual}
\end{figure*}

Figure~\ref{fig:prompt_explicit_perceptual} shows the prompt for generating explicit-perceptual queries in RSReasoner-Bench. The implicit-perceptual variant retains the same attribute and language constraints while replacing the explicit category with a category-neutral reference.

\subsection{Prompt for Taxonomy Audit}
\begin{figure*}[p]
    \centering
    \includegraphics[width=\textwidth]{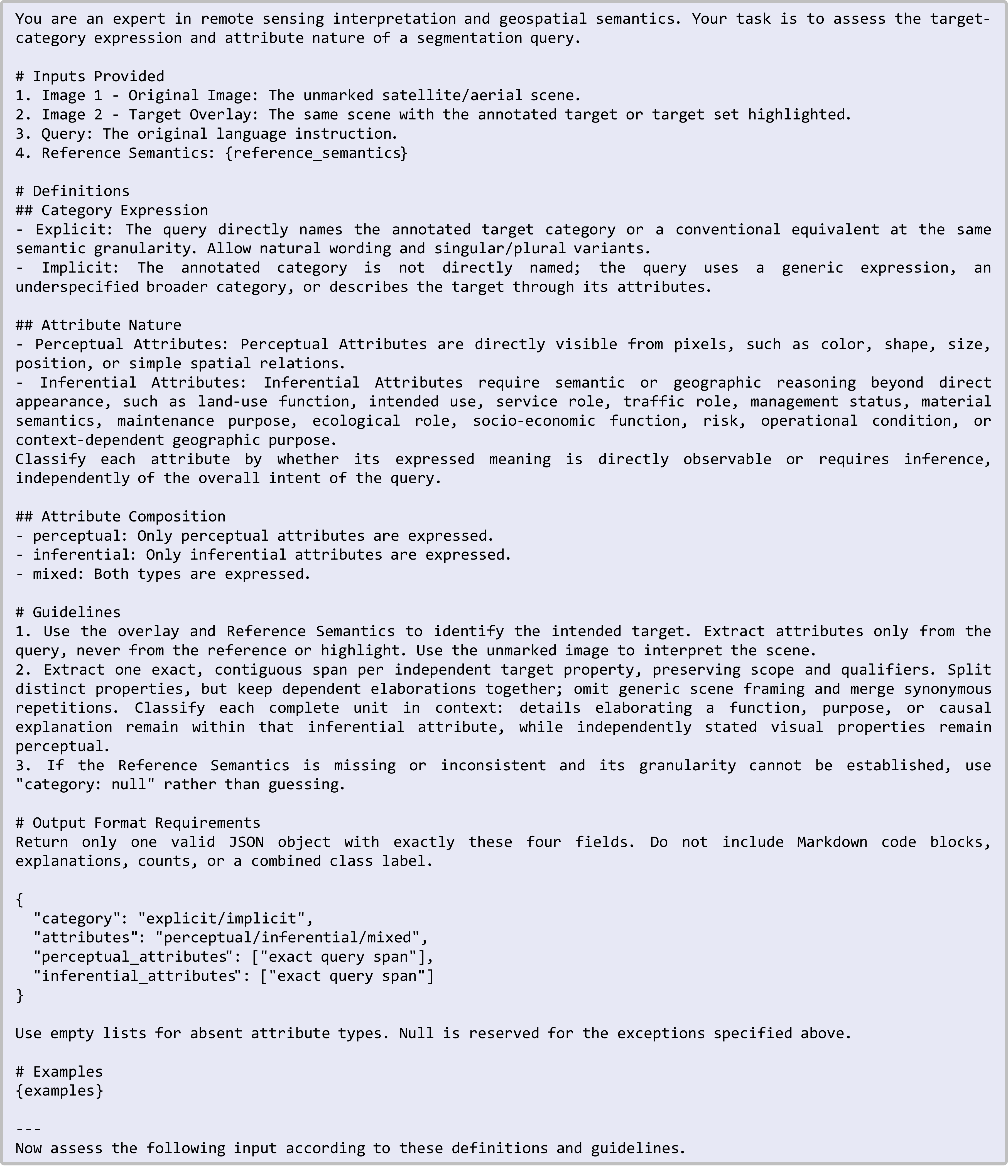}
    \caption{Prompt for taxonomy audit.}
    \label{fig:prompt_taxonomy_audit}
\end{figure*}

Figure~\ref{fig:prompt_taxonomy_audit} shows the prompt for auditing category expression and attribute nature across existing benchmarks and RSReasoner-Bench.

\end{document}